\documentclass[conference]{IEEEtran}
\IEEEoverridecommandlockouts
\usepackage{cite}
\usepackage[colorlinks,
            linkcolor=blue,       
            anchorcolor=blue,          
            citecolor=blue,      
            urlcolor=blue]{hyperref}
            
\usepackage{amsmath,amssymb,amsfonts}
\usepackage{algorithmic}
\usepackage{graphicx}
\usepackage{textcomp}
\usepackage{xcolor}
\usepackage{mathtools}
\usepackage{algorithm}
\usepackage{algorithmic}
\usepackage{multirow}
\usepackage{aisecure-math}
\usepackage{color}
\usepackage{diagbox}
\usepackage{xspace}
\newcommand{\pV}{{\mathcal{V}}}
\newcommand{\pT}{{\mathcal{T}}}

\usepackage{fancyhdr}
\usepackage{xspace}
\usepackage{enumitem}
\usepackage{xcolor}
\newcommand{\name}{CARE\xspace}
\newcommand{\blackhref}[1]{\hypersetup{urlcolor=black}\href{#1}} 

\def\BibTeX{{\rm B\kern-.05em{\sc i\kern-.025em b}\kern-.08em
    T\kern-.1667em\lower.7ex\hbox{E}\kern-.125emX}}
\begin{document}

\title{\name: Certifiably Robust Learning with Reasoning via Variational Inference}
\author{
    \IEEEauthorblockN{
        Jiawei Zhang\IEEEauthorrefmark{1} 
        \quad
        Linyi Li\IEEEauthorrefmark{1}
        \quad
        Ce Zhang\IEEEauthorrefmark{2}
        \quad
        Bo Li\IEEEauthorrefmark{1}
    }

  \IEEEauthorblockA{\IEEEauthorrefmark{1} University of Illinois Urbana-Champaign, USA, \{\blackhref{mailto:jiaweiz7@illinois.edu}{jiaweiz7}, \blackhref{mailto:linyi2@illinois.edu}{linyi2}, \blackhref{mailto:lbo@illinois.edu}{lbo}\}@illinois.edu}
  \IEEEauthorblockA{\IEEEauthorrefmark{2} ETH Zürich, Switzerland, \blackhref{mailto:ce.zhang@inf.ethz.ch}{ce.zhang@inf.ethz.ch}}
}

\maketitle
\thispagestyle{fancy} 
\lhead{}
\chead{} 
\rhead{}
\lfoot{}
\renewcommand{\headrulewidth}{0pt}
\renewcommand{\footrulewidth}{0pt}
\pagestyle{fancy}
\cfoot{\thepage}

\begin{abstract}
Despite great recent advances achieved by deep neural networks (DNNs), they are often vulnerable to adversarial attacks.
Intensive research efforts have been made to improve the robustness of DNNs; however, most \textit{empirical} defenses can be adaptively attacked again, and the \textit{theoretically} certified robustness is limited, especially on large-scale datasets. 
One potential root cause of such vulnerabilities for DNNs is that although they have demonstrated powerful expressiveness, they lack the reasoning ability to make robust and reliable predictions. 
In this paper, we aim to integrate domain knowledge to enable robust learning with the \textit{reasoning} paradigm. In particular, we propose a \underline{c}ertifiably robust le\underline{a}rning with \underline{re}asoning pipeline (\name), which consists of a \textit{learning} component and a \textit{reasoning} component.
Concretely, we use a set of standard DNNs to serve as the learning component to make semantic predictions (e.g., whether the input is furry), and we leverage the probabilistic graphical models, such as Markov logic networks (MLN), to serve as the reasoning component to enable knowledge/logic reasoning (e.g., 
\texttt{IsPanda} $\implies$ \texttt{IsFurry}).
However, it is known that the exact inference of  MLN (reasoning) is  \#P-complete, which limits the scalability of the pipeline. To this end, we propose to approximate the MLN inference via variational inference based on an efficient expectation maximization algorithm. In particular, we leverage graph convolutional networks (GCNs) to encode the posterior distribution during variational inference and update the parameters of GCNs (E-step) and the weights of knowledge rules in MLN (M-step) iteratively. 
We conduct extensive experiments on different datasets such as AwA2, Word50, GTSRB, and PDF malware, and we show that \name achieves significantly higher certified robustness (e.g., the certified accuracy is improved from $36.0\%$ to $61.8\%$ under $\ell_2$ radius $2.0$ on AwA2) compared with the state-of-the-art baselines. We additionally conducted different ablation studies to demonstrate the empirical robustness of \name and the effectiveness of different knowledge integration. The official code is available at~\textcolor{blue}{\url{https://github.com/javyduck/CARE}}.

\end{abstract}


\section{Introduction}
Despite that machine learning (ML), especially deep neural networks (DNNs), 
have achieved great successes in different applications, they are also found
to be vulnerable to small and adversarial perturbations that could lead to incorrect predictions~\cite{biggio2013evasion,szegedy2014intriguing,xiao2018generating,xiao2018spatially}. 
Given the massive deployment of machine learning systems, especially in safety-critical scenarios such as automatic driving~\cite{cao2021invisible,eykholt2018robust} and medical diagnosis~\cite{erickson2017machine,magoulas1999machine}, improving the robustness of ML models is of great importance, and a reliable defense mechanism is in dire need. 

To overcome such adversarial attacks,  significant efforts have been made to develop different defense approaches, both empirically and theoretically~\cite{xiao2018characterizing,yang2021trs,yang2022on,samangouei2018defense,cohen2019certified}. However, most of these existing empirical defenses have been attacked successfully again by strong adaptive attacks~\cite{carlini2017adversarial,athalye2018obfuscated}; and the theoretically certifiably robust models are usually limited on large-scale data~\cite{gowal2018effectiveness,zhang2019towards,salman2019convex,yang2020randomized}. 
On the other hand, existing ML models lack logical reasoning abilities, which may be one main root cause of their vulnerabilities. For instance, a human would be able to recognize a stop sign by just seeing the octagon shape, while DNNs cannot reason based on such knowledge.
Thus, in this paper, we aim to explore the question: \emph{Can we integrate domain knowledge into statistical learning with DNNs to improve their robustness? Will the certified robustness of ML models be improved when composed with a reasoning component?} 
\textit{Can we do such integration in an efficient and scalable way?}

\begin{figure*}
    \centering
    \includegraphics[width=\textwidth]{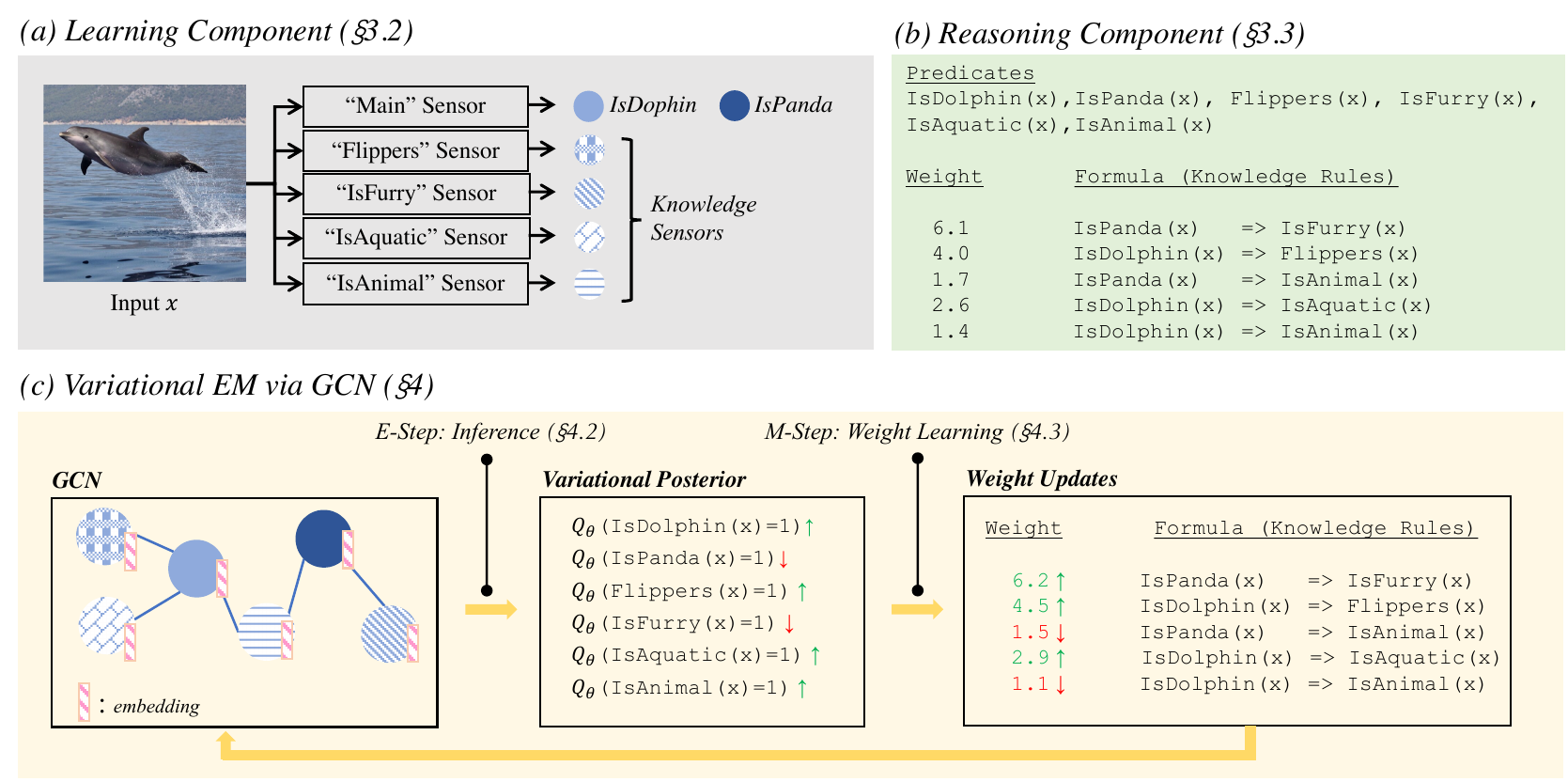}
    \vspace{-8mm}
    \caption{The overview of our \textit{learning with reasoning} framework \name. 
    }
    \label{fig:pipeline}
    \vspace{-7mm}
\end{figure*}

To effectively integrate knowledge rules to enable reasoning ability for existing DNN-based statistical learning, in this work, we propose 
a \textit{learning with reasoning} pipeline \name, which contains both a \textit{learning} and a \textit{reasoning} component. 
In particular, the learning component contains one \textit{main sensor} that is in charge of the main classification task (e.g., $d$-way animal prediction) and several \textit{knowledge sensors} that identify different semantic entities or attributes (e.g., whether the input is furry).
The output of different sensors will be taken as the input of the \textit{reasoning} component, which can be realized using probabilistic graphical models such as Markov logic networks (MLN)~\cite{richardson2006markov}. 
Concretely, different knowledge rules (e.g., ``Panda is furry") can be represented as the first-order logic rules and then embedded in the MLN to help perform logic reasoning. The overall pipeline of \name is shown in Figure~\ref{fig:pipeline}.
The advantage of such a pipeline is that the predictions of different sensors are dependent, following the logical relationships among them. Thus, given an attack against, say, the main sensor, the adversary not only needs to attack a set of sensors additionally but also needs to ensure that the attacked predictions of these sensors satisfy the logical relationships, making the attack much more challenging in practice.

Although such reasoning integration is very promising, as illustrated in 
a few recent seminal explorations~\cite{gurel2021knowledge,yang2020end},
scalability and efficiency 
hinder their real-world applications ---
the inference of MLN is \textsf{\#P}-complete~\cite{richardson2006markov}, which is exponential in the number of the possible predictions of sensors within the logical relationships, and thus impedes the scalability of such pipelines. 
As a result, 
recent attempts in using reasoning to improve
robustness can only handle relatively small-scale problems~\cite{yang2020end}.
The key technical contribution
of this paper is to
approximate the MLN inference with variational inference based on parametrized graphical convolutional networks (GCN)~\cite{2016tc}.
Currently, in addition to the classical inference approximation such as Markov chain Monte Carlo (MCMC)~\cite{gilks1995markov,6413813}, and loopy belief propagation~\cite{murphy2013loopy}, variational method~\cite{jordan1999introduction} which approximates probability densities through optimization has become more and more efficient and convenient in practice owing to advanced learning strategies. 
On a high level, the variational method approximates the posterior distribution with a given approximating function family $\mathcal{Q}$, and thus the design of such function family largely affects the final approximation. 
More specifically, this function family should satisfy the following two requirements: (1)~it should capture the topology of the knowledge/logic relationship structure;  (2)~it should be scalable and can be optimized effectively on large-scale datasets. To this end, we follow the observations of existing works~\cite{qu2019gmnn,zhang2019efficient} and adopt GCN to serve as the approximating function, which can efficiently represent the large knowledge graph structure. 

Concretely, we will map each sensor prediction (e.g., Panda) as a node within GCN and the logical relationships between sensors as edges (e.g., an edge between sensors predicting ``Panda" and ``furry" to represent the rule 
\texttt{IsPanda(x)} $\implies$ \texttt{IsFurry(x)}). 
Since the inference of GCN scales linearly in the number of graph edges,  the corresponding approximated inference of MLNs can thus scale linearly in the number of knowledge rules, which makes the \name scalable to large-scale problems.
In particular, we propose an efficient expectation maximization algorithm to iteratively update the weights of GCN (E-step) and the weights of logic rules within MLN (M-step).


This allows us to apply \name 
to an unprecedented scale on problems that
involve logical reasoning to improve 
robustness.
To demonstrate the robustness of \name, we conduct extensive experiments on four large-scale datasets: Animals with Attributes~(AwA2)~\cite{xian2018zero}, Word50~\cite{chen2015learning}, 
GTSRB~\cite{stallkamp2012man}, and PDF malware dataset from Contagio~\cite{parkour201716}. For AwA2, we leverage the annotated attributes of each animal class and the hierarchy relationship among different animal categories extracted from WordNet~\cite{miller1995wordnet} as our knowledge rules. For Word50, we leverage the positions of each character in the known words as knowledge rules. For GTSRB, we use the road sign properties such as shape and the content of each sign as knowledge rules; while for the PDF malware dataset, the common benign/malicious traces (features)  (e.g., $\texttt{/Root/Pages/Contents/Filter}$ for benign trace and $\texttt{/Root/OpenAction}$ for malicious trace) and their relationships are used to construct the knowledge rules. 
We show that \name significantly outperforms the SOTA \emph{certified} defenses~\cite{salman2019provably,jeong2020consistency,liu2020enhancing,lee2019tight} under different perturbation radii. 
We also conduct different ablation studies to further understand the impacts of the number of integrated knowledge rules, the empirical robustness of \name, and the robustness and explanation properties of \name based on case studies.

\underline{\textbf{Technical Contributions.}} In this paper, we provide a scalable certifiably robust \textit{learning with reasoning} pipeline \name, which has demonstrated significantly higher certified robustness than baselines on large-scale image datasets as well as  PDF malware dataset. 
\begin{itemize}[leftmargin = *]
    \item We propose a scalable and certifiably robust \textit{learning with reasoning} pipeline \name, which is able to integrate knowledge rules to enable reasoning ability for reliable prediction.
        
    \item We propose an efficient expectation maximization algorithm to approximate the reasoning (MLN) inference via variational inference using GCN. 
    
    
        
    \item We conduct extensive experiments on a wide range of datasets and demonstrate that \name achieves \textit{significantly} higher certified robustness than SOTA baselines. For instance,~\name improves the certified accuracy from $36.0\%$ (SOTA) to $61.8\%$ under $\ell_2$ radius $2.0$ on  AwA2; and for the word-level classification on Word50,~\name improves the certified accuracy from $24.8\%$ (SOTA) to $73.6\%$ under $\ell_2$ radius $0.5$; on the GTSRB dataset, \name improves the certified accuracy from $83.3\%$ (SOTA) to $84.4\%$ under $\ell_2$ radius $0.4$; for PDF malware,~\name improves the certified accuracy from $22.6\%$ (SOTA) to $54.5\%$ under $\ell_0$ radius $7$.
    
    \item We conduct a set of ablation studies to explore the impact of the number of integrated knowledge rules; demonstrate the high empirical robustness of \name compared with baselines; and showcase the robustness and explanation properties of \name based on case studies.
    
\end{itemize}

\vspace{-3mm}
\section{Background}
 \vspace{-1mm}
 
\textbf{Markov Logic Networks} (MLN) provides an effective approach to combining first-order logic and probabilistic graphical models in a unified representation. Concretely, MLN
can be viewed as a first-order knowledge base with a weight attached to each logic formula, where the first-order logic formula can be used to model different types of domain knowledge such as ``\texttt{IsPanda(x)} $\implies$ \texttt{IsFurry(x)}".
Formally, in MLN, the mapping (prediction) among entities can be represented as \textit{predicates} $t(\cdot)$, which is a logic function defined over the entity variable set $\mathcal{V} = \{v_1, ..., v_N\}$ where $v_i$ denotes a \textit{constant} in the logic world. For instance, a constant can be a ``stop sign" or a ``octagon shape". The predicate is thus defined as 
$t(\cdot): \pV \times ... \times \pV  \to \{0,1\}$.
In the meantime, a logic \textit{formula} in MLN is defined over the composition of a set of predicates as $f(\cdot): \pV \times ... \times \pV  \to \{0,1\}$. 
For instance, given an input instance $x$, a model that is trained to predict whether the input is of the octagon shape $g_{octagon} (x)$ can be viewed as a predicate. The related knowledge rules, such as ``stop sign $\implies$ octagon shape" can be represented as a formula as 
\vspace{-2mm}
$$
g_{stop} (x) \implies g_{octagon} (x).
\vspace{-2mm}
$$
A formula consists of different predicates.
We denote the assignments of variables to the arguments of a formula $f$ as $a_f$, and all the possible consistent assignments are represented as set $\mathcal{A}_f = \{a_f^1, a_f^2, ... \}$.
With a particular set of constants assigned to the arguments of a predicate, it is called a \textit{ground predicate}. For instance, with an assignment for a predicate $a_t = (c_1, c_2)$, we can simply write a ground predicate $t(c_1, c_2) = t(a_t)$.
Similarly, a formula with an assignment to its arguments is called a \textit{ground formula} (e.g., $f(a_f)$). 
Based on the probabilistic logic representations, MLN can thus be defined formally as a joint distribution over all possible assignments in set $\mathcal{A}_f$ for the formula set $\mathcal{F}$:
\vspace{-2.5mm}
\begin{equation}
\label{eq:mln_standard}
    P_w(t_1, ..., t_L)=\frac{1}{Z(w)} \exp \left(\sum_{f \in \mathcal{F}} w_{f} \sum_{a_f \in \mathcal{A}_f} \phi_f(a_f)\right),
    \vspace{-2mm}
\end{equation}
where $t_1, ..., t_L$ denote the $L$ ground predicates (with assignment $\mathcal{A}_f$) that are used to form the formulas, $w_f$ represents the corresponding weight for each formula,  
Note that $t_i(x)$ is a predicate function given input $x$, and for notation simplicity we will use $t_i$ throughout this work to represent $t_i(x)$ when there is no ambiguity.
$\phi_f(\cdot)$ represents the potential function for the given assignment, which takes 1 when the formula is true and 0 when it is false, and $Z(w)$ is the \textit{partition function} summing over 
all possible assignments.
Based on the grounding predicates, we can define a \textit{possible world} by assigning a truth value to each possible ground predicate. 

\textbf{Robustness certification.} 
The robustness certification technique aims to provide a certified robustness guarantee: given a robust radius $r \in \mathbb{R}_+$,  any perturbation within $r$ will not change the classifier's prediction~\cite{li2023sok,liu2021algorithms}.
Formally, such technique takes a classifier $h: \mathbb{R}^d \to \mathcal{Y}$ and a clean~(i.e., unperturbed) input $x$.
It outputs $r$ such that $h(x) = h(x')$ holds for any perturbed input $x'$ with $d(x,x') < r$ under a specific metric $d$~(e.g., $\ell_p$ norm).
We provide more details about existing robustness certification techniques in our related work \Cref{sec:related-work}.
In this paper, we mainly utilize the randomized smoothing technique~\cite{cohen2019certified}, one of the state-of-the-art certification methods that can scale to large-scale datasets~\cite{jeong2020consistency,li2023sok,carlini2022certified}, to evaluate the certified robustness for different learning pipelines as follows.
First, we will wrap a given learning model $h$ to a new smoothed model $g(x)=\argmax_{c\in \gY} \sP(h(x+\delta)=c)$ where the $\delta \sim \mathcal{N}\left(0, \sigma^{2} I\right)$ and the $\sigma$ here control the variance of the added noise. Then, the resulting Gaussian smoothed classifier $g(x)$ can be certified by leveraging Neyman-Pearson lemma with no further assumption. Assume $p_A$ is the probability of the returning class $c_{A}$, i.e., $p_A = \mathbb{P}\left(h(x+\delta)=c_{A}\right)$, and $p_B$ is the ``runner-up'' probability, i.e., $p_{B}=\max _{c \neq c_{A}} \mathbb{P}(h(x+\varepsilon)=c)$,  the smoothed classifier $g$ is robust around $x$ with the radius~\cite{cohen2019certified}:
\vspace{-2mm}
\begin{equation}
 R=\frac{\sigma}{2}\left(\Phi^{-1}\left(p_{A}\right)-\Phi^{-1}\left(p_{B}\right)\right),
 \label{eq:radius}
 \vspace{-3mm}
\end{equation}
where $\Phi^{-1}$ is the inverse of the standard Gaussian CDF. That is to say, it is guaranteed that there is no further adversarial perturbation within $R$, and thus the robustness can be certified.

\vspace{-3mm}
\section{\name: Scalable Robust Learning with Reasoning}
\label{sec:sector-pipeline}
\vspace{-1mm}
In this section, we first provide an overview of the proposed  \textit{learning with reasoning} pipeline \name, followed by the detailed construction of the \textit{learning} and \textit{reasoning} components within the pipeline.
\vspace{-2.5mm}
\subsection{Overview of \name}
\vspace{-1mm}
To effectively integrate domain knowledge into statistical machine learning models (e.g., DNNs), we propose \name, which consists of a \textit{learning} and a \textit{reasoning} components. 
    In particular, the  \textit{learning} component consists of a \textit{main sensor} which serves for the main classification task and makes a multi-class prediction given an input; and several \textit{knowledge sensors} which make predictions for the individual semantic objects requested by different knowledge rule given the same input. 
    For instance, if we want to integrate the knowledge ``Panda is furry'' into the learning process, we will train a main sensor to predict the class of the input (e.g., different animal categories), and a knowledge sensor to predict whether the input ``is furry", respectively.
    We then represent the knowledge as the first-order logic rule ``\texttt{IsPanda(x)} $\implies$ \texttt{IsFurry(x)}" via a reasoning component.
    
    Concretely, the \textit{reasoning} component 
    can be realized by different probabilistic graphical models, such as Markov logic networks (MLN) and Bayesian networks. In this work, we  focus on MLN as the reasoning component for different applications. However, it is  known that the inference of MLN is computationally expensive due to the exponential cost of constructing the ground Markov network. Thus, we propose a scalable variational inference approach based on GCN to approximate the inference of MLN (Section~\ref{sec:sector-gnn}).
    In particular, we propose an EM algorithm to jointly improve the accuracy of GCN in terms of learning the MLN~(E step) and optimize the weights of formulas in the latent MLN towards better inference-time robustness~(M step).
    Moreover, we consider different types of domain knowledge, such as attributes-based knowledge and category hierarchy knowledge, to improve the robustness of \name (details in Section~\ref{sec:exp}). 
 
    Since the proposed \name can be viewed as a general machine learning pipeline, we are able to certify its robustness using standard certification approaches~\cite{cohen2019certified,salman2019provably,jeong2020consistency,lee2019tight}. 
    As the pure data-driven based machine learning approaches have reached a bottleneck for certified robustness so far due to the lack of additional information or prior knowledge, here we show that the proposed \name is able to \textit{significantly} improve the certified robustness on datasets including the high-resolution dataset AwA2~\cite{xian2018zero},  the standard Word50~\cite{chen2015learning}, the GTSRB~\cite{stallkamp2012man} for road sign classification, and the security application of PDF Malware classification~\cite{parkour201716}. In addition, we will also show that simply adding more prediction models as an ensemble without explicit knowledge integration or reasoning,  which is shown to obtain only marginal robustness improvement by existing work~\cite{liu2020enhancing, yang2022on}, will not help achieve such high performance on certified robustness. We believe such knowledge integration and enabling reasoning ability is a promising way to break the existing robustness barriers.

\vspace{-4mm}
\subsection{\textit{Learning} in \name}
\vspace{-2mm}
Within the \textit{learning} component of \name, we construct a set of statistical learning models (e.g., DNNs, logistic regression, SVMs, etc.) to predict the main classification task and other knowledge sensors' classification tasks. For instance, as shown in Fig~\ref{fig:pipeline}, the goal is to perform the animal category classification. In particular, we will train one main sensor to predict the main animal classes, say,~\emph{IsPanda},~\emph{Dolphin}, and others. In order to integrate domain knowledge and logical reasoning ability into this learning process, we need to embed domain knowledge, such as ``\texttt{IsPanda(x)} $\implies$ \texttt{IsFurry(x)}", into the pipeline. Thus, we will train a knowledge sensor to predict ``whether the input is furry". Similarly, we will train other knowledge sensors for different knowledge rules. 
Here the main/knowledge sensors can be viewed as predicates, and the output of each knowledge sensor is a binary truth value; while for multiclass classification, we will map the $d$-way prediction of the main sensor to several binary truth values (detailed mapping process and constraints in Section~\ref{sec:sector-gnn}).

Formally, we define the prediction output of i-\textit{th} sensor $t_i(\cdot)$ as $t_i$, and the corresponding prediction confidence as $z_i$.
Given an input $x$, the corresponding sensor predictions $t_i(x)$ are shown in Figure~\ref{fig:pipeline} (i.e., $t_i(x)=$ \texttt{IsFurry}($x$)).

\vspace{-2mm}
\subsection{\textit{Reasoning} in \name}
\vspace{-1mm}
Given an input $x$ and predictions from different sensors $t_i(x)$, we will connect these predictions based on their logical relationships to enable the \textit{reasoning} ability of our learning pipeline \name.  
Such a logical relationship can be realized by different types of probabilistic graphical models, and in this paper, we will focus on MLN.

Concretely, as mentioned above, we will construct one main sensor and several knowledge sensors $t_i(x)$ as the predicates in MLN.
We then build logical relationships among the predicates to form different formulas. 
Assume we have $L$ sensors,
an MLN will define a joint distribution based on the predefined logical formulas. 
For simplicity, we denote the collection of  formulas as $\mathcal{F}$, and thus the joint distribution defined by MLN can be represented as below simplified from~\Cref{eq:mln_standard}:
\vspace{-2mm}
\begin{equation}
\label{eq:joint_p}
    P_w(t_1,...,t_L) \coloneqq \frac{1}{Z(w)} \exp \left( \sum_{f \in \mathcal{F}} w_f f(t_1,...,t_L)\right),
    \vspace{-2mm}
\end{equation}
where $Z(w)$ denotes the partition function summing over all the possible assignments of the predicates. Since in our learning pipeline, each formula will only have one unique correct assignment by construction to ensure robustness, we can simplify this joint distribution based on~\Cref{eq:mln_standard}.

The reasoning component of \name can handle 
logic formulas expressed 
as first-order logic rules.
In this paper, we further optimize 
for four types of logic rules that
popularly used in practice as follows. 


$\bullet$ \emph{Attribute rule ($t_i \implies t_j \lor t_k \lor... $}):
Some prediction classes have specific attributes, which can be leveraged to construct knowledge rules.
For instance, one attribute rule could be $\texttt{IsPanda(x)} \implies \texttt{IsFurry(x)}$.

$\bullet$ \emph{Hierarchy rule ($t_i \implies t_j$}):
In general, there exist hierarchical relationships between different classes, based on which we can build formula $f(t_i, t_j) = \lnot t_i \lor t_j $. For instance,  $\texttt{IsDog($x$)} \implies \texttt{IsAnimal($x$)}$, or a slightly more complicated example as $\texttt{IsChihuahua($x$)} \lor \texttt{IsCollie($x$)}  \lor \texttt{IsDalmatian($x$)} \implies \texttt{IsDog($x$)}$. For instance, we can build such hierarchical knowledge rules based on relationships extracted from WordNet.

$\bullet$ \emph{Exclusion rule ($t_i \oplus t_j$}):
Some class predictions are naturally exclusive from others. For instance, an animal cannot be a panda and dolphin at the same time, so we will exclude the possible world where $\texttt{IsPanda}(x)$ and $\texttt{IsDolphin}(x)$ are both true. In particular, we will introduce constraint $\texttt{IsPanda}(x) \land \texttt{IsDolphin}(x) = \texttt{False}$.

Next, we will discuss the \textit{weight training} of sensors and formulas. For the weight of sensors, we aim to take the influence of the prediction confidence $z_i$ of sensor $t_i(\cdot)$ into account, and thus we assign $\log [z_i/(1-z_i)]$ to be the weight of sensor $t_i$. As a result, if there is no other formula, the marginal probability of the predicate $t_i$ to be true will be its corresponding prediction confidence $z_i$.
For other formulas, we will train the weights given observed variables based on variational EM steps (M-step) based on a trained GCN, and the details are illustrated in~\Cref{sec:mstep}.

\vspace{-3mm}
\section{Scalable Reasoning via Variational Inference Using GCN}
\label{sec:sector-gnn}
\vspace{-1mm}
In this section, we will illustrate in detail how to approximate the inference of our reasoning component MLN via variational inference based on GCN.

 \vspace{-2mm}
\subsection{Variational EM Based on GCN}
\vspace{-1mm}
In order to conduct efficient inference and learning for MLN, existing work has introduced different approaches, including variational inference and Monte Carlo sampling~\cite{domingos2019unifying,liu2014generalizable}.
In particular, as MLN models the joint probability distribution of all predicates as defined in Eq.~\ref{eq:joint_p}, it is possible to train the weights of knowledge rules (formulas) $w$ within MLN by maximizing the log-likelihood of all the observed predicates (facts) $\log P_w(\mathcal{O})$.
However, it is intractable to maximize the overall objective directly since it requires computing the whole partition function $Z(w)$ and integrating over all observed predicates $\mathcal{O}$ and unobserved ones $\mathcal{U}$. Thus, some works propose to instead optimize the variational evidence lower bound (ELBO)~\cite{zhang2019efficient} of the data log-likelihood as below:
\vspace{-1mm}
\begin{equation}
\label{eq:old_elbo}
\begin{aligned}
   \log P_{w}(\mathcal{O}) \ge \mathcal{L}_{\mathrm{ELBO}}(Q_{\theta}&, P_{w}) :=\mathbb{E}_{Q_{\theta}(\mathcal{U} | \mathcal{O})}\left[\log P_{w}(\mathcal{O}, \mathcal{U})\right] \\
   &- \mathbb{E}_{Q_{\theta}(\mathcal{U} | \mathcal{O})}\left[\log Q_{\theta}(\mathcal{U} | \mathcal{O})\right],
\end{aligned}
\vspace{-1mm}
\end{equation}
where $Q_{\theta}(\mathcal{U}|\mathcal{O})$ represents the variational posterior distribution, and the equality in Eq~\ref{eq:old_elbo} holds if $Q_{\theta}(\mathcal{U}|\mathcal{O})$ equals to the true posterior $P_w(\mathcal{U}| \mathcal{O})$. 
Here since the sensor output variables together with the knowledge rules among them can be represented as a knowledge graph, we will use graphical convolutional networks (GCN) to encode the posterior distribution $Q_\theta(\cdot)$.

Now we need to learn the weights of MLN $w$, based on which we will make the inference for MLN to enforce the reasoning process given the knowledge rules. Thus, we leverage a variational EM algorithm~\cite{ghahramani2000graphical} to optimize the ELBO in Eq.~\ref{eq:old_elbo}. In particular, since the input $x$ is unknown, all the variables such as $\texttt{IsPanda}(x)$ are unobserved. Thus, we  consider all variables to be unobserved by setting $\mathcal{O}$ to be empty set $\emptyset$, and optimize over the outputs of sensors  $\pT=\{t_1, t_2, ...,t_L \}$ with the following optimization objective:
\vspace{-5mm}
\begin{small}
\begin{equation}
\label{eq:new_elbo}
   \mathcal{L}_{\mathrm{ELBO}}\left(Q_{\theta}, P_{w}\right):=\mathbb{E}_{Q_{\theta}(\pT) }\left[\log P_{w}(\pT)\right]-\mathbb{E}_{Q_{\theta}(\pT )}\left[\log Q_{\theta}(\pT)\right],
   \vspace{-1mm}
\end{equation}
\end{small}
which is the negative KL divergence between $Q_{\theta}(\pT)$ and $P_w(\pT)$. On the other hand, it can also be viewed as to directly approximate $P_w(\pT)$ via a variational distribution $Q_{\theta}(\pT)$. 

Next, we will discuss in detail the EM steps. On the high level, in the  E-step, we will fix the MLN weights $w$ for the knowledge rules and optimize GCN parameters $Q_{\theta}$ to minimize the KL distance between $Q_{\theta}(\pT)$ and $P_w(\pT)$;  in the M-step, we will fix $Q_{\theta}$ and update the weights $w$ to maximize the the log-likelihood function $\mathbb{E}_{Q_{\theta}(\pT) }\left[\log P_{w}( \pT)\right]$. The E-step and the M-step will be executed alternately multiple times until convergence.

\vspace{-2mm}
\subsection{E-step: Optimizing $Q_\theta$}
\vspace{-1mm}

In E-step, we aim to minimize the KL distance between the variational distribution $Q_{\theta}(\pT)$ and the true distribution $P_w(\pT)$. Since the inference of the MLN is \textsf{\#P}-complete~\cite{richardson2006markov}, we approximate $P_w(\pT)$ with a mean-field distribution, which has been shown to scale up to large graphical models~\cite{qu2019gmnn,qu2019probabilistic,zhang2019efficient}. In the mean-field variational distribution where the variables are independent, the joint distribution of outputs of sensors (unobserved variables) can be formed as the following,
\vspace{-3mm}
\begin{equation}
\label{eq:mean-field}
   Q_{\theta}(\pT):= \prod\nolimits_{t_i \in \pT} Q_{\theta}(t_i),
\vspace{-2mm}
\end{equation}
We constrain the sum of the $Q_{\theta}(t_i)$, whose associated class confidence $z_i$ comes from the same main sensor (i.e., multi-class classifier), to be $1$, in order to model 
key-constraints, namely the exclusion rules, induced 
by a $d$-way classifier.

To further improve the efficiency of inference and take into account the knowledge graph structure, we parameterize the $Q_{\theta}$ here with graph convolutional networks~(GCNs), where $\theta$  represents the parameters of GCN. In particular, we will construct nodes for each class prediction based on both main and knowledge sensor outputs as shown in~\Cref{fig:pipeline} (c). In other words, each node will be associated with a scalar which represents the confidence of the corresponding class. However, the message passing between different nodes on the GCN will not be effective if the input feature is only a scalar, and thus it will be hard for the GCN to learn the variational posterior distribution well. So following~\cite{zhang2019efficient}, we also train a class embedding vector $\vec{{\mu_i}}$ for each node and use the scalar multiplication of the class prediction confidence {$z_i$} and the corresponding class embedding vector $\vec{{\mu_i}}$ as the input of each node for further encouraging the expressivity of the model.

Based on the mean-field approximation, and joint distribution $\log P_{w}(\pT) = \log (\frac{1}{Z(w)} \exp(\sum_{f \in \mathcal{F}} w_ff(t_1,...,t_L)) )$, the ELBO from Eq.~\ref{eq:new_elbo} can thus be rewritten as:
\vspace{-2mm}
\begin{small}
\begin{equation}
\begin{aligned}
\label{eq:reelbo}
\mathcal{L}_{\mathrm{ELBO}}\left(Q_{\theta}, P_{w}\right) = \mathbb{E}_{Q_{\theta}(\mathcal{P})} & \left(  \sum_{f \in \mathcal{F}} w_ff(t_1,...,t_L) - \log Z(w) \right) \\
&- \mathbb{E}_{Q_{\theta}(\pT)}\log Q_{\theta}(\pT) .
\end{aligned}
\vspace{-1mm}
\end{equation}
\end{small}

Since the MLN weights $w$ is fixed during the E-step, the $\log Z(w)$ here is a constant and can be ignored during the optimization. However, with this new optimization objective, we cannot obtain the gradient of it w.r.t. the parameters $\theta$ in GCN through backpropagation directly. Thus, we first derive the explicit form of the gradient as bellows, and the full proof is deferred to~\Cref{sec:lemmaproof}.

\vspace{-1mm}
\begin{lemma}
\label{le:gradientofelbo}
The gradient of $\mathcal{L}_{\mathrm{ELBO}}\left(Q_{\theta}, P_{w}\right)$ w.r.t. the GCN parameters $\theta$ can be derived as: 
\vspace{-3mm}
\begin{small}
\begin{equation}
\mathbb{E}_{Q_{\theta}(\pT)}\left(\sum_{f \in \mathcal{F}} w_ff(t_1,...,t_L) -\log Q_{\theta}(\pT)\right)\nabla_\theta\log Q_{\theta}(\pT).
\vspace{-2mm}
\end{equation}
\end{small}
\end{lemma}

This shows that the gradient can be estimated through multiple sampling from $Q_\theta(\pT)$. In specific, the term $\nabla_\theta\log Q_{\theta}(\pT)$ can be directly derived through backpropagation, and then the question remains as how to calculate $\sum_{f \in \mathcal{F}} f(t_1,...,t_L)$. For the single sensor (formula), it can be calculated by the dot multiplication of $\bm{t}$ and $\log[\bm{z}/(1-\bm{z})]$ where $\bm{t} = [t_1,...,t_L]$; for the exclusion rule based formula, as mentioned before, we can imply it by constraining the sum of $Q_{\theta}(t_i)$ to be $1$ for the corresponding classes. For the attribute rule and  hierarchy rule based formulas, 
they can be reduced to the combination of four basic kinds of formulas: 
\begin{equation}
    \label{eq:allowed_clause}  
    \begin{aligned}
     &t_i \implies t_j\lor t_k\lor...\lor t_l,\\
      &t_i \implies t_j\wedge t_k\wedge...\wedge t_l, \\
     &t_j\lor t_k\lor...\lor t_l\implies t_i,\\
      &t_j\wedge t_k\wedge...\wedge t_l\implies t_i.\\
    \end{aligned}
    \vspace{-2mm}
\end{equation}
We thus provide an efficient score calculation for them as follows, and the detailed proof is deferred to~\Cref{sec:thmproof}.

\vspace{-2mm}
\begin{theorem}
    \label{thm:1nnreasoning}
    The function $\sum_{f \in \mathcal{F}} f(t_1,...,t_L)$ based on the four types of formulas defined in~\Cref{eq:allowed_clause} can be efficiently calculated as follows:
    \begin{equation}
    \vspace{-2mm}
    \label{eq:matrix_form}
    \sum_{f \in \mathcal{F}} w_ff(t_1,...,t_L) =  \bm{w}\operatorname{Neg}(A\bm{t}^T+B),
    \end{equation}
    where $\bm{w}$ is the row vector of the concatenation of all $w_f$ for $f \in \gF$, $\operatorname{Neg}(\cdot)$ is an indicator function which maps the values larger than $0$ to $0$  and maps the other values to $1$, $A$ and $B$ are the matrices determined by the pre-defined formulas with the shape $|\gF|\times L$ and $|\gF|\times 1$, respectively.
    \vspace{-2mm}
    \end{theorem}

In practice, we will shift the term $\left(\sum_{f \in \mathcal{F}} w_ff(t_1,...,t_L) -\log Q_{\theta}(\mathcal{T})\right)$ by subtracting the sample mean for reducing the variance in the estimation for the gradient with Monte Carlo based on the fact that $\mathbb{E}_{Q_{\theta}(\mathcal{T})}\nabla_\theta\log Q_{\theta}(\mathcal{T})=0$.

 Since the tasks here are supervised, and there is label information for each input during training, we can add a supervised negative likelihood to encourage the overall learning and help guide the direction of the optimization:
 \vspace{-3mm}
\begin{equation}
\label{eq:old_label}
\mathcal{L}_{\text{label }}(Q_{\theta})= -\sum_{i=1}^L \log Q_{\theta}\left( \text{GT}(t_i)\right),
\vspace{-2mm}
\end{equation}
where $\text{GT}(t_i)$ is the corresponding ground truth for predicate $t_i$ during training.  Thus, the final E-step training optimization objective is:
\vspace{-3mm}
\begin{equation}
\label{eq:label}
\mathcal{L}_{\text{new}}(Q_{\theta}) =\mathcal{L}_{\text {ELBO }}(Q_{\theta}, P_{w})-\eta\mathcal{L}_{\text {label }}(Q_{\theta}),
\vspace{-2mm}
\end{equation}
where  $\eta$ is a hyperparameter to balance the trade-off of these two likelihood terms. The embedding $\vec{\mu}$ will also be updated during the optimization of GCN through the chain rule for better expressiveness. 

During the test stage, the prediction for each class will be based on the marginal probability $Q_{\theta}(t_i)$,  and it can be seen as a  knowledge-enhanced correction for the original prediction.

\begin{figure*}[th]
    \centering
    \includegraphics[width=\textwidth]{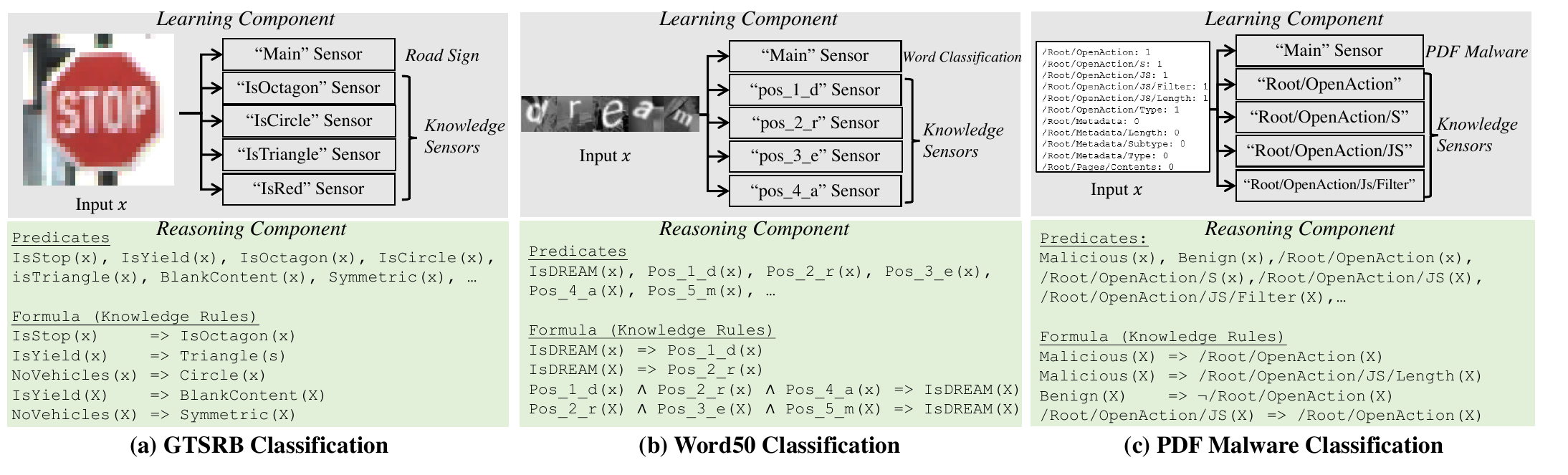}
    \vspace{-9mm}
    \caption{\small Learning and reasoning components of \name on GTSRB, Word50, and PDF malware classification. (AwA2 is illustrated in~\Cref{fig:pipeline}).}
    \label{fig:application}
    \vspace{-5mm}
    \end{figure*}

\setlength{\textfloatsep}{0pt}
\begin{algorithm}[t]
\caption{The whole training procedure for the variational EM based on GCN.}
\label{alg:pipeline}
\begin{algorithmic}[1]
\renewcommand{\algorithmicrequire}{\textbf{Input:}}
 \renewcommand{\algorithmicensure}{\textbf{Output:}}
 \REQUIRE Input $x$, a set of trained sensors (predicates) $\gT$, model GCN, sensor output confidence $\bm{z}=[z_1,...,z_L]$, number of training epochs $K$
 \ENSURE Trained GCN; the weight of MLN formulas $w$
  \STATE Initialize the node embedding of GCN
  $\bm{\mu} = [\vec{\mu_1},...,\vec{\mu_L}]$. 
   \STATE $\bm{m}=[\vec{m_1},...,\vec{m_L}] \gets [z_1\vec{\mu_1},...,z_1\vec{\mu_L}]$ \# Initialize  node features in GCN.
    \STATE $Q_\theta(\gT) \gets \text{GCN}(\bm{m};\theta)$ \# Get variational distribution.
  \FOR{$j = 1$ to $K$}
\STATE $\theta \gets \argmax_{\theta} \mathcal{L}_{\text{new}}\left(Q_{\theta}, P_{w}\right)$ \# E-step.
\STATE Update node embedding $\bm{\mu}$ from $Q_\theta (\gT)$
\STATE  $\bm{m}=[\vec{m_1},...,\vec{m_L}] \gets [z_1\vec{\mu_1},...,z_1\vec{\mu_L}]$ \# Update the input feature to GCN.
\STATE  $Q_\theta(\gT) \gets \text{GCN}(\bm{m};\theta)$ \# Update variational distribution.
\STATE $w \gets \argmax_{w} \mathbb{E}_{Q_{\theta}}\left[\log P_w\left(\mathcal{T}\right)\right]$ \# M-step.
 \ENDFOR
\STATE \textbf{return} GCN parameter $\theta$; weight of MLN formulas $w$.
 \end{algorithmic}
 \end{algorithm}

\vspace{-2mm} 
\subsection{M-step: Optimizing $w$}
\label{sec:mstep}
\vspace{-1mm}
In M-step, the GCN model is fixed, and we update the weight of the formula $w$ by maximizing the log-likelihood function, i.e., $\mathbb{E}_{Q_{\theta}(\mathcal{\pT})}\left[\log P_w\left(\mathcal{\pT}\right)\right]$, which is to maximize the term 
\begin{equation}
\label{eq:loglikelihood}
\mathbb{E}_{Q_{\theta}(\mathcal{T})} \log \frac{\exp\{\sum_{f \in \mathcal{F}} w_f f(t_1,...,t_L)\}}{\sum_{t'_1,...t'_L} \exp\{\sum_{f \in \mathcal{F}} w_ff(t_1',...,t_L')\}}.
\end{equation}
However, the partition function, namely, the denominator that involves an integration of all the variables, is intractable to compute. We  optimize the pseudo-likelihood~\cite{besag1975statistical} as an alternative, which is defined as:
\vspace{-3mm}
\begin{equation}
\begin{aligned}
P_{w}^*(t_1,..., t_L) := \prod_{i=1}^{L} P_{w}\left(t_{i} | MB\left(t_{i}\right)\right), 
\end{aligned}
\end{equation}
where $\mathrm{MB}(t_i)$ is the Markov blanket of the predicate $t_i$. In other words,  $\mathrm{MB}(t_i)$ is the set of formulas where the predicate $t_i$ appears. Then, following~\cite{richardson2006markov}, given a formula $f$, the gradient of the pseudo-log-likelihood w.r.t. its associated weight $w$, namely $\frac{\partial}{\partial w} \log P_{w}^{*}(t_1,...,t_L) $, is
\vspace{-2mm}
\begin{equation}
\label{eq:weght_update}
\begin{aligned}
 \sum_{i=1}^{L}  \left[f(\right.& t_1,...,t_L) -P_{w}\left(t_{i}=0 | MB\left(t_{i}\right)\right) f\left(\left[t_{i}=0\right]\right) \\
&\quad \left.-P_{w}\left(t_{i}=1 | MB\left(t_{i}\right)\right) f\left(\left[t_{i}=1\right]\right)\right],
\end{aligned}
\vspace{-2mm}
\end{equation}
where $f\left(\left[t_{i}=0\right]\right)$ represents the truth value of the formula $f$ when we force $t_i = 0$ while leaving the remaining $t_{j, j\neq i}$ unchanged;  similar for $f\left([t_{i}=1\right])$.  
Finally, we will maximize the original intractable log-likelihood function through optimizing the expectation of the pseudo-log-likelihood $\mathbb{E}_{Q_{\theta}(\mathcal{T})}\left[\log P_w^*\left(\mathcal{T}\right)\right]$, and the gradient w.r.t. the weight $w$ of the formula will be estimated through multiple sampling from the variational distribution $Q_{\theta}$. The algorithm for the whole training pipeline is provided in~\Cref{alg:pipeline}.


\vspace{-1mm}
\section{Experimental Evaluation}
\label{sec:exp}
\begin{figure}[t]
    \centering
    \includegraphics[width=0.4\textwidth]{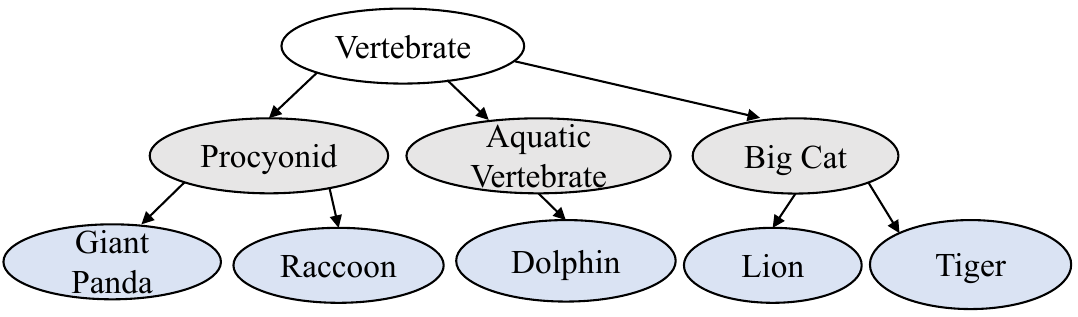}
    \vspace{-2mm}
    \caption{\small Hierarchy knowledge rule of AwA2. The blue nodes represent the main task of animal categories, and the grey nodes represent part of the knowledge sensors.}
    \label{fig:awa_hierarchy}
    \vspace{-1mm}
    \end{figure}

\vspace{-1mm}
In this section, we present experimental evaluation of \name on four large  datasets: Animals with Attributes~(AwA2)~\cite{xian2018zero}, Word50~\cite{chen2015learning}, GTSRB~\cite{stallkamp2012man}, and PDF malware dataset from Contagio~\cite{parkour201716}. 
The illustration of these datasets and the corresponding construction of \name are shown in~\Cref{fig:pipeline,fig:application,fig:awa_hierarchy,fig:pdfmalware}.
With the knowledge integration and reasoning, \name achieves significantly higher certified robustness than the state-of-the-art methods under different radii. We also conduct a set of ablation studies to explore the influence of the number of integrated knowledge rules, the empirical robustness of \name, and the explanation properties of \name via case studies. All experiments are run on four GeForce RTX 2080 Ti GPUs.

    
\renewcommand\arraystretch{1}
\begin{table*}[t]
    \centering
    \caption{\small  Certified accuracy under different $\ell_2$ perturbation radii on AwA2 dataset.}
    \label{tab:awa}
    \vspace{-2mm}
\resizebox{0.9\linewidth}{!}{
\begin{tabular}{c|c|cccccccccccccc}
\hline
\multirow{2}{*}{$\sigma$} & \multirow{2}{*}{Method} & \multirow{2}{*}{ACR} & \multicolumn{12}{c}{Certified Accuracy under Radius $r$} &      \\
 &                         &                      & 0.00 & 0.20 & 0.40 & 0.60 & 0.80 & 1.00 & 1.20 & 1.40 & 1.60 & 1.80 & 2.00 & 2.20 & 2.40 \\ \hline\hline
\multirow{6}{*}{0.25} & Gaussian                & 0.544                                    & 84.0                     & 77.6                     & 71.4                     & 58.6                     & 40.0                     & 0.0                      & 0.0                      & 0.0                      & 0.0                      & 0.0                      & 0.0                      & 0.0                      & 0.0                      \\
& SWEEN                & 0.552                                    & 84.2                     & 78.8                     & 71.2                     & 60.8                     & 43.0                     & 0.0                      & 0.0                      & 0.0                      & 0.0                      & 0.0                      & 0.0                      & 0.0                      & 0.0                      \\
 & SmoothAdv               & 0.574 & 78.6 & 74.8 & 71.6 & 69.4 & 62.2 & 0.0 & 0.0 & 0.0 & 0.0 & 0.0 & 0.0 & 0.0 & 0.0             \\
& Consistency             & 0.587 & 81.6 & 78.2 & 74.0 & 69.8 & 58.2 & 0.0 & 0.0 & 0.0 & 0.0 & 0.0 & 0.0 & 0.0 & 0.0                      \\
& {MultiTask}             & {0.593} & {79.8} & {78.2} & {76.2} & {71.0} & {58.0} & {0.0} & {0.0} & {0.0} & {0.0} & {0.0} & {0.0} & {0.0} & {0.0} \\
 & \name    & \textbf{0.709} & \textbf{96.6} & \textbf{94.2} & \textbf{91.4} & \textbf{85.4} & \textbf{75.0} & 0.0 & 0.0 & 0.0 & 0.0 & 0.0 & 0.0 & 0.0 & 0.0     \\ \hline\hline
\multirow{6}{*}{0.50}                  & Gaussian                & 0.827                                    & 75.6                     & 71.2                     & 64.6                     & 58.2                     & 53.0                     & 46.2                     & 38.8                     & 32.0                     & 21.2                     & 0.0                      & 0.0                      & 0.0                      & 0.0                      \\
 & SWEEN                & 0.854                                    & 76.4                     & 73.8                     & 67.8                     & 60.4                     & 53.6                     & 47.4                     & 39.6                     & 34.6                     & 22.4                     & 0.0                      & 0.0                      & 0.0                      & 0.0                      \\
  & SmoothAdv               & 0.949 & 72.0 & 69.8 & 66.6 & 62.8 & 60.2 & 56.8 & 52.2 & 47.6 & 40.2 & 0.0 & 0.0 & 0.0 & 0.0    \\
 & Consistency             & 0.953 & 74.0 & 71.2 & 68.8 & 64.6 & 61.2 & 56.0 & 51.2 & 46.8 & 40.4 & 0.0 & 0.0 & 0.0 & 0.0                \\
 & {MultiTask}             & {0.842} & {69.6} & {67.6} & {63.2} & {58.2} & {53.4} & {49.4} & {42.4} & {36.8} & {27.2} & {0.0} & {0.0} & {0.0} & {0.0}\\
  & \name    & \textbf{1.141} & \textbf{91.2} & \textbf{88.2} & \textbf{84.2} & \textbf{78.8} & \textbf{73.4} & \textbf{68.4} & \textbf{63.2} & \textbf{56.2} & \textbf{44.0} & 0.0 & 0.0 & 0.0 & 0.0                   \\
\hline\hline
\multirow{6}{*}{1.00}                  & Gaussian                & 0.994                                    & 59.6                     & 54.6                     & 51.6                     & 49.0                     & 44.8                     & 40.8                     & 36.6                     & 32.6                     & 29.6                     & 26.4                     & 22.8                     & 20.0                     & 17.2                     \\
  & SWEEN                & 1.059                                    & 62.2                     & 57.6                     & 54.8                     & 50.2                     & 45.8                     & 41.8                     & 39.2                     & 34.4                     & 32.0                     & 29.0                     & 26.8                     & 22.0                     & 18.8                     \\
   & SmoothAdv               & 1.231 & 57.2 & 54.0 & 53.0 & 49.8 & 47.2 & 45.4 & 42.2 & 40.8 & 38.2 & 36.8 & 34.0 & 32.6 & 30.2                   \\
 & Consistency             & 1.247 & 54.0 & 52.0 & 50.0 & 48.0 & 45.6 & 44.0 & 42.0 & 40.6 & 39.4 & 37.8 & 36.0 & 33.8 & 31.6              \\
  & {MultiTask}             & {1.192} & {51.6} & {49.8} & {48.4} & {46.8} & {46.0} & {45.0} & {42.0} & {40.0} & {38.2} & {36.0} & {34.0} & {31.2} & {29.2}\\
& \name   & \textbf{2.127} & \textbf{87.0} & \textbf{85.2} & \textbf{84.0} & \textbf{82.0} & \textbf{80.4} & \textbf{78.2} & \textbf{75.6} & \textbf{71.4} & \textbf{68.6} & \textbf{65.8} & \textbf{61.8} & \textbf{59.4} & \textbf{56.0}       \\ \hline
\end{tabular}}
\vspace{-6mm}
\end{table*}

\vspace{-2mm}
\subsection{Experimental Setup}
\label{subsec:exp-setup}
\vspace{-1mm}
\textbf{Datasets and the implementation of \textit{learning} component.}
 For AwA2, all sensors, including the main sensor and knowledge sensors, are trained with the architecture of ResNet-50~\cite{he2016deep}; while for Word50 and PDF malware datasets, we build the feed-forward neural network with two hidden layers activated by ReLU for all sensors. Specifically, for Word50, following the similar setting in~\cite{chen2015learning}, the number of hidden neurons is set to $512$ for character classification and  $1024$ for word classification; for the PDF malware dataset, following the same setting in~\cite{chen2020training}, the number of hidden neurons is set to $200$ for both main sensor and the knowledge sensors. For GTSRB, we use the GTSRB-CNN~\cite{eykholt2018robust} for all sensors.
    
\vspace{-1mm}
\textbf{The implementation of \textit{reasoning} component.} The dimension of the embedding $\vec{\mu}$ for each predicate is fixed to $512$, {and it is initialized with the He uniform initialization~\cite{he2015delving}}. For all datasets, we use the GCN with two hidden layers, and the hidden dimension is also set to $512$. For the construction of the graph, we introduce a node for each predicate, and each predicate corresponds to one class that appeared in the main sensor or knowledge sensor exactly. The edge will connect associated predicates that appeared in the same knowledge rule (formula). 
We train the GCN with $60$ epochs, and the learning rate is set to $0.01$ in the first $40$ epochs and set to $0.001$ for the last $20$ epochs. 

\vspace{-1mm}
\textbf{Baselines.}
For the dataset AwA2, Word50, and GTSRB, whose features are continuous, we consider four state-of-the-art $\ell_2$ certification baselines: (1) \emph{Gaussian smoothing}~\cite{cohen2019certified} trains smoothed classifiers by directly augmenting the input images with the Gaussian noise during training; (2) \emph{SWEEN}~\cite{liu2020enhancing} employs weighted ensemble to improve the certified robustness; (3) \emph{SmoothAdv}~\cite{salman2019provably} further applies adversarial training to improve the certified robustness based on the Gaussian smoothing; (4) \emph{Consistency}~\cite{jeong2020consistency} adds a consistency regularization term in the training loss based on standard Gaussian smooth training; {(5)~\emph{MultiTask} incorporates the additional sensor information by adding more classification heads in the main sensor and training it together with the knowledge classification tasks under Gaussian augmentation}. For the PDF malware dataset, whose features are binary, we consider the state-of-the-art $\ell_0$ certification baseline: Lee et al.~\cite{lee2019tight}. 
To train a $\ell_0$ smoothed robust model, we smooth each feature by replacing it with a range of discrete random values with probability ($1-\alpha$) following existing work~\cite{lee2019tight}.
SWEEN serves as another baseline to represent the SOTA robust ensembles.

\vspace{-1mm}
\textbf{Evaluation metrics.}
We mainly focus on the \emph{certified robustness} of different methods. 
For AwA2, Word50, and GTSRB, we report the certified accuracy of \name and other pipelines under different $\ell_2$ radius $r$, following the standard certification setting~\cite{cohen2019certified}. For the PDF malware dataset, we will report the certified accuracy under different $\ell_0$ radius $r$, which follows the certification setting in~\cite{lee2019tight}. In addition, we also report the average certified radius (ACR) following~\cite{zhai2019macer}.
{The whole certification procedure is provided in~\Cref{sec:certification_alg}.}

\renewcommand\arraystretch{1}
\begin{table*}[htbp]
    \centering
    \caption{\small Certified accuracy under different $\ell_2$ perturbation radii on word classification in Word50.}
    \label{tab:word}
    \vspace{-3mm}
\resizebox{0.9\linewidth}{!}{
\begin{tabular}{c|c|cccccccccccccc}
\hline
\multirow{2}{*}{$\sigma$} & \multirow{2}{*}{Method} & \multicolumn{1}{c}{\multirow{2}{*}{ACR}} & \multicolumn{12}{c}{Certified Accuracy under Radius $r$}      & \multicolumn{1}{c}{}     \\ &     & \multicolumn{1}{c}{}    & \multicolumn{1}{c}{0.00} & \multicolumn{1}{c}{0.10} & \multicolumn{1}{c}{0.20} & \multicolumn{1}{c}{0.30} & \multicolumn{1}{c}{0.40} & \multicolumn{1}{c}{0.50} & \multicolumn{1}{c}{0.60} & \multicolumn{1}{c}{0.70} & \multicolumn{1}{c}{0.80} & \multicolumn{1}{c}{0.90} & \multicolumn{1}{c}{1.00} & \multicolumn{1}{c}{1.10} & \multicolumn{1}{c}{1.20} \\ \hline\hline
\multirow{6}{*}{0.12}                  & Gaussian                & 0.115 & 48.6 & 38.0 & 26.4 & 16.6 & 10.0 & 0.0 & 0.0 & 0.0 & 0.0 & 0.0 & 0.0 & 0.0 & 0.0     \\
& SWEEN                & 0.152                              & 51.4                & 44.6                & 36.6                & 27.4                & 18.4                & 0.0                  & 0.0                  & 0.0                  & 0.0                  & 0.0                  & 0.0                  & 0.0                  & 0.0                  \\
& SmoothAdv               & 0.197 & 59.0 & 53.8 & 45.2 & 38.0 & 29.0 & 0.0 & 0.0 & 0.0 & 0.0 & 0.0 & 0.0 & 0.0 & 0.0                 \\
 & Consistency             & 0.157 & 53.4 & 45.2 & 36.0 & 28.4 & 20.8 & 0.0 & 0.0 & 0.0 & 0.0 & 0.0 & 0.0 & 0.0 & 0.0                \\
  & {MultiTask}             & {0.142} & {55.6} & {43.0} & {32.6} & {23.6} & {15.6} & {0.0} & {0.0} & {0.0} & {0.0} & {0.0} & {0.0} & {0.0} & {0.0}  \\
 & \name    & \textbf{0.391} & \textbf{97.0} & \textbf{96.0} & \textbf{91.4} & \textbf{81.4} & \textbf{70.4} & 0.0 & 0.0 & 0.0 & 0.0 & 0.0 & 0.0 & 0.0 & 0.0                 \\ \hline\hline
\multirow{6}{*}{0.25}                  & Gaussian                & 0.125 & 42.0 & 32.6 & 24.0 & 17.6 & 13.0 & 8.0 & 4.6 & 2.8 & 2.2 & 0.6 & 0.0 & 0.0 & 0.0          \\
& SWEEN                & 0.194                             & 48.2                & 41.8                & 35.6               & 28.4                & 22.0                & 16.6                & 10.8                & 7.6                 & 4.8                 & 2.4                  & 0.0                  & 0.0                  & 0.0                  \\
& SmoothAdv               & 0.246 & 55.6 & 47.0 & 40.0 & 35.8 & 29.0 & 24.8 & 17.6 & 11.8 & 8.4 & 4.8 & 0.0 & 0.0 & 0.0           \\
& Consistency             & 0.201 & 47.6 & 40.2 & 36.0 & 29.0 & 24.0 & 17.4 & 13.2 & 9.0 & 6.6 & 2.4 & 0.0 & 0.0 & 0.0                  \\
  & {MultiTask} & {0.166} & {49.2} & {40.8} & {34.2} & {25.6} & {17.4} & {11.8} & {7.2} & {4.2} & {2.0} & {1.2} & {0.0} & {0.0} & {0.0} \\
& \name    & \textbf{0.674} &\textbf{97.2} & \textbf{94.8} &\textbf{92.6} & \textbf{89.4 }& \textbf{81.8} & \textbf{73.6} & \textbf{64.4} & \textbf{55.2} & \textbf{43.6} & \textbf{30.8} & 0.0 & 0.0 & 0.0      \\ \hline\hline
\multirow{6}{*}{0.50}                  & Gaussian                & 0.082 & 27.8 & 20.4 & 14.6 & 11.2 & 7.8 & 4.4 & 3.4 & 2.2 & 1.6 & 1.0 & 0.4 & 0.4 & 0.4              \\
& SWEEN                & 0.143                              & 35.0                & 29.0                & 23.8                & 17.8                & 14.2                & 11.0                & 8.6                 & 6.4                  & 4.4                  & 3.8                  & 2.0                  & 1.4                  & 0.8                  \\ 
 & SmoothAdv               & 0.168 & 38.0 & 32.6 & 27.0 & 21.0 & 16.0 & 12.6 & 9.4 & 7.4 & 6.0 & 5.4 & 3.8 & 2.8 & 1.6        \\
 & Consistency             & 0.146 & 34.8 & 28.4 & 22.8 & 17.8 & 13.0 & 9.8 & 8.8 & 7.8 & 5.8 & 4.6 & 3.2 & 2.4 & 1.6                \\
   & {MultiTask} & {0.108} & {32.0} & {24.8} & {18.6} & {13.4} & {9.8} & {8.2} & {5.2} & {4.2} & {2.8} & {1.8} & {1.2} & {0.4} & {0.4}  \\
& \name    & \textbf{0.697} & \textbf{87.6}& \textbf{84.4} & \textbf{78.4} & \textbf{73.6} & \textbf{69.0} & \textbf{63.0} & \textbf{56.6} & \textbf{50.0} & \textbf{44.0} & \textbf{36.6} & \textbf{30.0}& \textbf{24.0} & \textbf{18.0}      \\
\hline
\end{tabular}}
\vspace{-5mm}
\end{table*}

\renewcommand\arraystretch{1}
\begin{table*}[htbp]
    \centering
    \caption{\small Certified  accuracy under different $\ell_2$ perturbation radii  on character classification in Word50.}
    \label{tab:character}
    \vspace{-3mm}
\resizebox{0.9\linewidth}{!}{
\begin{tabular}{c|c|cccccccccccccc}
\hline
\multirow{2}{*}{$\sigma$} & \multirow{2}{*}{Method} & \multicolumn{1}{c}{\multirow{2}{*}{ACR}} & \multicolumn{12}{c}{Certified Accuracy under Radius $r$}      & \multicolumn{1}{c}{}     \\ &     & \multicolumn{1}{c}{}    & \multicolumn{1}{c}{0.00} & \multicolumn{1}{c}{0.10} & \multicolumn{1}{c}{0.20} & \multicolumn{1}{c}{0.30} & \multicolumn{1}{c}{0.40} & \multicolumn{1}{c}{0.50} & \multicolumn{1}{c}{0.60} & \multicolumn{1}{c}{0.70} & \multicolumn{1}{c}{0.80} & \multicolumn{1}{c}{0.90} & \multicolumn{1}{c}{1.00} & \multicolumn{1}{c}{1.10} & \multicolumn{1}{c}{1.20} \\ \hline\hline
\multirow{6}{*}{0.12}                  & Gaussian                & 0.234 & 72.0 & 64.0 & 53.8 & 45.8 & 34.8 & 0.0 & 0.0 & 0.0 & 0.0 & 0.0 & 0.0 & 0.0 & 0.0           \\
& SWEEN                & 0.250                              & 72.8                & 64.2                & 58.6                & 50.2                & 39.8                & 0.0                  & 0.0                  & 0.0                  & 0.0                  & 0.0                  & 0.0                  & 0.0                  & 0.0                  \\
& SmoothAdv               & 0.228 & 63.8 & 59.0 & 52.0 & 45.4 & 37.8 & 0.0 & 0.0 & 0.0 & 0.0 & 0.0 & 0.0 & 0.0 & 0.0      \\
 & Consistency             & 0.226 & 66.2 & 59.0 & 51.6 & 44.4 & 37.6 & 0.0 & 0.0 & 0.0 & 0.0 & 0.0 & 0.0 & 0.0 & 0.0            \\
& {MultiTask} & {0.191} & {62.8} & {53.6} & {45.2} & {35.4} & {25.8} & {0.0} & {0.0} & {0.0} & {0.0} & {0.0} & {0.0} & {0.0} & {0.0} \\
 & \name    & \textbf{0.341} & \textbf{90.2} & \textbf{85.2} & \textbf{78.0} & \textbf{70.8} & \textbf{60.0} & 0.0 & 0.0 & 0.0 & 0.0 & 0.0 & 0.0 & 0.0 & 0.0          \\ \hline\hline
\multirow{6}{*}{0.25}                  & Gaussian                & 0.290 & 63.4 & 57.2 & 51.0 & 42.2 & 35.0 & 26.8 & 20.2 & 13.4 & 9.0 & 4.2 & 0.0 & 0.0 & 0.0               \\
& SWEEN                & 0.315                              & 65.8                & 58.6                & 52.4                & 46.0                & 37.8                & 30.0                & 23.6                & 16.6                 & 10.8                 & 7.2                  & 0.0                  & 0.0                  & 0.0                  \\
& SmoothAdv               & 0.289 & 55.8 & 50.0 & 44.6 & 37.6 & 33.4 & 27.4 & 23.6 & 19.8 & 15.0 & 10.0 & 0.0 & 0.0 & 0.0            \\
& Consistency             & 0.285 & 57.0 & 52.2 & 46.2 & 39.6 & 35.2 & 27.8 & 23.8 & 16.6 & 11.0 & 5.6 & 0.0 & 0.0 & 0.0           \\
& {MultiTask} & {0.246} & {60.2} & {53.2} & {44.6} & {35.8} & {28.8} & {21.8} & {15.2} & {8.8} & {5.4} & {2.4} & {0.0} & {0.0} & {0.0}\\
& \name    & \textbf{0.539} & \textbf{87.6} & \textbf{83.2} & \textbf{77.4} & \textbf{73.0} & \textbf{63.2} & \textbf{55.6} & \textbf{50.2} & \textbf{41.8}& \textbf{32.4} & \textbf{21.2} & 0.0 & 0.0 & 0.0       \\\hline\hline
\multirow{6}{*}{0.50} & Gaussian & 0.165 & 40.0 & 33.6 & 29.8 & 23.6 & 18.2 & 13.2 & 9.2 & 7.2 & 4.8 & 2.8 & 1.8 & 0.6 & 0.6               \\
& SWEEN                & 0.184                              & 42.0                & 36.8                & 31.4                & 25.2                & 20.4                & 14.8                & 11.4                 & 8.4                  & 5.2                  & 3.6                  & 2.8                  & 1.2                  & 0.8                  \\
 & SmoothAdv               & 0.165 & 31.2 & 27.0 & 23.6 & 20.8 & 16.8 & 14.6 & 11.6 & 9.8 & 7.4 & 5.0 & 4.6 & 2.8 & 1.2            \\
 & Consistency             & 0.162 & 37.0 & 32.4 & 27.0 & 21.2 & 17.0 & 14.2 & 10.0 & 6.8 & 4.8 & 3.0 & 2.2 & 1.0 & 0.8   \\
 & {MultiTask} & {0.209} & {47.6} & {42.2} & {35.0} & {27.4} & {23.0} & {17.0} & {13.2} & {9.6} & {6.4} & {4.4} & {3.0} & {1.6} & {1.6}\\
& \name    & \textbf{0.539} & \textbf{80.6} & \textbf{76.8} & \textbf{70.4} & \textbf{65.4} & \textbf{59.2} & \textbf{53.0} & \textbf{44.8} & \textbf{37.0} & \textbf{29.6} & \textbf{21.6} & \textbf{17.2} & \textbf{11.6} & \textbf{6.6}        \\ \hline
\end{tabular}}
\vspace{-7mm}
\end{table*}

\vspace{-2mm}
\subsection{Evaluation on AwA2}
\vspace{-1mm}

\textbf{Dataset description.} AwA2~\cite{xian2018zero} contains $37322$ images of $50$ animal categories and provides $85$ class attributes for each class. For example, animal \emph{``Fox''} is assigned with the attribute labels such as \emph{``brown''}, \emph{``has a tail''} and \emph{``has no spots''}.
In addition to the attribute knowledge, we also construct hierarchical relations between the classes based on WordNet~\cite{miller1995wordnet} as another type of domain knowledge.

    \textbf{Task and the implementation of \textit{learning} component.}
    The main task here is to classify the $50$ animal classes for the input image. First, we will train one main sensor for classifying these $50$ animals and train $85$ knowledge sensors for classifying each binary attribute, respectively. We utilize  WordNet~\cite{miller1995wordnet} to build a hierarchy tree by iteratively searching the inherited hypernyms of the $50$ leaf animal classes, part of the nodes are shown in~\Cref{fig:awa_hierarchy}. Then, we perform additional hierarchy classification tasks on the $28$ internal nodes (gray nodes in~\Cref{fig:awa_hierarchy}). We conduct additional ablation studies to evaluate the effect of training different numbers of knowledge sensors  in~\Cref{sec:ablation}. The final sensing vector $\bm{z}$ is of $50+85+28=163$ dimensions by concatenating the output confidence of all the sensors (predicates).

    \textbf{The implementation of \textit{reasoning} component.}
    For each possible sensor task, we introduce a corresponding predicate. Thus, the number of the predicates is $163$ here.
    
    For the \textit{attribute-based} knowledge/formulas, we let each class imply its owned attributes. For example, if the animal class~\emph{``raccoon''} has the attributes~\emph{gray} and~\emph{furry}, we will  construct two formulas: 
    $\texttt{IsRaccoon}(x) \implies \texttt{IsGray}(x)$ and $\texttt{IsRaccoon}(x) \implies \texttt{IsFurry}(x)$. The total number of attribute-based formulas is the number of possible attributes of all animals, which is $1562$.
    
    For the \textit{hierarchy-based} knowledge/formulas, the classes in the internal nodes (the gray node as shown in~\Cref{fig:awa_hierarchy}) will imply at least one of its children to be true. For example, $\texttt{IsProcyonid}(x)  \implies \texttt{IsPanda}(x) \lor \texttt{IsRaccoon}(x)$; $\texttt{IsBigCat}(x) \implies \texttt{IsLion}(x) \lor \texttt{IsTiger}(x)$. The number of hierarchy-based formulas is the number of the internal nodes, which is $28$, and thus the total number of the overall formulas is $1562+28=1590$.
    
    
    \textbf{Certification details.}
    All the images are resized to $224\times224$ for training the sensors. We randomly sample $80\%$ images from each animal class as the training data while picking $10$ images from each class within the remaining unsampled set for certification. Following the standard setting~\cite{cohen2019certified}, we certify these $500$ images with confidence $99.9\%$ (the results are certified with  $N=10,000$ samples of smoothing noise). We test all methods based on three levels of smoothing noise  $\sigma = 0.25,0.50,1.00$. For small $\sigma =0.25$, the $\eta$ in~\Cref{eq:label} is set to $0.2$, and for $\sigma = 0.50$ and $1.00$, the $\eta$ is set to $0.6$. Generally, with larger training noise, the $\eta$ needs to be larger to help maintain the benign accuracy and guide the training of  GCN. The details for other baselines are deferred to~\Cref{sec:exp_baseline}.
    
    \textbf{Certification results.}
    The certification results of \name and baselines are shown in~\Cref{tab:awa}, and as we can see, our method \name improves the certified accuracy under different radii with different smoothing levels. 
    In addition, we can also replace the main sensor of \name with different training methods, and detailed results are in~\Cref{sec:exp_for_different_main}. {Out of interest, we also explore the importance of the reasoning module by comparing it with the method that replaces the GCN in~\name with a simple linear classifier while maintaining others as the same, the corresponding results are provided in~\Cref{adx:n-way}}.

    
\vspace{-2mm}
\subsection{Evaluation on Word50}
\vspace{-1mm}
    \textbf{Dataset description.}
    Word50~\cite{chen2015learning} is created by randomly choosing $50$ words, and each consists of $5$ lower case characters, which are extracted from the Chars74K dataset~\cite{de2009character}. The background of each character is inserted with random patches, and the whole character image is further perturbed with scaling, rotation, and translation to increase the difficulty of recognition, making it more challenging than traditional digit recognition tasks. The size of each character image is $28\times28$, and examples of the word images are shown in~\Cref{fig:application} (b). The intriguing property of this dataset is that the relationship between nearby characters can be treated as prior knowledge to help build reliable predictions. 
    
    \textbf{Task and the implementation of \textit{learning} component.}
    We conduct two tasks here, one is for the~\emph{word} classification, and one is for the~\emph{character} classification. We train one main sensor for classifying the $50$ words and $5$ knowledge sensors for classifying the character on each position of the word. The sensing vector $\bm{z}$  can be represented as $[\bm{u};\bm{e}_1;...;\bm{e}_5]$, where $\bm{u} \in \{0,1\}^{50}$ is the output confidence of the main sensor with $50$ dimensions, $\bm{e}_i \in \{0,1\}^{26}$ is the output confidence of the knowledge sensor which classifies the character on the $i$-th position with $26$ dimensions. Therefore, the total dimension of the sensing vector $\bm{z}$ here is $50+26\times5=180$. 
    
    \textbf{The implementation of \textit{reasoning} component.}
   For each word prediction, we will use a predicate to represent it. For instance, we will use $\texttt{DREAM}(x)$ to denote if the input word is $\texttt{DREAM}$. While for the character appeared in the word, we will use $\texttt{Pos\_i\_\$}(x)$  to represent if the character appeared in the $i$-th position of the input image $x$ is character `$\$$'. For example, the predicate $\texttt{Pos\_2\_a}(x)$ predicts if the second character appeared in the input word image is $\texttt{`a'}$. Thus, the number of the predicates is  $180$.
   
    Given the $50$ known words, we will construct knowledge rules based on the word and the corresponding characters in each position. For the attribute-based knowledge/formulas, we will build  them like $\texttt{IsDREAM}(x) \implies \texttt{Pos\_1\_d}(x)$, ..., $\texttt{DREAM}(x) \implies \texttt{Pos\_5\_m}(x)$, and the total number of such formulas is $50\times5=250$. In addition, for this dataset, we find that the identification of the characters on at least three positions is enough to determine the  whole word, thus we also construct the knowledge/formulas like $\texttt{Pos\_1\_d}(x) \land \texttt{Pos\_4\_a}(x) \land \texttt{Pos\_5\_m}(x) \implies \texttt{IsDREAM}(x)$ for further enriching the prediction robustness. The number of such  formulas is $50\times
   (\tbinom{5}{3}+\tbinom{5}{4}+\tbinom{5}{5})=800$, and thus the number of the overall formulas is $250+800=1050$.

    \textbf{Certification results.} The certified accuracy  on  word-level and character-level classifications are shown in \Cref{tab:word} and~\Cref{tab:character}, respectively. 
    As we can see, ~\name significantly outperforms all other baselines under different perturbation radii and smoothing noise levels. Using different models for the main sensor in \name can be found in~\Cref{sec:exp_for_different_main}.  The training and the certification details are deferred to~\Cref{sec:exp_baseline} and~\Cref{sec:certification_detail}, respectively.

\begin{figure*}
    \centering
    \includegraphics[width=\textwidth]{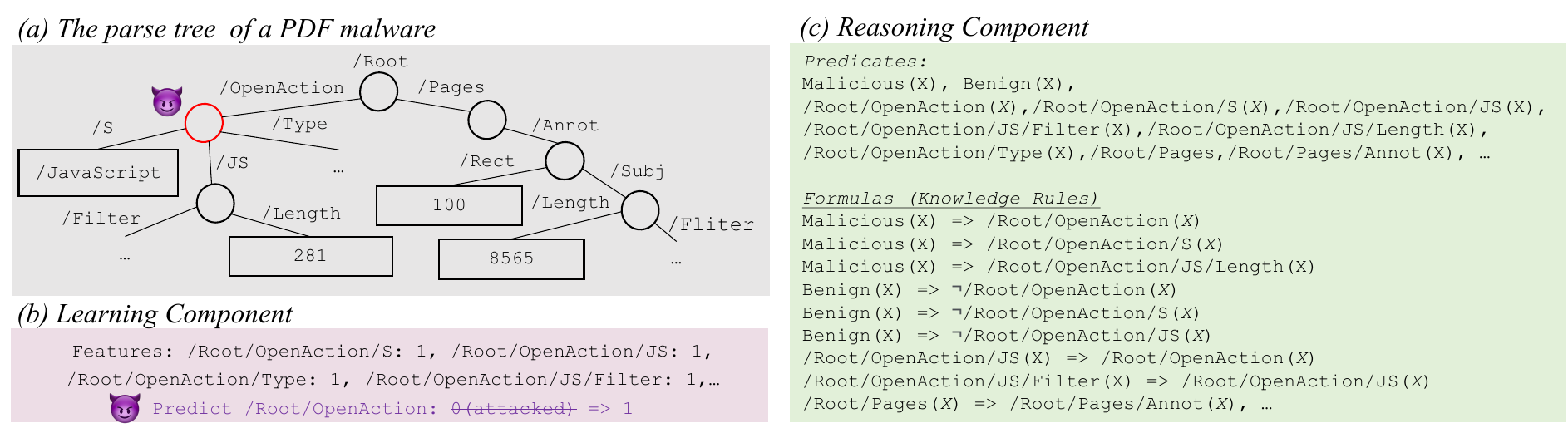}
    \vspace{-8mm}
    \caption{(a) The parsed tree structure of a PDF malware and an attack example; (b) the learning component of \name under attack; (c) the reasoning component of \name and some examples of corresponding knowledge rules.
    }
    \label{fig:pdfmalware}
    \vspace{-8mm}
\end{figure*} 

\vspace{-2mm}
\subsection{Evaluation on GTSRB}
\vspace{-1mm}
    \textbf{Dataset description.}  
    Here we use the GTSRB  dataset~\cite{stallkamp2012man} following~\cite{gurel2021knowledge},  which contains $12$ types of the road signs: \emph{``Stop'', ``Priority Road'', ``Yield'', ``Construction Area'', ``Keep Right'', ``Turn Left'', ``Do not Enter'', ``No Vehicles'', ``Speed Limit 20'', ``Speed Limit 50'', ``Speed Limit 120'', ``End of Previous Limitation''}. Each image is resized to $32\times32$ for training, and an example is shown in~\Cref{fig:application} (a). 
    
    \vspace{-1mm}
    \textbf{Task and the implementation of \textit{learning} component.}
    The main task is to classify the $12$ types of German road signs. For the learning component, first, we will train a main sensor to classify those $12$ road signs. Next, we manually construct $20$ knowledge sensors such as \texttt{IsOctagon()}, \texttt{IsSqaure()}, and \texttt{IsCircle()} based on the border patterns and the contents of the road signs. The full knowledge construction is provided in~\Cref{sec:gtsrb_details}. The final sensing vector $\bm{z}$ is of $12+20=32$ dimensions by concatenating the output confidence of all the sensors.
    
    \renewcommand\arraystretch{1.1}
    \begin{table}[t!]
        \centering
        \caption{\small Certified accuracy on GTSRB under $\ell_2$ radii.}
        \label{tab:stop_sign}
        \vspace{-2mm}
    \resizebox{\linewidth}{!}{
    \begin{tabular}{c|ccccccccccccc}
    \hline
     \multirow{2}{*}{Method} & \multicolumn{12}{c}{Certified Accuracy under $\ell_2$ Radius $r$}      & \multicolumn{1}{c}{}     \\  & \multicolumn{1}{c}{0.00}   & \multicolumn{1}{c}{0.10} & \multicolumn{1}{c}{0.20} & \multicolumn{1}{c}{0.30} & \multicolumn{1}{c}{0.40} & \multicolumn{1}{c}{0.50} & \multicolumn{1}{c}{0.60} & \multicolumn{1}{c}{0.70} & \multicolumn{1}{c}{0.80} & \multicolumn{1}{c}{0.90} & \multicolumn{1}{c}{1.00} & \multicolumn{1}{c}{1.10} & \multicolumn{1}{c}{1.20} \\ \hline\hline
    Gaussian& 97.9 & 96.5 & 92.6 &86.8 &82.5 & 78.6 & 74.3 & 68.7 & 63.2 &57 & 53.3 & 50.4 & 47.7     \\
     SWEEN &99.2 & 97.1 &94.7 & 87.9 & 82.9 & 78.8& 74.1 & 69.1 &65.8 & 58.6 & 55.8 & 52.5 & 49.6              \\
    SmoothAdv   & 97.1 & 95.9&91.8 & 86.6 & 82.9& 78.6 & \textbf{75.3} & 71.0 & 66.9 & \textbf{63.0} & 56.8 & 53.9 & 51.6              \\
      Consistency     & 99.4 & 98.8 & 95.3 & 90.7 & 83.3 & 78.0 & 74.3 & 71.0 & 65.6 & 60.5 &57.4 & 54.5 & 51.2            \\
            {MultiTask}     & {96.5} & {94.4} & {90.3} & {85.8} & {81.3} & {76.3} & {71.8} & {67.3} & {62.8} & {57.8} & {55.3} & {51.4} & {48.6}           \\
      \name   & \textbf{99.6} & \textbf{99.2}& \textbf{96.7} &\textbf{91.2} & \textbf{84.4} & \textbf{79.4} & \textbf{75.3} & \textbf{72.0} & \textbf{67.9} & \textbf{63.0} & \textbf{57.8} & \textbf{55.1} & \textbf{51.9}           \\ \hline
    \end{tabular}}
    \vspace{-0.5mm}
    \end{table}
    
    \vspace{-1mm}
    \textbf{The implementation of \textit{reasoning} component.}
    For each road sign and attribute (e.g., \texttt{IsStop()} and \texttt{IsOctagon()}), we will build one associated predicate and thus the number of predicates is $32$.
    
    For the attribute-based knowledge/formulas,  following~\cite{gurel2021knowledge}, we treat the attribute that only one of the road signs owns as the permissive attribute and allow each of them to imply its associated road sign. For example, only stop sign is of octagon shape, and thus the corresponding formulas would be  $\texttt{IsOctagon}(x)\implies\texttt{IsStop}(x)$. Next, we treat the remaining attributes as preventative attributes and let each road sign imply them. For example, both~\emph{``do not enter''} sign and~\emph{``no vehicles''} sign are circle, so the corresponding formulas are constructed like $\texttt{IsDoNotEnter}(x)\implies\texttt{IsCircle}(x)$ and $\texttt{IsNoVehicles}(x)\implies\texttt{IsCircle}(x)$. The detailed knowledge construction is in~\Cref{sec:gtsrb_details}, and the final number of the constructed formulas here is $44$.
    
    The hierarchy knowledge/formulas are constructed between the attributes that have inclusion relations. For examples, both the attributes octagon and square are one kind of polygon instead of circle, so we will construct two formulas $\texttt{IsOctagon}(x) \implies \texttt{IsPolygon}(x)$ and $\texttt{IsSquare}(x) \implies \texttt{IsPolygon}(x)$. The full inclusion relations between the attributes are provided in~\Cref{sec:gtsrb_details}, and the final number of the hierarchy formulas is $14$, so the total number of formulas is $44+14=58$.
    
    
    \textbf{Certification results.}
    We report the best certified accuracy of each method under $\sigma \in \{0.12, 0.25, 0.50\}$ for each radius, and the results are shown in \Cref{tab:stop_sign}. As we can see, ~\name is consistently better than if not equal to the best of the baseline approaches under different perturbation radii. The full certification results for each method under different $\sigma$ are in~\Cref{sec:exp_for_different_main}.  The training and the certification details are deferred to~\Cref{sec:exp_baseline} and~\Cref{sec:certification_detail}, respectively.
    
    \renewcommand\arraystretch{1}
    \begin{table}[t!]
        \centering
        \caption{\small Certified accuracy on  PDF malware under  $\ell_0$  radii.}
        \label{tab:pdfmalware}
        \vspace{-2mm}
    \resizebox{\linewidth}{!}{
    \begin{tabular}{c|cccccccccc}
    \hline
    \multirow{2}{*}{Method} & \multicolumn{9}{c}{Certified Accuracy under Radius $r$} \\ \cline{2-11} 
                           &0 & 1    & 2    & 3    & 4    & 5   & 6   & 7   & 8   & 9   \\ \hline
    Lee et al.~\cite{lee2019tight}  & \textbf{99.8}  &   99.0   &  96.1     &   80.0   &   80.0 & 
      68.0 	 & 46.5  &	15.1  &	5.7  &	5.7    \\
    SWEEN  &\textbf{99.8} & 99.0 &	\textbf{97.7} &	85.2 &	80.3 &	72.5 &	57.2 	&22.6 &	8.9 &	8.9 
     \\
     {MultiTask} & {99.7}  & {99.0} & {97.2} & {82.8}   &  {80.5}    &  {72.7}  &  {59.0}    &    {53.8} &  {9.9}  &  {9.9}   \\
    \name   & 99.5 & \textbf{99.3} &	96.9 &	\textbf{85.5} 	&\textbf{84.2} 	&\textbf{77.4} &	\textbf{63.4} &	\textbf{54.5} &	\textbf{13.5} &	\textbf{13.5} 
        \\ \hline
    \end{tabular}}
    \end{table}

\vspace{-3mm}
\subsection{Evaluation on PDF Malware Dataset}
\vspace{-1mm}
    \textbf{Dataset description.} The PDF malware dataset from Contagio~\cite{parkour201716} contains $16800$ clean and $11960$  malicious PDFs. Following the standard setting in~\cite{chen2020training}, we use Hidost~\cite{vsrndic2013detection} to extract the binary structural path features from the parsed tree structure of each PDF with the default~\texttt{compact} path option~\cite{hidost}. An example of the parsed tree structure and the corresponding extracted binary Hidost features are shown in~\Cref{fig:pdfmalware} (a) and~\Cref{fig:pdfmalware} (b) respectively. The final extracted features have $3514$ dimensions, and the feature in each position is a binary value indicating the existence of a specific path.
    
    \textbf{Task and the implementation of \textit{learning} component.}
    The main task is to detect PDF malware. First, we train one main sensor based on the whole $3514$ binary features extracted from PDF with Hidost. Then, we manually pick $6$ malicious features shared with most malicious PDFs but not in most benign PDFs and $8$ benign features shared with most benign PDFs but not in most malicious PDFs. Detailed information on these $14$ selected features is provided in~\Cref{sec:pdf_details}. We assume the adversary can arbitrarily manipulate some of the whole $3514$ features. The final sensing vector $\bm{z}$ is the concatenation of the output confidence of all the sensor output confidence with $2+14=16$ dimensions.
    
    \textbf{The implementation of \textit{reasoning} component.}
    As shown in~\Cref{fig:pdfmalware} (c), first, we construct the predicates $\texttt{IsMalicious}(x)$ and $\texttt{IsBenign}(x)$ to indicate if the input PDF $x$ is malicious or benign. Then, for each of the $14$ picked features  we will construct one predicate to indicate if the structural path exists, such as $\texttt{/Root/OpenAction}(x)$,  $\texttt{/Root/OpenAction/JS}(x)$, $\texttt{/Root/Metadata}(x)$. And thus, the number of predicates is $16$ here.
    
    For the $6$ malicious features, we let the malicious PDF  imply each of them, while the benign PDF will imply the non-existence of them. For instance, $\texttt{IsMalicious}(x) \implies \texttt{/Root/OpenAction/JS}(x)$ and $\texttt{IsBenign}(x) \implies \lnot \texttt{/Root/OpenAction/JS}(x)$. Similarly for the 8 benign features: $\texttt{IsBenign}(x) \implies \texttt{/Root/Metadata/JS}(x)$ and $\texttt{Malicious}(x) \implies \lnot \texttt{/Root/Metadata/JS}(x)$. The total number of the attribute-based formulas here is $14\times2=28$, and more details in~\Cref{sec:pdf_details}
    
    In addition, we also construct the hierarchy-based formulas based on the parsed PDF tree structure, which constructs $12$ more formulas. In specific, for each internal node in the parsed tree, every descendant of it will imply its existence. For instance, $\texttt{/Root/OpenAction/JS}(x) \implies \texttt{/Root/OpenAction}(x)$,  $\texttt{/Root/Metadata/Length}(x)$ $\implies \texttt{/Root/Metadata}(x)$. The number of the overall formulas is $28+12=40$. 

    
    \textbf{Certification results.}
    We report the best certified accuracy of each method under  $\alpha \in \{0.80, 0.85, 0.90\}$ for each radius where $\alpha$ is defined in \Cref{subsec:exp-setup}, and the results are shown in \Cref{tab:pdfmalware}. As we can see, with knowledge integration,~\name  enhances the certified robustness significantly, especially under  large $\ell_0$ perturbation. The full certification results for each method under different  $\alpha$ are in~\Cref{sec:exp_for_different_main}.  The training and the certification details are deferred to~\Cref{sec:exp_baseline} and~\Cref{sec:certification_detail}, respectively.

\renewcommand\arraystretch{1.1}
\begin{table}[t]
    \centering
    \caption{\small Certified accuracy of \name using different number of knowledge sensors on  AwA2.}
    \vspace{-2mm}
    \label{tab:pick}
\resizebox{\linewidth}{!}{
\begin{tabular}{c|c|cccccccccccccc}
\hline
\multirow{2}{*}{$\sigma$} & \multirow{2}{*}{Method} & \multirow{2}{*}{ACR} & \multicolumn{12}{c}{Certified Accuracy under Radius $r$}                          &      \\
                                       &                         &                      & 0.00 & 0.20 & 0.40 & 0.60 & 0.80 & 1.00 & 1.20 & 1.40 & 1.60 & 1.80 & 2.00 & 2.20 & 2.40 \\ \hline\hline
\multirow{6}{*}{0.25}                  & Gaussian                & 0.544                                    & 84.0                     & 77.6                     & 71.4                     & 58.6                     & 40.0                     & 0.0                      & 0.0                      & 0.0                      & 0.0                      & 0.0                      & 0.0                      & 0.0                      & 0.0                      \\
& \name-10               & 0.594 & 89.4 & 84.6 & 76.2 & 65.8 & 46.2 & 0.0 & 0.0 & 0.0 & 0.0 & 0.0 & 0.0 & 0.0 & 0.0                  \\
& \name-30               & 0.639 & 93.8 & 89.2 & 82.4 & 71.6 & 52.8 & 0.0 & 0.0 & 0.0 & 0.0 & 0.0 & 0.0 & 0.0 & 0.0                  \\
  & \name-50               & 0.671 & 94.8 & 91.6 & 87.8 & 77.6 & 55.4 & 0.0 & 0.0 & 0.0 & 0.0 & 0.0 & 0.0 & 0.0 & 0.0                  \\
 & \name-70               & 0.703 & 96.0 & \textbf{94.2} & 90.2 & 83.4 & 65.6 & 0.0 & 0.0 & 0.0 & 0.0 & 0.0 & 0.0 & 0.0 & 0.0             \\
 & \name-All              &\textbf{0.709} & \textbf{96.6} & \textbf{94.2} & \textbf{91.4} & \textbf{84.8} & \textbf{67.4} & 0.0 & 0.0 & 0.0 & 0.0 & 0.0 & 0.0 & 0.0 & 0.0                    \\ \hline\hline
\multirow{6}{*}{0.50}                  & Gaussian                & 0.827                                    & 75.6                     & 71.2                     & 64.6                     & 58.2                     & 53.0                     & 46.2                     & 38.8                     & 32.0                     & 21.2                     & 0.0                      & 0.0                      & 0.0                      & 0.0                      \\
 & \name-10               & 0.892 & 79.2 & 75.0 & 69.2 & 62.4 & 56.2 & 50.0 & 43.6 & 36.0 & 25.6 & 0.0 & 0.0 & 0.0 & 0.0                      \\
 & \name-30               & 0.956 & 85.0 & 80.0 & 75.0 & 65.8 & 60.6 & 53.6 & 47.8 & 37.6 & 28.2 & 0.0 & 0.0 & 0.0 & 0.0                  \\
 & \name-50               & 1.015 & 88.0 & 82.8 & 78.0 & 72.2 & 65.6 & 58.2 & 51.0 & 41.4 & 29.0 & 0.0 & 0.0 & 0.0 & 0.0                    \\
 & \name-70               & 1.047 & 89.0 & 86.4 & 80.4 & 73.2 & 67.4 & 60.4 & 50.6 & 43.0 & 31.6 & 0.0 & 0.0 & 0.0 & 0.0                   \\
 & \name-All              & \textbf{1.114} & \textbf{91.2}& \textbf{88.2} & \textbf{84.2} & \textbf{78.8} & \textbf{71.2 }& \textbf{66.4} & \textbf{56.8 }& \textbf{46.8 }& \textbf{34.6} & 0.0 & 0.0 & 0.0 & 0.0      \\ \hline\hline
\multirow{6}{*}{1.00}                  & Gaussian                & 0.994                                    & 59.6                     & 54.6                     & 51.6                     & 49.0                     & 44.8                     & 40.8                     & 36.6                     & 32.6                     & 29.6                     & 26.4                     & 22.8                     & 20.0                     & 17.2                     \\
 & \name-10               & 1.196 & 67.6 & 62.0 & 59.2 & 55.2 & 50.4 & 46.6 & 44.6 & 41.6 & 37.8 & 33.8 & 29.2 & 26.0 & 21.8                     \\
 & \name-30               &1.542 & 76.8 & 73.8 & 71.6 & 67.8 & 64.8 & 61.0 & 57.0 & 51.6 & 48.4 & 45.2 & 40.8 & 38.0 & 32.4                 \\
 & \name-50               & 1.796 & 81.8 & 80.2 & 76.8 & 74.6 & 72.8 & 69.0 & 67.4 & 62.6 & 57.6 & 54.0 & 49.0 & 45.4 & 41.6                \\
  & \name-70               & 2.033 & 86.0 & \textbf{85.2} & 82.6 & 80.8 & 78.6 & 76.4 & 72.4 & 70.4 & 67.4 & 64.2 & 57.8 & 53.8 & 50.2                   \\
& \name-All              & \textbf{2.092} & \textbf{87.0} & \textbf{85.2} & \textbf{84.0} & \textbf{82.0} & \textbf{80.4} & \textbf{78.2} & \textbf{75.6} & \textbf{71.2} & \textbf{68.0} & \textbf{64.4} & \textbf{61.0} & \textbf{57.0} & \textbf{52.8}            \\ \hline
\end{tabular}}
\vspace{-5mm}
\end{table}

\renewcommand\arraystretch{1}
\vspace{-5mm}
\begin{table}[t]
    \centering
    \caption{\small Empirical robust accuracy of different methods on AwA2, Word50, and GTSRB under $\ell_{2}$ attacks.}
    \vspace{-2mm}
    \label{tab:attack_l2_all}
\resizebox{\linewidth}{!}{
\begin{tabular}{c|cccc|cccc|cccc}
\hline
\multirow{3}{*}{Method} & \multicolumn{4}{c|}{AwA2}& \multicolumn{4}{c|}{Word50}& \multicolumn{4}{c}{GTSRB}\\ \cline{2-13} 
& \multicolumn{1}{c|}{\multirow{2}{*}{$\sigma$}} & \multicolumn{3}{c|}{$\epsilon$} & \multicolumn{1}{c|}{\multirow{2}{*}{$\sigma$}} & \multicolumn{3}{c|}{$\epsilon$} & \multicolumn{1}{c|}{\multirow{2}{*}{$\sigma$}} & \multicolumn{3}{c}{$\epsilon$} \\ 
& \multicolumn{1}{c|}{}& 1.8           & 2.4           & 3.0          & \multicolumn{1}{c|}{}                                       & 0.4          & 0.8         & 1.2          & \multicolumn{1}{c|}{}& 0.6          & 1.2        & 1.8          \\ \hline
Gaussian& \multicolumn{1}{c|}{\multirow{6}{*}{0.25}}& 29.8          & 16.6          & 8.6          & \multicolumn{1}{c|}{\multirow{6}{*}{0.12}} & 10.4 & 1.0 & 0.0           & \multicolumn{1}{c|}{\multirow{6}{*}{0.12}} &67.9 & 41.8 & 31.3 \\
SWEEN& \multicolumn{1}{c|}{}& 33.0          & 17.4          & 10.4         & \multicolumn{1}{c|}{}                                       & 27.8 & 11.2 & 3.8        & \multicolumn{1}{c|}{} & 78.0 &    58.2  & 46.1              \\
SmoothAdv& \multicolumn{1}{c|}{}& 48.4          & 36.4          & 25.4         & \multicolumn{1}{c|}{}    &  30.0 & 8.4 & 2.2         & \multicolumn{1}{c|}{}& 72.2             &  51.9 & 39.9             \\
Consistency& \multicolumn{1}{c|}{}& 40.4          & 29.0          & 17.8         & \multicolumn{1}{c|}{}&  21.0 & 4.6 & 0.6  & \multicolumn{1}{c|}{}&   73.3    & 52.9              &  52.3             \\
MultiTask& \multicolumn{1}{c|}{}& 36.4         & 24.6          & 15.6          & \multicolumn{1}{c|}{}&  15.2 & 1.6 & 0.2  & \multicolumn{1}{c|}{}&   64.0   & 35.2              &  18.5             \\
\name& \multicolumn{1}{c|}{}&\textbf{66.4}    &  \textbf{42.4}   &  \textbf{26.2}         & \multicolumn{1}{c|}{} &  \textbf{86.4} & \textbf{82.4} & \textbf{80.4}          & \multicolumn{1}{c|}{} & \textbf{79.8}& \textbf{58.8}              &   \textbf{52.5}           \\ \hline\hline

Gaussian& \multicolumn{1}{c|}{\multirow{6}{*}{0.50}}& 40.4    &   31.4     &   21.4         & \multicolumn{1}{c|}{\multirow{6}{*}{0.25}}& 13.6 & 2.2 & 0.0         & \multicolumn{1}{c|}{\multirow{6}{*}{0.25}}&     72.0          &     47.7         &    28.2          \\
SWEEN& \multicolumn{1}{c|}{}&  43.0   &   33.4    &  23.0          & \multicolumn{1}{c|}{}&  27.6 & 14.4 & 3.2        & \multicolumn{1}{c|}{} &     72.6          &       49.4       &         32.1     \\
SmoothAdv& \multicolumn{1}{c|}{}&  46.0 & 38.6 & 31.6          & \multicolumn{1}{c|}{}&  29.8 & 10.2 & 2.4  & \multicolumn{1}{c|}{}&     73.9          &       54.1       &    35.4          \\
Consistency& \multicolumn{1}{c|}{}&  47.0 & 37.0 & 29.4       & \multicolumn{1}{c|}{}&  24.2 & 8.2 & 0.4    & \multicolumn{1}{c|}{}&        73.7       &    51.0          &    34.8          \\
MultiTask& \multicolumn{1}{c|}{}& 42.2    & 32.8              &  25.6         & \multicolumn{1}{c|}{}&  16.2 & 1.8 & 0.4  & \multicolumn{1}{c|}{}&  68.9   & 45.5              &  27.0             \\
\name& \multicolumn{1}{c|}{}& \textbf{67.2}     &  \textbf{55.6}    &   \textbf{44.6}      & \multicolumn{1}{c|}{}& \textbf{90.0} & \textbf{84.2} & \textbf{80.2}         & \multicolumn{1}{c|}{}&      \textbf{75.1}         &        \textbf{56.2}      &       \textbf{37.4}       \\ \hline\hline

Gaussian& \multicolumn{1}{c|}{\multirow{6}{*}{1.00}}& 39.2    &   32.8     &   25.8          & \multicolumn{1}{c|}{\multirow{6}{*}{0.50}}&  12.6 & 3.6 & 1.0         & \multicolumn{1}{c|}{\multirow{6}{*}{0.50}}                  &     67.3          &   45.9           &    26.5          \\
SWEEN& \multicolumn{1}{c|}{} &  40.0   &   34.2    &  28.6      & \multicolumn{1}{c|}{}&  20.0 & 10.6 & 4.4         & \multicolumn{1}{c|}{} &    67.7           &       47.5       &      29.4        \\
SmoothAdv& \multicolumn{1}{c|}{} &  39.4 & 34.0 & 30.8    & \multicolumn{1}{c|}{}                                       &  20.0 & 9.8 & 3.6      & \multicolumn{1}{c|}{}&  65.6             & 49.4  & 34.6              \\
Consistency& \multicolumn{1}{c|}{}&  39.8 & 35.0 & 31.6       & \multicolumn{1}{c|}{} &  15.8 & 7.6 & 3.2 & \multicolumn{1}{c|}{}& 70.2 & 50.8              &     32.5         \\
MultiTask& \multicolumn{1}{c|}{}& 35.2          & 31.4          & 26.8         & \multicolumn{1}{c|}{}&  14.6 & 5.2 & 0.6  & \multicolumn{1}{c|}{}&   67.5    & 47.9              &  28.6             \\
\name& \multicolumn{1}{c|}{}& \textbf{76.8}    & \textbf{73.4}   &   \textbf{68.6}   & \multicolumn{1}{c|}{}&  \textbf{80.0} & \textbf{72.6} & \textbf{63.2}         & \multicolumn{1}{c|}{}&       \textbf{72.0}        & \textbf{56.6} & \textbf{35.0}\\ \hline
\end{tabular}}
\end{table}

 \begin{figure*}[t]
    \centering
    \includegraphics[width=\textwidth]{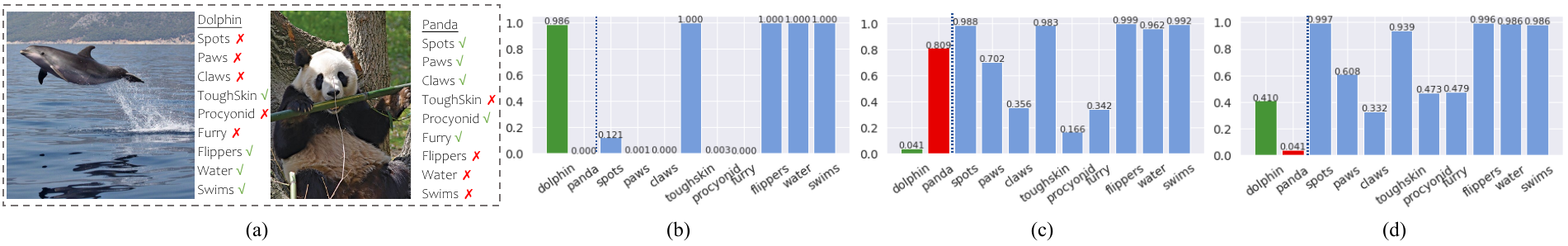}
    \vspace{-8mm}
    \caption{A case study on AwA2. (a) Attributes for dolphin and panda classes; (b) prediction confidence of a DNN before attack; (c) prediction confidence of the DNN after the adversarial attack; (d) prediction confidence of \name given the same adversarial input. (The ground truth is~\emph{``dolphin''}).} 
    \label{fig:hist}
    \vspace{-7mm}
\end{figure*} 

\vspace{1.7mm}
\subsection{Ablation Studies}
\vspace{-1mm}
\label{sec:ablation}
In addition to the improved certified robustness of \name, we conduct a set of ablation studies to further explore the impact of the integrated number of knowledge rules, the empirical robustness of different methods, and the interesting explanation properties of \name.

 \textbf{Number of knowledge rules.}
    Generally, the certified robustness will improve with the increase of the number of integrated knowledge rules. Here we quantitatively analyze this phenomenon on the AwA2 dataset. In particular, we randomly pick $k$ knowledge sensors and build the formulas based on their corresponding provided knowledge rules. Then, we will retrain the GCN with these limited confidence vectors and knowledge rules. For simplicity, we denote \name-$k$ as the model with $k$ knowledge sensors. As shown in~\Cref{tab:pick}, the certified robustness of \name improves with the increase of the number of knowledge rules (sensors). Interestingly, \name still beats the baselines with only $10$ knowledge sensors. {However, simply increasing the number of the base model for the ensemble method will only provide marginal performance as shown in~\Cref{adx:ensemble}, and this phenomenon has also been theoretically proven in~\cite{yang2022on}.}
    
    \textbf{Empirical robustness.}
    Except for the certified robustness, we also evaluate the empirical robustness of \name on AwA2, Word50, and GTSRB to demonstrate the effectiveness of our method further. Concretely, we will first attack the main sensor to generate adversarial instances and then test them on the \name pipeline. Since the main sensor here is a smoothed model, we follow~\cite{hayes2020extensions} to sample $100$ Monte Carlo samples to estimate the gradient, and the detailed procedure of the empirical attack is provided in~\Cref{sec:emp_attack_detail}.
    
    We provide the experimental results for both $\ell_2$ and $\ell_{\infty}$ attacks on our model and report the empirical robust accuracy on the main task (i.e., Animal class prediction on AwA; word prediction on Word50; road sign prediction on GTSRB) for baselines and our method. Under $\ell_2$ attack, the number of update steps in PGD is fixed to $100$, and the attack step size is set to $0.2$. The $\ell_2$ attack results under different perturbation magnitude $\epsilon$ for dataset AwA2, Word50 and GTSRB are shown in~\Cref{tab:attack_l2_all}. 
    The results for $\ell_{\infty}$ attack {and the exploration for attacking with the attributes} are deferred to~\Cref{sec:emp_attack_detail}.

    \textbf{Transferability between sensors.}
    The large improvement of the empirical robustness with knowledge integration can be attributed to: $(a)$  low attack transferability between different sensors;  $(b)$ high difficulty of attacking all sensors at the same time such that their predictions still satisfy the knowledge rules among them. Here we evaluate the attack transferability between $12$ sensors (one main sensor and eleven random picked attribute sensors, all are trained under $\sigma = 0.50$) on AwA2 under $\ell_2$ perturbation size $\epsilon = 3.0$, and the corresponding results are shown in~\Cref{fig:heatmap} (in~\Cref{sec:emp_attack_detail}). All the sensors here are trained under  Gaussian noise with $\sigma = 0.5$, and the result in the $i$-th row and the $j$-th column represents the empirical robust accuracy when we conduct a PGD attack on the $i$-th sensor while testing the obtained adversarial images on $j$-th sensor. As we can see, when we attack one of the sensors, only a few other sensors will be influenced, and the reason can be attributed to the fact that different knowledge sensors will rely on different features.
    As the adversary usually can only flip a small set of the sensors, which could be again corrected by the knowledge rules, the overall pipeline \name is still robust against perturbations.

    \textbf{Case studies on AwA2.}
    To better demonstrate the knowledge correction process in \name, we provide some case studies against adversarial attacks. As shown in~\Cref{fig:hist}, the ground label for the input is~\emph{``Dolphin''} and it is predicted correctly by the main sensor originally (\Cref{fig:hist} (b)); then we perform a strong targeted attack with $100$ update steps to make it be misclassified as~\emph{``Panda''} (\Cref{fig:hist} (c)). All the sensors here are trained under $\sigma = 0.25$, the perturbation size $\epsilon$ and the attack step size here are set to $6.0$ and $0.4$, respectively. Some attributes for these two classes are shown in~\Cref{fig:hist} (a). As we can see, although some attributes like~\emph{``Spots''} and~\emph{``Paws''} are attacked, most of the attributes remain unattacked (e.g.~\emph{``Toughskin'', ``Flippers'', ``Water'', ``Swims''}), and thus after passing through the \name pipeline, the knowledge rules among these attributes would correct the sensor predictions (\Cref{fig:hist} (d)).
    More examples are shown in~\Cref{sec:case_awa}.

\vspace{-4mm}
\section{Related Work}
\vspace{-2mm}
    \label{sec:related-work}
    \textbf{Knowledge integration and logical reasoning.}
    There is abundant domain knowledge in real-world data. For instance, the labels in ImageNet~\cite{deng2009imagenet} contains a semantic hierarchy structure based on the lexical database WordNet~\cite{miller1995wordnet}.
    Thus, how to quantitatively represent and effectively integrate such knowledge is an important research direction.
    In particular, Bayesian logic programs~\cite{kersting2001towards}, relational Markov networks~\cite{taskar2007relational} and Markov logic networks~\cite{richardson2006markov} have been used for knowledge reasoning. In addition, with the development of deep learning, some works have started to introduce structured logic rules into a neural network to improve performance. For example, Deng et al.~\cite{deng2014large}  construct hierarchy and exclusion graphs, which are based on the hierarchical relations between classes, to improve the classification on ImageNet. Hu et al.~\cite{hu2016harnessing} develop a distillation method to encode the knowledge into the weight of neural networks. However, leveraging such domain knowledge and relationships to improve the certified robustness of DNNs has not been well explored yet, and this work provides the first learning with a reasoning pipeline to improve the \textit{certified} robustness of DNNs. 

    \textbf{Markov Logic Networks.} 
    MLNs, which extend the probabilistic graphical model with first-order logic, has
    been largely used  for solving the problems like collective classification~\cite{crane2012investigating}, link prediction~\cite{domingos20071} and  entity resolution~\cite{singla2006entity}. 
    However, the inference of MLN is \textsf{\#P}-complete, and it can be either solved with variable elimination-based methods like belief propagation~\cite{yedidia2000generalized,yedidia2003understanding} and junction tree algorithm~\cite{kahle2008junction} or approximated by random sampling like Markov chain Monte Carlo (MCMC)~\cite{gilks1995markov} and importance sampling~\cite{venugopal2014scaling}. Nevertheless, MLN is still hard to be scaled for large knowledge graphs in practice, and the combination of deep neural networks and MLN is still constrained on small dataset~\cite{yang2020end}. Therefore, our method~\name aims to provide a more robust and scalable framework for such MLN-based logical reasoning via variational inference~\cite{jordan1999introduction} and equip it with a more powerful posterior parameterization by graph neural network.
    
    \textbf{Graph Neural Networks.}
    GNN~\cite{scarselli2008graph} is well recognized for its superior performance in handling large-scale knowledge graphs and the effective encoding ability. Different from classic knowledge graph embedding  such as TransE~\cite{bordes2013translating}, DistMult~\cite{kadlec2017knowledge}, and RotatE~\cite{sun2018rotate}, which cannot leverage prior domain knowledge, GNN models such as Graph Convolutional Network (GCN)~\cite{2016tc} can be used to learn semantically-constrained embeddings~\cite{xie2019embedding}. In addition, Qu et al.~\cite{qu2019gmnn} propose a Graph Markov Neural Network (GMNN), which combines GNN with a conditional random field to improve the performance of semi-supervised object classification and link classification in relational data. These works have provided valuable experience in projecting traditional graphical models to deep neural networks for efficient training and inference.
    
    \textbf{Certified robustness.}
    The robustness certification aims to ensure that the prediction of a classifier is consistent within a certain perturbation radius~\cite{li2023sok}. Currently, there are mainly two types of certification methods. The \textit{complete} certification, which guarantees to find the perturbation if it exists, is usually based on satisfiability modulo theories~\cite{katz2017reluplex, ehlers2017formal}, or mixed integer-linear programming~\cite{lomuscio2017approach, fischetti2017deep}. 
    However, the exact certification is \texttt{NP-complete} for feed-forward networks.
    The \textit{incomplete} certification, guarantees to find non-certifiable instances, while may miss some certifiable ones based on different relaxed optimization. With such relaxation, incomplete certification is usually more practical and efficient, which is mainly based on linear programming relaxation~\cite{zhang2018efficient, salman2019convex} or semi-definite programming~\cite{raghunathan2018certified, raghunathan2018semidefinite}. However, these incomplete certification approaches are only applicable for specific architecture and can not scale to a large dataset like ImageNet. 
    Later, Cohen et al.~\cite{cohen2019certified} provide a \textit{probabilistic} certification method based on randomized smoothing, which can be scaled to ImageNet and is further improved with adversarial training~\cite{salman2019provably} and consistency regularization~\cite{jeong2020consistency}.
   
\section{Conclusion}
\vspace{-1mm}
In this work, we propose the first scalable certifiably robust machine learning pipeline \name by integrating knowledge to enable reasoning ability for reliable prediction. We show that when combining learning with reasoning, \name can effectively scale to large datasets and achieve both high certified robustness and empirical robustness. We believe our observations and findings will inspire interesting future directions on leveraging domain knowledge to improve ML robustness.

\vspace{-1mm}
\section*{Acknowledgment}
\vspace{-1mm}
This work is partially supported by NSF grant No.1910100, NSF CNS No.2046726, a C3.ai DTI Award, and the Alfred P. Sloan Foundation.

\newpage


\bibliographystyle{IEEEtran}
\bibliography{bib}

\begin{thebibliography}{10}
\providecommand{\url}[1]{#1}
\csname url@samestyle\endcsname
\providecommand{\newblock}{\relax}
\providecommand{\bibinfo}[2]{#2}
\providecommand{\BIBentrySTDinterwordspacing}{\spaceskip=0pt\relax}
\providecommand{\BIBentryALTinterwordstretchfactor}{4}
\providecommand{\BIBentryALTinterwordspacing}{\spaceskip=\fontdimen2\font plus
\BIBentryALTinterwordstretchfactor\fontdimen3\font minus
  \fontdimen4\font\relax}
\providecommand{\BIBforeignlanguage}[2]{{%
\expandafter\ifx\csname l@#1\endcsname\relax
\typeout{** WARNING: IEEEtran.bst: No hyphenation pattern has been}%
\typeout{** loaded for the language `#1'. Using the pattern for}%
\typeout{** the default language instead.}%
\else
\language=\csname l@#1\endcsname
\fi
#2}}
\providecommand{\BIBdecl}{\relax}
\BIBdecl

\bibitem{biggio2013evasion}
B.~Biggio, I.~Corona, D.~Maiorca, B.~Nelson, N.~{\v{S}}rndi{\'c}, P.~Laskov,
  G.~Giacinto, and F.~Roli, ``Evasion attacks against machine learning at test
  time,'' in \emph{Joint European conference on machine learning and knowledge
  discovery in databases}.\hskip 1em plus 0.5em minus 0.4em\relax Springer,
  2013, pp. 387--402.

\bibitem{szegedy2014intriguing}
C.~Szegedy, W.~Zaremba, I.~Sutskever, J.~Bruna, D.~Erhan, I.~Goodfellow, and
  R.~Fergus, ``Intriguing properties of neural networks,'' in \emph{2nd
  International Conference on Learning Representations, ICLR 2014}, 2014.

\bibitem{xiao2018generating}
C.~Xiao, B.~Li, J.-Y. Zhu, W.~He, M.~Liu, and D.~Song, ``Generating adversarial
  examples with adversarial networks,'' \emph{AAAI}, 2018.

\bibitem{xiao2018spatially}
C.~Xiao, J.~Y. Zhu, B.~Li, W.~He, M.~Liu, and D.~Song, ``Spatially transformed
  adversarial examples,'' in \emph{6th International Conference on Learning
  Representations, ICLR 2018}, 2018.

\bibitem{cao2021invisible}
\BIBentryALTinterwordspacing
Y.~Cao, N.~Wang, C.~Xiao, D.~Yang, J.~Fang, R.~Yang, Q.~Chen, M.~Liu, and
  B.~Li, ``Invisible for both camera and lidar: Security of multi-sensor fusion
  based perception in autonomous driving under physical-world attacks,'' in
  \emph{2021 IEEE Symposium on Security and Privacy (SP)}.\hskip 1em plus 0.5em
  minus 0.4em\relax Los Alamitos, CA, USA: IEEE Computer Society, may 2021, pp.
  1302--1320. [Online]. Available:
  \url{https://doi.ieeecomputersociety.org/10.1109/SP40001.2021.00076}
\BIBentrySTDinterwordspacing

\bibitem{eykholt2018robust}
K.~Eykholt, I.~Evtimov, E.~Fernandes, B.~Li, A.~Rahmati, C.~Xiao, A.~Prakash,
  T.~Kohno, and D.~Song, ``Robust physical-world attacks on deep learning
  visual classification,'' in \emph{Proceedings of the IEEE Conference on
  Computer Vision and Pattern Recognition}, 2018, pp. 1625--1634.

\bibitem{erickson2017machine}
B.~J. Erickson, P.~Korfiatis, Z.~Akkus, and T.~L. Kline, ``Machine learning for
  medical imaging,'' \emph{Radiographics}, vol.~37, no.~2, p. 505, 2017.

\bibitem{magoulas1999machine}
G.~D. Magoulas and A.~Prentza, ``Machine learning in medical applications,'' in
  \emph{Advanced course on artificial intelligence}.\hskip 1em plus 0.5em minus
  0.4em\relax Springer, 1999, pp. 300--307.

\bibitem{xiao2018characterizing}
C.~Xiao, R.~Deng, B.~Li, F.~Yu, M.~Liu, and D.~Song, ``Characterizing
  adversarial examples based on spatial consistency information for semantic
  segmentation,'' in \emph{Proceedings of the European Conference on Computer
  Vision (ECCV)}, 2018, pp. 217--234.

\bibitem{yang2021trs}
Z.~Yang, L.~Li, X.~Xu, S.~Zuo, Q.~Chen, P.~Zhou, B.~I.~P. Rubinstein, C.~Zhang,
  and B.~Li, ``Trs: Transferability reduced ensemble via promoting gradient
  diversity and model smoothness,'' in \emph{Advances in Neural Information
  Processing Systems}, 2021.

\bibitem{yang2022on}
Z.~Yang, L.~Li, X.~Xu, B.~Kailkhura, T.~Xie, and B.~Li, ``On the certified
  robustness for ensemble models and beyond,'' in \emph{International
  Conference on Learning Representations}, 2022.

\bibitem{samangouei2018defense}
P.~Samangouei, M.~Kabkab, and R.~Chellappa, ``Defense-gan: Protecting
  classifiers against adversarial attacks using generative models,''
  \emph{arXiv preprint arXiv:1805.06605}, 2018.

\bibitem{cohen2019certified}
J.~Cohen, E.~Rosenfeld, and Z.~Kolter, ``Certified adversarial robustness via
  randomized smoothing,'' in \emph{International Conference on Machine
  Learning}.\hskip 1em plus 0.5em minus 0.4em\relax PMLR, 2019, pp. 1310--1320.

\bibitem{carlini2017adversarial}
N.~Carlini and D.~Wagner, ``Adversarial examples are not easily detected:
  Bypassing ten detection methods,'' in \emph{Proceedings of the 10th ACM
  workshop on artificial intelligence and security}, 2017, pp. 3--14.

\bibitem{athalye2018obfuscated}
A.~Athalye, N.~Carlini, and D.~Wagner, ``Obfuscated gradients give a false
  sense of security: Circumventing defenses to adversarial examples,'' in
  \emph{International conference on machine learning}.\hskip 1em plus 0.5em
  minus 0.4em\relax PMLR, 2018, pp. 274--283.

\bibitem{gowal2018effectiveness}
S.~Gowal, K.~Dvijotham, R.~Stanforth, R.~Bunel, C.~Qin, J.~Uesato,
  R.~Arandjelovic, T.~Mann, and P.~Kohli, ``On the effectiveness of interval
  bound propagation for training verifiably robust models,'' \emph{arXiv
  preprint arXiv:1810.12715}, 2018.

\bibitem{zhang2019towards}
H.~Zhang, H.~Chen, C.~Xiao, S.~Gowal, R.~Stanforth, B.~Li, D.~Boning, and C.-J.
  Hsieh, ``Towards stable and efficient training of verifiably robust neural
  networks,'' in \emph{International Conference on Learning Representations},
  2019.

\bibitem{salman2019convex}
H.~Salman, G.~Yang, H.~Zhang, C.-J. Hsieh, and P.~Zhang, ``A convex relaxation
  barrier to tight robustness verification of neural networks,'' \emph{Advances
  in Neural Information Processing Systems}, vol.~32, pp. 9835--9846, 2019.

\bibitem{yang2020randomized}
G.~Yang, T.~Duan, J.~E. Hu, H.~Salman, I.~Razenshteyn, and J.~Li, ``Randomized
  smoothing of all shapes and sizes,'' in \emph{International Conference on
  Machine Learning}.\hskip 1em plus 0.5em minus 0.4em\relax PMLR, 2020, pp.
  10\,693--10\,705.

\bibitem{richardson2006markov}
M.~Richardson and P.~Domingos, ``Markov logic networks,'' \emph{Machine
  learning}, vol.~62, no.~1, pp. 107--136, 2006.

\bibitem{gurel2021knowledge}
N.~M. G{\"u}rel, X.~Qi, L.~Rimanic, C.~Zhang, and B.~Li, ``Knowledge enhanced
  machine learning pipeline against diverse adversarial attacks,'' \emph{ICML},
  2021.

\bibitem{yang2020end}
Z.~Yang, Z.~Zhao, H.~Pei, B.~Wang, B.~Karlas, J.~Liu, H.~Guo, B.~Li, and
  C.~Zhang, ``End-to-end robustness for sensing-reasoning machine learning
  pipelines,'' \emph{arXiv preprint arXiv:2003.00120}, 2020.

\bibitem{2016tc}
T.~N. Kipf and M.~Welling, ``Semi-supervised classification with graph
  convolutional networks,'' in \emph{5th International Conference on Learning
  Representations, ICLR 2017}, 2017.

\bibitem{gilks1995markov}
W.~R. Gilks, S.~Richardson, and D.~Spiegelhalter, \emph{Markov chain Monte
  Carlo in practice}.\hskip 1em plus 0.5em minus 0.4em\relax CRC press, 1995.

\bibitem{6413813}
F.~Niu, C.~Zhang, C.~Ré, and J.~Shavlik, ``Scaling inference for markov logic
  via dual decomposition,'' in \emph{2012 IEEE 12th International Conference on
  Data Mining}, 2012, pp. 1032--1037.

\bibitem{murphy2013loopy}
K.~Murphy, Y.~Weiss, and M.~I. Jordan, ``Loopy belief propagation for
  approximate inference: An empirical study,'' \emph{arXiv preprint
  arXiv:1301.6725}, 2013.

\bibitem{jordan1999introduction}
M.~I. Jordan, Z.~Ghahramani, T.~S. Jaakkola, and L.~K. Saul, ``An introduction
  to variational methods for graphical models,'' \emph{Machine learning},
  vol.~37, no.~2, pp. 183--233, 1999.

\bibitem{qu2019gmnn}
M.~Qu, Y.~Bengio, and J.~Tang, ``Gmnn: Graph markov neural networks,'' in
  \emph{International conference on machine learning}.\hskip 1em plus 0.5em
  minus 0.4em\relax PMLR, 2019, pp. 5241--5250.

\bibitem{zhang2019efficient}
Y.~Zhang, X.~Chen, Y.~Yang, A.~Ramamurthy, B.~Li, Y.~Qi, and L.~Song,
  ``Efficient probabilistic logic reasoning with graph neural networks,'' in
  \emph{International Conference on Learning Representations}, 2019.

\bibitem{xian2018zero}
Y.~Xian, C.~H. Lampert, B.~Schiele, and Z.~Akata, ``Zero-shot learning—a
  comprehensive evaluation of the good, the bad and the ugly,'' \emph{IEEE
  transactions on pattern analysis and machine intelligence}, vol.~41, no.~9,
  pp. 2251--2265, 2018.

\bibitem{chen2015learning}
L.-C. Chen, A.~Schwing, A.~Yuille, and R.~Urtasun, ``Learning deep structured
  models,'' in \emph{International Conference on Machine Learning}.\hskip 1em
  plus 0.5em minus 0.4em\relax PMLR, 2015, pp. 1785--1794.

\bibitem{stallkamp2012man}
J.~Stallkamp, M.~Schlipsing, J.~Salmen, and C.~Igel, ``Man vs. computer:
  Benchmarking machine learning algorithms for traffic sign recognition,''
  \emph{Neural networks}, vol.~32, pp. 323--332, 2012.

\bibitem{parkour201716}
\BIBentryALTinterwordspacing
M.~Parkour, ``16,800 clean and 11,960 malicious files for signature testing and
  research.'' [Online]. Available:
  \url{http://contagiodump.blogspot.com/2013/03/16800-clean-and-11960-malicious-files.html}
\BIBentrySTDinterwordspacing

\bibitem{miller1995wordnet}
G.~A. Miller, ``Wordnet: a lexical database for english,'' \emph{Communications
  of the ACM}, vol.~38, no.~11, pp. 39--41, 1995.

\bibitem{salman2019provably}
H.~Salman, J.~Li, I.~P. Razenshteyn, P.~Zhang, H.~Zhang, S.~Bubeck, and
  G.~Yang, ``Provably robust deep learning via adversarially trained smoothed
  classifiers,'' in \emph{NeurIPS}, 2019.

\bibitem{jeong2020consistency}
J.~Jeong and J.~Shin, ``Consistency regularization for certified robustness of
  smoothed classifiers,'' in \emph{34th Conference on Neural Information
  Processing Systems (NeurIPS) 2020}.\hskip 1em plus 0.5em minus 0.4em\relax
  NeurIPS committee, 2020.

\bibitem{liu2020enhancing}
C.~Liu, Y.~Feng, R.~Wang, and B.~Dong, ``Enhancing certified robustness via
  smoothed weighted ensembling,'' \emph{arXiv preprint arXiv:2005.09363}, 2020.

\bibitem{lee2019tight}
G.-H. Lee, Y.~Yuan, S.~Chang, and T.~Jaakkola, ``Tight certificates of
  adversarial robustness for randomly smoothed classifiers,'' \emph{Advances in
  Neural Information Processing Systems}, vol.~32, 2019.

\bibitem{li2023sok}
L.~Li, T.~Xie, and B.~Li, ``Sok: Certified robustness for deep neural
  networks,'' in \emph{44th {IEEE} Symposium on Security and Privacy, {SP}
  2023, San Francisco, CA, USA, 22-26 May 2023}.\hskip 1em plus 0.5em minus
  0.4em\relax {IEEE}, 2023.

\bibitem{liu2021algorithms}
C.~Liu, T.~Arnon, C.~Lazarus, C.~Strong, C.~Barrett, M.~J. Kochenderfer
  \emph{et~al.}, ``Algorithms for verifying deep neural networks,''
  \emph{Foundations and Trends{\textregistered} in Optimization}, vol.~4, no.
  3-4, pp. 244--404, 2021.

\bibitem{carlini2022certified}
N.~Carlini, F.~Tramer, J.~Z. Kolter \emph{et~al.}, ``(certified!!) adversarial
  robustness for free!'' \emph{arXiv preprint arXiv:2206.10550}, 2022.

\bibitem{domingos2019unifying}
P.~Domingos and D.~Lowd, ``Unifying logical and statistical ai with markov
  logic,'' \emph{Communications of the ACM}, vol.~62, no.~7, pp. 74--83, 2019.

\bibitem{liu2014generalizable}
Z.~Liu and G.~von Wichert, ``A generalizable knowledge framework for semantic
  indoor mapping based on markov logic networks and data driven mcmc,''
  \emph{Future Generation Computer Systems}, vol.~36, pp. 42--56, 2014.

\bibitem{ghahramani2000graphical}
Z.~Ghahramani and M.~J. Beal, ``Graphical models and variational methods,'' in
  \emph{Advanced Mean Field Methods-Theory and Practice}.\hskip 1em plus 0.5em
  minus 0.4em\relax Citeseer, 2000.

\bibitem{qu2019probabilistic}
M.~Qu and J.~Tang, ``Probabilistic logic neural networks for reasoning,''
  \emph{Advances in neural information processing systems}, vol.~32, 2019.

\bibitem{besag1975statistical}
J.~Besag, ``Statistical analysis of non-lattice data,'' \emph{Journal of the
  Royal Statistical Society: Series D (The Statistician)}, vol.~24, no.~3, pp.
  179--195, 1975.

\bibitem{he2016deep}
K.~He, X.~Zhang, S.~Ren, and J.~Sun, ``Deep residual learning for image
  recognition,'' in \emph{Proceedings of the IEEE conference on computer vision
  and pattern recognition}, 2016, pp. 770--778.

\bibitem{chen2020training}
Y.~Chen, S.~Wang, D.~She, and S.~Jana, ``On training robust $\{$PDF$\}$ malware
  classifiers,'' in \emph{29th USENIX Security Symposium (USENIX Security 20)},
  2020, pp. 2343--2360.

\bibitem{he2015delving}
K.~He, X.~Zhang, S.~Ren, and J.~Sun, ``Delving deep into rectifiers: Surpassing
  human-level performance on imagenet classification,'' in \emph{Proceedings of
  the IEEE international conference on computer vision}, 2015, pp. 1026--1034.

\bibitem{zhai2019macer}
R.~Zhai, C.~Dan, D.~He, H.~Zhang, B.~Gong, P.~Ravikumar, C.-J. Hsieh, and
  L.~Wang, ``Macer: Attack-free and scalable robust training via maximizing
  certified radius,'' in \emph{International Conference on Learning
  Representations}, 2019.

\bibitem{de2009character}
T.~E. De~Campos, B.~R. Babu, M.~Varma \emph{et~al.}, ``Character recognition in
  natural images.'' \emph{VISAPP (2)}, vol.~7, no.~2, 2009.

\bibitem{vsrndic2013detection}
N.~{\v{S}}rndic and P.~Laskov, ``Detection of malicious pdf files based on
  hierarchical document structure,'' in \emph{Proceedings of the 20th Annual
  Network \& Distributed System Security Symposium}.\hskip 1em plus 0.5em minus
  0.4em\relax Citeseer, 2013, pp. 1--16.

\bibitem{hidost}
``Hidost: Toolset for extracting document structures from pdf and swf files,''
  \url{https://github.com/srndic/hidost}.

\bibitem{hayes2020extensions}
J.~Hayes, ``Extensions and limitations of randomized smoothing for robustness
  guarantees,'' in \emph{Proceedings of the IEEE/CVF Conference on Computer
  Vision and Pattern Recognition Workshops}, 2020, pp. 786--787.

\bibitem{deng2009imagenet}
J.~Deng, W.~Dong, R.~Socher, L.-J. Li, K.~Li, and L.~Fei-Fei, ``Imagenet: A
  large-scale hierarchical image database,'' in \emph{2009 IEEE conference on
  computer vision and pattern recognition}.\hskip 1em plus 0.5em minus
  0.4em\relax Ieee, 2009, pp. 248--255.

\bibitem{kersting2001towards}
K.~Kersting and L.~D. Raedt, ``Towards combining inductive logic programming
  with bayesian networks,'' in \emph{International Conference on Inductive
  Logic Programming}.\hskip 1em plus 0.5em minus 0.4em\relax Springer, 2001,
  pp. 118--131.

\bibitem{taskar2007relational}
B.~Taskar, P.~Abbeel, M.-F. Wong, and D.~Koller, ``Relational markov
  networks,'' \emph{Introduction to statistical relational learning}, vol. 175,
  p. 200, 2007.

\bibitem{deng2014large}
J.~Deng, N.~Ding, Y.~Jia, A.~Frome, K.~Murphy, S.~Bengio, Y.~Li, H.~Neven, and
  H.~Adam, ``Large-scale object classification using label relation graphs,''
  in \emph{European conference on computer vision}.\hskip 1em plus 0.5em minus
  0.4em\relax Springer, 2014, pp. 48--64.

\bibitem{hu2016harnessing}
Z.~Hu, X.~Ma, Z.~Liu, E.~H. Hovy, and E.~P. Xing, ``Harnessing deep neural
  networks with logic rules,'' in \emph{ACL (1)}, 2016.

\bibitem{crane2012investigating}
R.~Crane and L.~McDowell, ``Investigating markov logic networks for collective
  classification.'' in \emph{ICAART (1)}, 2012, pp. 5--15.

\bibitem{domingos20071}
P.~Domingos and M.~Richardson, ``1 markov logic: A unifying framework for
  statistical relational learning,'' \emph{Statistical Relational Learning}, p.
  339, 2007.

\bibitem{singla2006entity}
P.~Singla and P.~Domingos, ``Entity resolution with markov logic,'' in
  \emph{Sixth International Conference on Data Mining (ICDM'06)}.\hskip 1em
  plus 0.5em minus 0.4em\relax IEEE, 2006, pp. 572--582.

\bibitem{yedidia2000generalized}
J.~S. Yedidia, W.~Freeman, and Y.~Weiss, ``Generalized belief propagation,''
  \emph{Advances in neural information processing systems}, vol.~13, 2000.

\bibitem{yedidia2003understanding}
J.~S. Yedidia, W.~T. Freeman, Y.~Weiss \emph{et~al.}, ``Understanding belief
  propagation and its generalizations,'' \emph{Exploring artificial
  intelligence in the new millennium}, vol.~8, no. 236-239, pp. 0018--9448,
  2003.

\bibitem{kahle2008junction}
D.~Kahle, T.~Savitsky, S.~Schnelle, and V.~Cevher, ``Junction tree algorithm,''
  \emph{Stat}, vol. 631, 2008.

\bibitem{venugopal2014scaling}
D.~Venugopal and V.~G. Gogate, ``Scaling-up importance sampling for markov
  logic networks,'' \emph{Advances in Neural Information Processing Systems},
  vol.~27, 2014.

\bibitem{scarselli2008graph}
F.~Scarselli, M.~Gori, A.~C. Tsoi, M.~Hagenbuchner, and G.~Monfardini, ``The
  graph neural network model,'' \emph{IEEE transactions on neural networks},
  vol.~20, no.~1, pp. 61--80, 2008.

\bibitem{bordes2013translating}
A.~Bordes, N.~Usunier, A.~Garcia-Duran, J.~Weston, and O.~Yakhnenko,
  ``Translating embeddings for modeling multi-relational data,'' \emph{Advances
  in neural information processing systems}, vol.~26, 2013.

\bibitem{kadlec2017knowledge}
R.~Kadlec, O.~Bajgar, and J.~Kleindienst, ``Knowledge base completion:
  Baselines strike back,'' \emph{ACL 2017}, p.~69, 2017.

\bibitem{sun2018rotate}
Z.~Sun, Z.-H. Deng, J.-Y. Nie, and J.~Tang, ``Rotate: Knowledge graph embedding
  by relational rotation in complex space,'' in \emph{International Conference
  on Learning Representations}, 2018.

\bibitem{xie2019embedding}
Y.~Xie, Z.~Xu, M.~S. Kankanhalli, K.~S. Meel, and H.~Soh, ``Embedding symbolic
  knowledge into deep networks,'' \emph{Advances in neural information
  processing systems}, vol.~32, 2019.

\bibitem{katz2017reluplex}
G.~Katz, C.~Barrett, D.~L. Dill, K.~Julian, and M.~J. Kochenderfer, ``Reluplex:
  An efficient smt solver for verifying deep neural networks,'' in
  \emph{International Conference on Computer Aided Verification}.\hskip 1em
  plus 0.5em minus 0.4em\relax Springer, 2017, pp. 97--117.

\bibitem{ehlers2017formal}
R.~Ehlers, ``Formal verification of piece-wise linear feed-forward neural
  networks,'' in \emph{International Symposium on Automated Technology for
  Verification and Analysis}.\hskip 1em plus 0.5em minus 0.4em\relax Springer,
  2017, pp. 269--286.

\bibitem{lomuscio2017approach}
A.~Lomuscio and L.~Maganti, ``An approach to reachability analysis for
  feed-forward relu neural networks,'' \emph{arXiv preprint arXiv:1706.07351},
  2017.

\bibitem{fischetti2017deep}
M.~Fischetti and J.~Jo, ``Deep neural networks as 0-1 mixed integer linear
  programs: A feasibility study,'' \emph{arXiv preprint arXiv:1712.06174},
  2017.

\bibitem{zhang2018efficient}
H.~Zhang, T.-W. Weng, P.-Y. Chen, C.-J. Hsieh, and L.~Daniel, ``Efficient
  neural network robustness certification with general activation functions,''
  in \emph{NeurIPS}, 2018.

\bibitem{raghunathan2018certified}
A.~Raghunathan, J.~Steinhardt, and P.~Liang, ``Certified defenses against
  adversarial examples,'' in \emph{International Conference on Learning
  Representations}, 2018.

\bibitem{raghunathan2018semidefinite}
A.~Raghunathan, J.~Steinhardt, and P.~S. Liang, ``Semidefinite relaxations for
  certifying robustness to adversarial examples,'' in \emph{NeurIPS}, 2018.

\bibitem{madry2018towards}
A.~Madry, A.~Makelov, L.~Schmidt, D.~Tsipras, and A.~Vladu, ``Towards deep
  learning models resistant to adversarial attacks,'' in \emph{International
  Conference on Learning Representations}, 2018.

\end{thebibliography}
\newpage
\appendix
\section{Appendix}
\subsection{Proof Details}
\label{sec:proof}

\subsubsection{Proof of~\Cref{le:gradientofelbo}}
\label{sec:lemmaproof}
\begin{proof}
\begin{footnotesize}
\begin{align*}
 \begin{aligned}
 \label{eq:opt}
 \centering
&\nabla_\theta \mathbb{E}_{Q_{\theta}(\mathcal{T})} \left( \sum_{f \in \mathcal{F}} w_ff(t_1,...,t_L) - \log Z(w) - \log Q_{\theta}(\mathcal{T}) \right)\\
&=\nabla_\theta \int\left( Q_{\theta}(\mathcal{T})\left( \sum_{f \in \mathcal{F}} w_ff(t_1,...,t_L) \right) -  Q_{\theta}(\mathcal{T})\log Q_{\theta}(\mathcal{T}) \right)\, d\bm{t}\\ 
&=\int Q_{\theta}(\mathcal{T})\nabla_\theta\log Q_{\theta}(\mathcal{T})\left( \sum_{f \in \mathcal{F}} w_ff(t_1,...,t_L) \right) \\ 
&\quad  - Q_{\theta}(\mathcal{T})\log Q_{\theta}(\mathcal{T})\nabla_\theta  \log Q_{\theta}(\mathcal{T}) - Q_{\theta}(\mathcal{T})\nabla_\theta\log Q_{\theta}(\mathcal{T})\, d\bm{t}\\
&=\mathbb{E}_{Q_{\theta}(\mathcal{T})}\left(\sum_{f \in \mathcal{F}} w_ff(t_1,...,t_L) -\log Q_{\theta}(\mathcal{T})-1\right)\nabla_\theta\log Q_{\theta}(\mathcal{T})
 \end{aligned}
 \end{align*}
 \end{footnotesize}
 
 Further, with the truth $\mathbb{E}_{Q_{\theta}(\mathcal{T})}\nabla_\theta\log Q_{\theta}(\mathcal{T})=0$, we can see the term $\left(\sum_{f \in \mathcal{F}} w_ff(t_1,...,t_L) - \log{Q_{\theta}(\mathcal{T})}-1\right)$ shown above can be shifted by any constant without changing the whole expectation, which just means we can ignore the $-1$ inside this term. 
 \end{proof}
 
 \subsubsection{Proof of~\Cref{thm:1nnreasoning}}
\label{sec:thmproof}
\begin{proof}
We only need to prove that for each type of the formula defined in~\Cref{eq:allowed_clause}, the truth value of it can be written in the form of $\operatorname{Neg}(\bm{a}\bm{t}^T+b)$ where $\bm{a}$ is a row vector with shape $1\times n$ and $b$ is a constant.

Then, for Type 1 formula, the truth value of it can be directly calculated by $\operatorname{Neg}\left(t_i-(t_j+t_k+...+t_l)\right)$;  and for Type 2 formula, the truth value can be calculated by $\operatorname{Neg}\left(t_i-(t_j+t_k+...+t_l)/m\right)$ where $m$ is the number of the appeared $t_j,t_k,...,t_l$ here; for Type 3 formula, the truth value can be calculated by $\operatorname{Neg}\left(-t_i+(t_j+t_k+...+t_l)/m\right)$; for Type 4 formula, the truth value can be calculated by $\operatorname{Neg}\left(-t_i+(t_j+t_k+...+t_l)-m+1\right)$.

 the proof still holds for the cases with negation on some predicates like $\lnot t_i$, which is equivalent to replacing the $t_i$ above with $1-t_i$.
\end{proof}

\renewcommand\arraystretch{1.15}
\begin{table*}[t]
    \centering
    \caption{\small Certified accuracy for~\name under different $\ell_2$ perturbation radii on the AwA2 dataset.}
    \label{tab:awa_ours}
\resizebox{\linewidth}{!}{
\begin{tabular}{c|c|cccccccccccccc}
\hline
\multirow{2}{*}{$\sigma$} & \multirow{2}{*}{Method} & \multirow{2}{*}{ACR} & \multicolumn{12}{c}{Certified Accuracy under Radius $r$} &      \\
 &                         &                      & 0.00 & 0.20 & 0.40 & 0.60 & 0.80 & 1.00 & 1.20 & 1.40 & 1.60 & 1.80 & 2.00 & 2.20 & 2.40 \\ \hline\hline
\multirow{3}{*}{0.25}
 & \name(Gaussian)     & \textbf{0.709} & \textbf{96.6} & \textbf{94.2} & \textbf{91.4} & 84.8 & 67.4 & 0.0 & 0.0 & 0.0 & 0.0 & 0.0 & 0.0 & 0.0 & 0.0     \\ 
 & \name(SmoothAdv)  & 0.707 & 95.4 & 92.4 & 89.8 & \textbf{85.4} & \textbf{75.0} & 0.0 & 0.0 & 0.0 & 0.0 & 0.0 & 0.0 & 0.0 & 0.0 \\
  &\name(Consistency)  & 0.693 & 95.0 & 92.0 & 87.2 & 83.0 & 70.0 & 0.0 & 0.0 & 0.0 & 0.0 & 0.0 & 0.0 & 0.0 & 0.0       \\\hline\hline
\multirow{3}{*}{0.50}  
  & \name(Gaussian)   & 1.114 & \textbf{91.2} & \textbf{88.2} & \textbf{84.2} & \textbf{78.8} & 71.2 & 66.4 & 56.8 & 46.8 & 34.6 & 0.0 & 0.0 & 0.0 & 0.0                   \\
  &\name(SmoothAdv)   & \textbf{1.141} & 88.2 & 85.2 & 80.8 & \textbf{78.8} & \textbf{73.4} & 67.6 & 62.2 & 54.8 & 43.2 & 0.0 & 0.0 & 0.0 & 0.0 \\
  & \name(Consistency)  & 1.138 & 87.8 & 84.6 & 80.0 & 76.8 & \textbf{73.4} & \textbf{68.4} & \textbf{63.2} & \textbf{56.2} & \textbf{44.0} & 0.0 & 0.0 & 0.0 & 0.0           \\
\hline\hline
\multirow{3}{*}{1.00} 
& \name(Gaussian)    & 2.092 & \textbf{87.0} & \textbf{85.2} & \textbf{84.0} & \textbf{82.0} & \textbf{80.4} & \textbf{78.2} & \textbf{75.6} & 71.2 & 68.0 & 64.4 & 61.0 & 57.0 & 52.8       \\ 
& \name(SmoothAdv)   & 2.087 & 85.0 & 83.0 & 81.6 & 79.6 & 76.6 & 75.0 & 73.2 & \textbf{71.4} & 68.0 & 64.8 & 59.2 & 56.0 & 53.8          \\
& \name(Consistency)  & \textbf{2.127} & 85.4 & 84.0 & 83.0 & 80.2 & 78.4 & 76.2 & 73.4 & 70.6 & \textbf{68.6} & \textbf{65.8 } & \textbf{61.8} & \textbf{59.4} & \textbf{56.0}         \\
\hline
\end{tabular}}
\end{table*}

\renewcommand\arraystretch{1.15}
\begin{table*}[htbp]
    \centering
    \caption{\small Certified Word accuracy for~\name under different $\ell_2$ perturbation radii on  Word50.}
    \label{tab:word_ours}
\resizebox{\linewidth}{!}{
\begin{tabular}{c|c|cccccccccccccc}
\hline
\multirow{2}{*}{$\sigma$} & \multirow{2}{*}{Method} & \multicolumn{1}{c}{\multirow{2}{*}{ACR}} & \multicolumn{12}{c}{Certified Accuracy under Radius $r$}      & \multicolumn{1}{c}{}     \\ &     & \multicolumn{1}{c}{}    & \multicolumn{1}{c}{0.00} & \multicolumn{1}{c}{0.10} & \multicolumn{1}{c}{0.20} & \multicolumn{1}{c}{0.30} & \multicolumn{1}{c}{0.40} & \multicolumn{1}{c}{0.50} & \multicolumn{1}{c}{0.60} & \multicolumn{1}{c}{0.70} & \multicolumn{1}{c}{0.80} & \multicolumn{1}{c}{0.90} & \multicolumn{1}{c}{1.00} & \multicolumn{1}{c}{1.10} & \multicolumn{1}{c}{1.20} \\ \hline\hline
\multirow{3}{*}{0.12}                  
 & \name(Gaussian)    & 0.360 & 94.8 & 89.6 & 82.6 & 75.8 & 62.0 & 0.0 & 0.0 & 0.0 & 0.0 & 0.0 & 0.0 & 0.0 & 0.0                 \\
 & \name(SmoothAdv)    & 0.371 & 93.2 & 89.4 & 85.0 & 79.4 & 67.8 & 0.0 & 0.0 & 0.0 & 0.0 & 0.0 & 0.0 & 0.0 & 0.0       \\
& \name(Consistency)   & \textbf{0.391} & \textbf{97.0} & \textbf{96.0} & \textbf{91.4} & \textbf{81.4} & \textbf{70.4} & 0.0 & 0.0 & 0.0 & 0.0 & 0.0 & 0.0 & 0.0 & 0.0  \\
\hline\hline
\multirow{3}{*}{0.25}                  
& \name(Gaussian)    & 0.624 & 96.0 & 92.4 & 88.0 & 82.2 & 75.6 & 68.8 & 58.0 & 48.0 & 37.8 & 25.6 & 0.0 & 0.0 & 0.0      \\ 
& \name(SmoothAdv)    & 0.577 & 91.6 & 86.4 & 81.0 & 74.6 & 69.2 & 63.0 & 53.6 & 45.4 & 35.6 & 24.0 & 0.0 & 0.0 & 0.0        \\
& \name(Consistency)    & \textbf{0.674} &\textbf{97.2} & \textbf{94.8} &\textbf{92.6} & \textbf{89.4 }& \textbf{81.8} & \textbf{73.6} & \textbf{64.4} & \textbf{55.2} & \textbf{43.6} & \textbf{30.8} & 0.0 & 0.0 & 0.0       \\
 \hline\hline
\multirow{3}{*}{0.50}
& \name(Gaussian)     & 0.671 & 87.0 & 83.0 & 77.2 & 73.4 & 68.2 & 60.6 & 54.2 & 48.2 & 40.6 & 34.0 & 28.0 & 20.6 & 15.0       \\
& \name(SmoothAdv)   & 0.690 & 85.2 & 82.0 & 77.4 & 71.6 & 66.6 & 60.8 & 54.2 & 48.0 & 42.6 & \textbf{36.6} & 29.4 & \textbf{24.0} & \textbf{18.0}       \\
& \name(Consistency)    & \textbf{0.697} & \textbf{87.6}& \textbf{84.4} & \textbf{78.4} & \textbf{73.6} & \textbf{69.0} & \textbf{63.0} & \textbf{56.6} & \textbf{50.0} & \textbf{44.0} & 36.4 & \textbf{30.0 }& 21.8 & 16.2    \\
\hline
\end{tabular}}
\end{table*}

\renewcommand\arraystretch{1.15}
\begin{table*}[htbp]
    \centering
    \caption{\small Certified Character accuracy for~\name under different $\ell_2$ perturbation radii  on  Word50.}
    \label{tab:character_ours}
\resizebox{\linewidth}{!}{
\begin{tabular}{c|c|cccccccccccccc}
\hline
\multirow{2}{*}{$\sigma$} & \multirow{2}{*}{Method} & \multicolumn{1}{c}{\multirow{2}{*}{ACR}} & \multicolumn{12}{c}{Certified Accuracy under Radius $r$}      & \multicolumn{1}{c}{}     \\ &     & \multicolumn{1}{c}{}    & \multicolumn{1}{c}{0.00} & \multicolumn{1}{c}{0.10} & \multicolumn{1}{c}{0.20} & \multicolumn{1}{c}{0.30} & \multicolumn{1}{c}{0.40} & \multicolumn{1}{c}{0.50} & \multicolumn{1}{c}{0.60} & \multicolumn{1}{c}{0.70} & \multicolumn{1}{c}{0.80} & \multicolumn{1}{c}{0.90} & \multicolumn{1}{c}{1.00} & \multicolumn{1}{c}{1.10} & \multicolumn{1}{c}{1.20} \\ \hline\hline
\multirow{3}{*}{0.12}        
 & \name(Gaussian)    & 0.306 & 86.6 & 80.6 & 71.8 & 60.4 & 50.0 & 0.0 & 0.0 & 0.0 & 0.0 & 0.0 & 0.0 & 0.0 & 0.0             \\
 & \name(SmoothAdv)   & \textbf{0.341} & \textbf{90.2} & \textbf{85.2} & \textbf{78.0} & \textbf{70.8} & \textbf{60.0} & 0.0 & 0.0 & 0.0 & 0.0 & 0.0 & 0.0 & 0.0 & 0.0        \\
 & \name(Consistency)    & 0.318 & 86.6 & 80.8 & 72.8 & 65.6 & 53.0 & 0.0 & 0.0 & 0.0 & 0.0 & 0.0 & 0.0 & 0.0 & 0.0       \\
\hline\hline
\multirow{3}{*}{0.25}   
& \name(Gaussian)    & 0.467 & 84.0 & 77.8 & 70.8 & 63.4 & 56.0 & 49.0 & 41.4 & 30.2 & 24.6 & 12.8 & 0.0 & 0.0 & 0.0      \\
& \name(SmoothAdv)    & \textbf{0.539} & 86.4 & 82.8 & 77.2 & 70.6 & \textbf{63.2} & \textbf{55.6} & \textbf{50.2} & \textbf{41.8}& \textbf{32.4} & \textbf{21.2} & 0.0 & 0.0 & 0.0       \\
& \name(Consistency)    & 0.522 & \textbf{87.6} & \textbf{83.2} & \textbf{77.4} & \textbf{73.0} & 62.8 & 54.4 & 46.2 & 37.0 & 29.0 & 17.8 & 0.0 & 0.0 & 0.0      \\
\hline\hline
\multirow{3}{*}{0.50}
& \name(Gaussian)    & 0.536 & 80.2 & \textbf{76.8} & \textbf{70.4} & 64.6 & \textbf{59.2} & 51.6 & \textbf{44.8} & \textbf{37.0} & 29.4 & \textbf{21.6} & 15.6 & 10.4 & 6.0        \\
& \name(SmoothAdv)    & 0.501 & 76.8 & 71.6 & 65.6 & 60.6 & 53.6 & 45.8 & 41.0 & 35.6 & 27.6 & 20.4 & 15.4 & 10.8 & \textbf{6.6}   \\
& \name(Consistency)    & \textbf{0.539} & \textbf{80.6} & 75.6 & 70.2 & \textbf{65.4} & \textbf{59.2} & \textbf{53.0} & 44.6 & 36.2 & \textbf{29.6} & \textbf{21.6} & \textbf{17.2} & \textbf{11.6} & 6.4      \\
\hline
\end{tabular}}
\end{table*}

\renewcommand\arraystretch{1.15}
    \begin{table*}[htbp]
        \centering
        \caption{\small Certified accuracy under different $\ell_2$ perturbation radii and sigma on  GTSRB.}
        \label{tab:stop_sign_full}
    \resizebox{\linewidth}{!}{
    \begin{tabular}{c|c|cccccccccccccc}
    \hline
    \multirow{2}{*}{$\sigma$} & \multirow{2}{*}{Method} & \multicolumn{1}{c}{\multirow{2}{*}{ACR}} & \multicolumn{12}{c}{Certified Accuracy under Radius $r$}      & \multicolumn{1}{c}{}     \\ &     & \multicolumn{1}{c}{}    & \multicolumn{1}{c}{0.00} & \multicolumn{1}{c}{0.10} & \multicolumn{1}{c}{0.20} & \multicolumn{1}{c}{0.30} & \multicolumn{1}{c}{0.40} & \multicolumn{1}{c}{0.50} & \multicolumn{1}{c}{0.60} & \multicolumn{1}{c}{0.70} & \multicolumn{1}{c}{0.80} & \multicolumn{1}{c}{0.90} & \multicolumn{1}{c}{1.00} & \multicolumn{1}{c}{1.10} & \multicolumn{1}{c}{1.20} \\ \hline\hline
    \multirow{8}{*}{0.12}& Gaussian& 0.410 & 97.9 & 96.5 & 92.6 & 86.8 & 79.8 & 0.0 & 0.0 & 0.0 & 0.0 & 0.0 & 0.0 & 0.0 & 0.0   \\
    & SWEEN &0.417 & 99.2 & 97.1 & 94.7 & 87.9 & 82.3 & 0.0 & 0.0 & 0.0 & 0.0 & 0.0 & 0.0 & 0.0 & 0.0       \\
    & SmoothAdv  & 0.410 & 97.1 & 95.9 & 91.8 & 86.6 & 81.3 & 0.0 & 0.0 & 0.0 & 0.0 & 0.0 & 0.0 & 0.0 & 0.0       \\
     & Consistency    & 0.422 & 99.4 & 98.8 & 95.3 & 90.7 & 83.1 & 0.0 & 0.0 & 0.0 & 0.0 & 0.0 & 0.0 & 0.0 & 0.0       \\
     & {MultiTask}    & {0.402} & {96.5} & {94.4} & {90.3} & {85.8} & {78.4} & {0.0} & {0.0} & {0.0} & {0.0} & {0.0} & {0.0} & {0.0} & {0.0}      \\ \cline{2-16}
     & \name(Gaussian)   & 0.414 & 99.0 & 98.1 & 92.8 & 87.2 & 82.1 & 0.0 & 0.0 & 0.0 & 0.0 & 0.0 & 0.0 & 0.0 & 0.0      \\
     & \name(SmoothAdv)   & 0.421 & 99.0 & 98.6 & 94.4 & 90.5 & 81.9 & 0.0 & 0.0 & 0.0 & 0.0 & 0.0 & 0.0 & 0.0 & 0.0      \\
     & \name(Consistency)   & \textbf{0.425} & \textbf{99.6} & \textbf{99.2} & \textbf{96.7} & \textbf{91.2} & \textbf{84.4} & 0.0 & 0.0 & 0.0 & 0.0 & 0.0 & 0.0 & 0.0 & 0.0        \\\hline\hline
    \multirow{8}{*}{0.25}                  & Gaussian & 0.742 & 96.5 & 94.7 & 90.3 & 85.6 & 82.5 & 78.6 & 74.3 & 68.7 & 63.2 & 55.3 & 0.0 & 0.0 & 0.0 \\
    & SWEEN & 0.750 & 97.7 & 94.2 & 90.7 & 86.6 & 82.9 & 78.8 & 74.1 & 69.1 & 65.8 & 58.4 & 0.0 & 0.0 & 0.0 \\
    & SmoothAdv & 0.754 & 93.8 & 91.6 & 90.1 & 86.6 & 82.9 & 78.6 & \textbf{75.3} & 71.0 & 66.9 & \textbf{63.0} & 0.0 & 0.0 & 0.0 \\
    & Consistency & 0.755 & 95.5 & 93.8 & 91.8 & 87.7 & 83.3 & 78.0 & 74.3 & 71.0 & 65.6 & 59.9 & 0.0 & 0.0 & 0.0\\ 
     & {MultiTask}    & {0.732} & {95.7} & {93.4} & {90.3} & {85.8} & {81.3} & {76.3} & {71.8} & {66.9} & {62.8} & {54.3} & {0.0} & {0.0} & {0.0}    \\\cline{2-16}
     & \name(Gaussian)   & 0.754 & \textbf{97.9} & \textbf{95.1} & 91.2 & 86.2 & 82.9 & 79.0 & 75.1 & 70.4 & 65.2 & 58.6 & 0.0 & 0.0 & 0.0      \\
     & \name(SmoothAdv)   & \textbf{0.762} & 95.5 & 93.8 & 91.4 & 87.0 & \textbf{83.7} & \textbf{79.4} & 75.1 & \textbf{72.0} & \textbf{67.9} & \textbf{63.0} & 0.0 & 0.0 & 0.0    \\
     & \name(Consistency)   & 0.761 & 96.9 & 94.7 & \textbf{92.2} & \textbf{87.7} & 82.9 & 78.4 & \textbf{75.3} & 71.8 & 66.9 & 60.5 & 0.0 & 0.0 & 0.0     \\\hline\hline
    \multirow{8}{*}{0.50}& Gaussian& 1.058 & 88.1 & 86.2 & 82.9 & 78.6 & 75.3 & 71.6 & 70.0 & 64.6 & 60.5 & 57.0 & 53.3 & 50.4 & 47.7   \\
    & SWEEN & 1.092 & 87.9 & 86.0 & \textbf{83.3} & 79.6 & 75.5 & 72.8 & 69.5 & 66.3 & 63.2 & 58.6 & 55.8 & 52.5 & 49.6     \\ 
     & SmoothAdv & 1.079 & 79.8 & 78.8 & 76.1 & 73.7 & 71.2 & 68.7 & 66.3 & 63.2 & 61.1 & 59.3 & 56.8 & 53.9 & 51.6 \\
     & Consistency & 1.098 & 82.7 & 81.1 & 79.0 & 76.5 & 74.9 & 73.3 & \textbf{70.8} & 67.5 & \textbf{64.6} & 60.5 & 57.4 & 54.5 & 51.2    \\
      & {MultiTask}    &{1.082} & {89.5} & {87.2} & {82.5} & {78.4} & {75.5} & {72.8} & {69.8} & {67.3} & {62.8} & {57.8} & {55.3} & {51.4} & {48.6} \\\cline{2-16}
     & \name(Gaussian)   &1.092 & \textbf{89.5} & \textbf{86.8} & 82.9 & \textbf{80.2} & {76.3} & 73.3 & 69.3 & 66.3 & 62.6 & 58.6 & 55.3 & 52.7 & 49.4       \\
     & \name(SmoothAdv)   & 1.111 & 87.9 & 84.6 & 81.7 & 78.2 & 74.7 & 72.0 & 70.0 & 65.8 & 63.0 & \textbf{60.9} & 57.2 & 54.3 & 51.6     \\
     & \name(Consistency)   & \textbf{1.117} & 88.5 & 86.0 & 81.9 & 78.6 & \textbf{76.5} & \textbf{73.5} & 70.6 & \textbf{67.7} & 64.0 & 60.7 & \textbf{57.8} & \textbf{55.1} & \textbf{51.9}        \\\hline
    \end{tabular}}
    \end{table*}
    
\subsection{Experiment Details}
\label{sec:exp_details}

\subsubsection{Training details} 
\label{sec:exp_baseline}
For AwA2, the sensor is initialized with the weight pretrained on ImageNet and finetuned with a learning rate $0.001$ for $30$ epochs; the batch size is also set to $256$. For Word50, the sensor is trained with $90$ epochs, and the initial learning rate is set to $0.01$ and will be decayed by $0.1$ at $30$-th and $60$-th epoch, and the batch size is set to $128$. For GTSRB, the model is trained with $150$ epochs, and the initial learning rate is set to $0.01$ and will be decayed by $0.1$ at $50$-th and $100$-th epoch, the batch size is set to $200$. For the PDF malware dataset, the sensor is trained with $90$ epochs, and the initial learning rate is set to $0.05$ and will be decayed by $0.1$ at $30$-th and $60$-th epoch, and the batch size is set to $128$. for all image datasets, we balance the number of training images from each class during the training. {For MultiTask, we add more classification heads in the main sensor and train it together with other knowledge tasks under Gaussian noise; the loss is defined as the mean of the classification loss for each task, the training epoch is set to $150$, the initial learning rate is still set the same for each dataset and will be decayed by $0.1$ at $50$-th and $100$-th epoch.} 

The number of the base models in SWEEN is fixed to $6$ for all experiments. And for the training with SmoothAdv, the $\epsilon$ is set to $255$ and the $m$ is set to $2$ under all sigmas on the datasets AwA2 and Word50; while on GTSRB, the $\epsilon$ is set to $127$ under $\sigma = 0.12,0.25$ and is set to $255$ under $\sigma = 0.50$. For the training with Consistency, the $\lambda$ is set to $10$ and the $m$ is set to $2$ on AwA2 under all sigmas; while on Word50, the $m$ is set to $2$ under all sigmas, and the $\lambda$ is set to $10$ for $\sigma = 0.12,0.25$ and is set to $5$ for $\sigma = 0.50$; for GTSRB, the $m$ and the $\lambda$ are set to $2$ and $5$, respectively under all sigmas. 

{
\subsubsection{Certification Procedure}
\label{sec:certification_alg}
The whole certification process is provided in ~\Cref{alg:rs_certification} following~\cite{cohen2019certified}, the auxiliary function $\textsc{SampleUnderNoise}$ is shown in~\Cref{alg:sample} and the $\textsc{LowerConfBound}(k,n,1-\alpha)$ is a function which returns a one-sided $(1-\alpha)$ lower confidence bound $\underline{p}$ for the Binomial parameter $p$ given $k \sim \operatorname{Binomial}(n, p)$.
}

\begin{algorithm}[t]
\caption{{$\textsc{SampleUnderNoise}(f, x, n, \sigma)$.}}
\label{alg:sample}
\begin{algorithmic}[1]
\renewcommand{\algorithmicrequire}{\textbf{Input:}}
 \renewcommand{\algorithmicensure}{\textbf{Output:}}
  \REQUIRE Base classifier $f$, clean input image $x$, the number of smoothing noise $n$, smoothing noise magnitude $\sigma$.
  \ENSURE A vector of class counts.
\STATE $\texttt{counts} \gets [0,0,...,0]$
\FOR{$i = 1$ to $n$}
\STATE $x_{rs} \gets x + \mathcal{N}\left(\mathbf{0}, \sigma^{2} \mathbf{I}\right)$
\STATE $y \gets f(x_{rs})$
\STATE $\texttt{counts}[y] += 1$
\ENDFOR
  \STATE \textbf{return} \texttt{counts}
 \end{algorithmic}
 \end{algorithm}
 
 \begin{algorithm}[t]
   \caption{{Certification Procedure for Randomized Smoothing.}}
   \label{alg:rs_certification}
   \begin{algorithmic}[1]
\renewcommand{\algorithmicrequire}{\textbf{Input:}}
 \renewcommand{\algorithmicensure}{\textbf{Output:}}
  \REQUIRE The magnitude of the smoothing noise $\sigma$, the magnitude of the local smoothing noise $\sigma'$, the base classifier $f$, the number of the smoothing noise for selection $n_0$, the number of the smoothing noise for estimation $n$, the certification confidence $(1-\alpha)$. 
  \ENSURE Certified prediction and its robust radius.
   \STATE $\texttt{counts0} \leftarrow \textsc{SampleUnderNoise}(f, x, n_0, \sigma)$
   \STATE $\hat{c}_A \leftarrow$ top index in \texttt{counts0}
   \STATE $\texttt{counts} \leftarrow \textsc{SampleUnderNoise}(f, x, n, \sigma)$
   \STATE $\underline{p_A} \leftarrow \textsc{LowerConfBound}$($\texttt{counts}[\hat{c}_A]$, $n$, $1 - \alpha$) 
   \IF{ $\underline{p_A} > \frac{1}{2}$}
   \STATE \textbf{return} $\hat{c}_A$ and radius $\sigma \, \Phi^{-1}(\underline{p_A})$ 
   \ELSE 
   \STATE \textbf{return} ABSTAIN
   \ENDIF
\end{algorithmic}
\end{algorithm}

\subsubsection{Certification details} 
\label{sec:certification_detail}

\textbf{Word50 certification details:}
The training, validation, and test sets contain $10,000$, $2,000$ and $2,000$ different word images, respectively. We randomly select $10$ images for each word from the test dataset for certification, and the total number of certified images is $500$ following the standard evaluation setting. All the results are certified with  $N=100,000$ samples of smoothing noise, and the confidence of the certification is set to $99.9\%$. We test our method on three levels of smoothing noise  $\sigma = 0.12, 0.25, 0.50$, and the $\eta$ is set to $0.6,0.9,1.0$, respectively.

\textbf{GTSRB certification details:}
The whole dataset contains $14880$ training samples, $972$ validation samples, and $3888$ testing samples. We randomly pick one out of every eight from the test dataset for certification, and following the standard setting~\cite{cohen2019certified}, we certify these $486$ images with confidence $99.9\%$, and all the results are certified with  $N=100,000$ samples of smoothing noise. We test our method on three levels of smoothing noise with $\sigma = 0.12,0.25,0.50$, and the $\eta$ is set to $0.10,0.15,0.25$, respectively.
    
\textbf{PDF Malware certification details:}
We split the whole Contagio dataset into $70\%$ train set and $30\%$ test set following~\cite{chen2020training}. In specific, the number of malicious PDFs for training and testing is $6,896$ and $3,448$, respectively, while the number of benign PDFs for training and testing is $6,296$ and $2,698$, respectively. We select $10\%$ images, namely, $615$ PDFs, from the test dataset for certification. 
Since the input extracted features from the PDF are binary, the certification is conducted based on $\ell_0$-norm. All the results are certified with  $N=100,000$ samples of smoothing noise, and the confidence of the certification is set to $99.9\%$. We test our method on three levels of retaining probability $\alpha = 0.80, 0.85, 0.90$, and the $\eta$ is set to $0.09,0.10,0.05$, respectively. All the sensors for both our method and the baselines are trained with  Bernoulli augmentation following~\cite{lee2019tight}, where each feature value will be replaced with a random value ~\{0,1\} with probability $1-\alpha$. 

\subsubsection{Detailed knowledge used in GTSRB}
\label{sec:gtsrb_details}
We demonstrate the $20$ manually constructed attributes here. Some of them are adapted from~\cite{gurel2021knowledge}. The $12$ permissive attributes are as follows:~\emph{``Octagon'', ``Square'', ``Blank Triangle'', ``Inverse Triangle'', ``Red Circle'', ``Gray Circle'', ``Blank Circle'', ``Digit 20'', ``Digit 50'', ``Digit 120'', ``Left Arrow'', ``Right Arrow''}. The $8$ preventative attributes are as follows:~\emph{``Red Hollow Circle'', ``Blue Filled Circle'', ``Circle'', ``Blank Content'', ``Digit Content'', ``Filled Content'', ``Symmetric'', ``Polygon''}.

The inclusion relations are shown as follows: 1. each of the attributes~\emph{``Octagon'', ``Square'', ``Blank Triangle'', ``Inverse Triangle''} would imply~\emph{``Polygon''}; 2. each of the attributes~\emph{``Red Circle'', ``Gray Circle'', ``Blank Circle'', ``Red Hollow Circle'', ```Blue Filled Circle''} would imply~\emph{``Circle''}; 3. each of the attributes~\emph{``Blank Triangle'', ``Blank Circle''} would imply~\emph{``Blank Content''}; 4. each of the attributes~\emph{``Digit 20'', ``Digit 50'', ``Digit 120''} would imply~\emph{``Digit Content''}. For better message passing, the edge for the inclusion relation on the graph is directed from the attribute to its corresponding implied attribute.

\subsubsection{The knowledge and the reasoning  details in PDF malware}
\label{sec:pdf_details}
The chosen $6$ malicious traces are \emph{``/Root/OpenAction'', ``/Root/OpenAction/S'', ``/Root/OpenAction/JS'', ``/Root/OpenAction/JS/Filter'', ``/Root/OpenAction/JS/Length'', ``/Root/OpenAction/Type''}. The chosen $8$ benign traces are \emph{``/Root/Metadata'', ``/Root/Metadata/Length'', ``/Root/Metadata/Subtype'', ``/Root/Metadata/Type'', ``/Root/Pages/Contents'', ``/Root/Pages/Contents/Filter'', ``/Root/Pages/Contents/Length'', ``/Root/Pages/CropBox''}.

For the reasoning part, notice we construct the formula like $t_i\implies \lnot t_j$. Then, instead of directly connecting the edge between the node representing $t_i$ and the node representing $t_j$, we construct an auxiliary node representing the predicate $\lnot t_j$ and choose to connect it with the node representing $t_i$ for better message passing. And the exclusion formula is naturally built for $t_j$ and $\lnot t_j$.

\begin{figure}[t]
\centering
\includegraphics[width=0.5\textwidth]{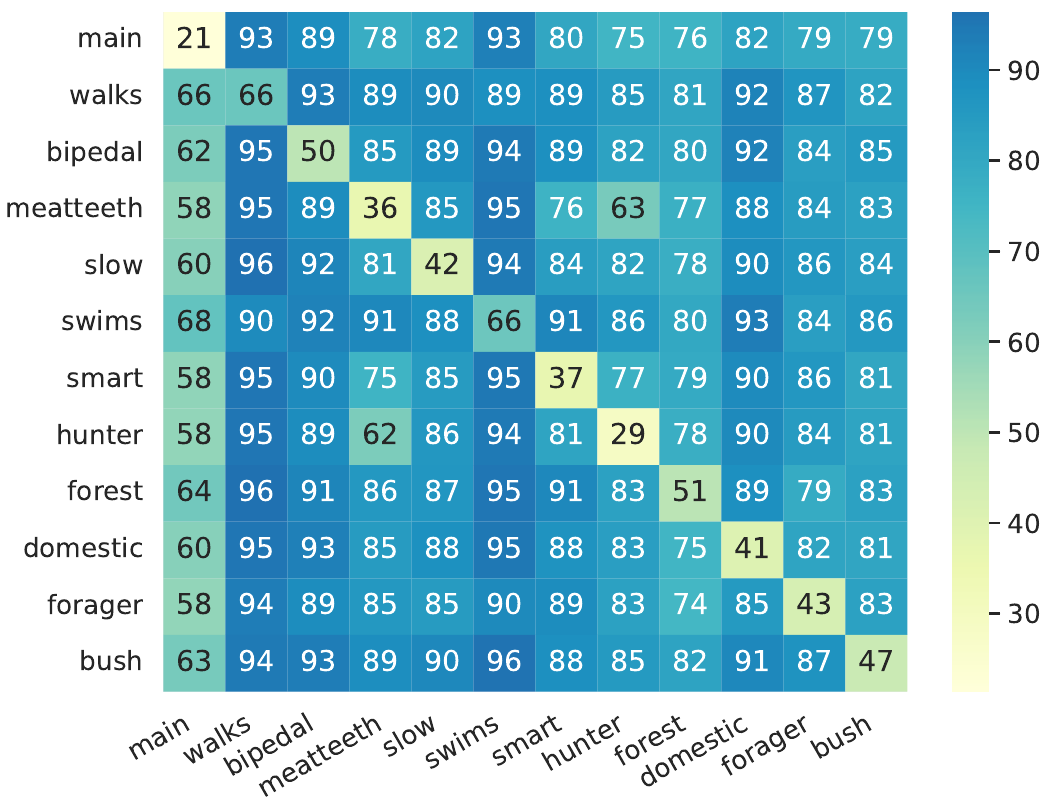}
\caption{The heatmap for the attack transferability between $13$ sensors. The number in the cell $(i,j)$ represents the empirical robust accuracy ($\%$) of $j$th sensor when tested with the adversarial attacks against the  $i$th sensor.}
\label{fig:heatmap}
\end{figure} 
    
\subsection{Detailed Experiment Results for~\name}
\label{sec:exp_for_different_main}
We demonstrate the detailed experiment results for our method~\name with different main sensors. For simplicity, we use~\name(Gaussian) to indicate the main sensor is trained with vanilla Gaussian augmentation~\cite{cohen2019certified}; use~\name(SmoothAdv) to indicate the main sensor is trained with adversarial training~\cite{salman2019provably}; and use~\name(Consistency) to indicate the main sensor is trained with consistency regularization~\cite{jeong2020consistency}. Notice that all the knowledge sensors are trained with vanilla Gaussian augmentation~\cite{cohen2019certified}. The detailed results on AwA are shown in~\Cref{tab:awa_ours}; the results for word classification and character classification are shown in~\Cref{tab:word_ours} and~\Cref{tab:character_ours}, respectively. 

 The full results for GTSRB are shown in~\Cref{tab:stop_sign_full}; while for the PDF malware dataset, we also report the Median Certified Radius (MCR) as a reference, and the full results are shown in~\Cref{tab:pdfmalware_full}.

\renewcommand\arraystretch{1.15}
\begin{table*}[htbp]
    \centering
    \caption{\small Certified accuracy under different retaining probability $\alpha$ and $\ell_0$ perturbation radii on the PDF malware dataset.}
    \label{tab:pdfmalware_full}
\resizebox{0.8\linewidth}{!}{
\begin{tabular}{c|c|ccccccccccc}
\hline
\multirow{2}{*}{$\alpha$} & \multirow{2}{*}{Method} & \multirow{2}{*}{ACR/MCR} & \multicolumn{10}{c}{Certified Accuracy under Radius $r$} \\
 & & & 0   & 1   & 2   & 3   & 4   & 5   & 6   & 7   & 8   & 9  \\ \hline
\multirow{4}{*}{0.90}   & Lee et al.~\cite{lee2019tight} & 3.008/3 & \textbf{99.8}  & 99.0 & 96.1 & 77.9   &  27.8    &  0.0  &  0.0    &    0.0 &  0.0  &   0.0  \\
& SWEEN & 3.159/3 & \textbf{99.8}  & 99.0 & \textbf{97.7} & 81.1   &  38.0    &  0.0  &  0.0    &    0.0 &  0.0  &  0.0   \\
& {MultiTask} & {3.341/4} & {99.7}  & {99.0} & {97.2} & {81.5}   &  {56.4}    &  {0.0}  &  {0.0}    &    {0.0} &  {0.0}  &  {0.0}   \\
& \name & \textbf{3.506/4} & 99.5  & \textbf{99.3} & 96.9 & \textbf{85.5}   &  \textbf{68.8}    &  0.0  &  0.0    &    0.0 &  0.0  &0.0  \\ \hline\hline
\multirow{4}{*}{0.85}   & Lee et al.~\cite{lee2019tight} & 3.842/4 & \textbf{99.7}  & 98.5 & 96.1 & 80.0   &  53.5    &  43.7  &  12.4    &    0.0 &  0.0  &0.0  \\
& SWEEN & 4.367/5 & \textbf{99.7}  & \textbf{98.9} & \textbf{96.4} & 82.2   &  68.6    &  65.4  & 22.3    &    0.0 &  0.0  &0.0 \\
& {MultiTask} & {4.626/6} & {99.7}  & {99.0} & {96.7} & {82.8}   &  {66.3}    &  {65.0}  &  {52.7}    &    {0.0} &  {0.0}  &  {0.0}   \\
& \name & \textbf{4.954/6} & 99.5  & \textbf{98.9} & 91.1 & \textbf{85.4}   &  \textbf{79.3}    &  \textbf{77.4}  &  \textbf{63.4}    &    {0.0} &  {0.0} & {0.0}  \\ \hline\hline
\multirow{4}{*}{0.80}   & Lee et al.~\cite{lee2019tight} & 4.945/5 & \textbf{99.5}  & 98.7 & 94.8 & 80.0   &  80.0    &  68.0  &  46.5    &    15.1 &  5.7  & 5.7\\
& SWEEN & 5.259/6 & \textbf{99.5}  & \textbf{98.9} & 95.8 & 80.7   &  80.3    &  72.5  &  57.2    &   22.6 &  8.9  & 8.9 \\
& {MultiTask} & {5.624/7} & {99.3}  & {98.7} & {96.1} & {81.8}   &  {80.5}    &  {72.7}  &  {59.0}    &    {53.8} &  {9.9}  &  {9.9}   \\
& \name & \textbf{5.789/7} & 99.2  & 98.4 & \textbf{96.6} & \textbf{84.2}   &  \textbf{84.2}    &  \textbf{74.5}  &  \textbf{59.5}    &    \textbf{54.5} &  \textbf{13.5}  & \textbf{13.5} \\ \hline
\end{tabular}}
\end{table*}

\renewcommand\arraystretch{1.15}
\begin{table*}[t]
    \centering
    \caption{{\small Certified accuracy for the method~\emph{SensingLinear} on the AwA2 dataset under different $\ell_2$ perturbation radii, and the used main sensor is indicated in the bracket.}}
    \label{tab:n-way}
\resizebox{\linewidth}{!}{
\begin{tabular}{c|c|cccccccccccccc}
\hline
\multirow{2}{*}{$\sigma$} & \multirow{2}{*}{Method} & \multirow{2}{*}{ACR} & \multicolumn{12}{c}{Certified Accuracy under Radius $r$} &      \\
 &                         &                      & 0.00 & 0.20 & 0.40 & 0.60 & 0.80 & 1.00 & 1.20 & 1.40 & 1.60 & 1.80 & 2.00 & 2.20 & 2.40 \\ \hline\hline
\multirow{4}{*}{0.25}
 & SensingLinear(Gaussian)     & 0.593 & 79.8 & 78.2 & 76.2 & 71.0 & 58.0 & 0.0 & 0.0 & 0.0 & 0.0 & 0.0 & 0.0 & 0.0 & 0.0    \\ 
 & SensingLinear(SmoothAdv)  & 0.596 & 80.2 & 78.4 & 76.6 & 71.0 & 60.4 & 0.0 & 0.0 & 0.0 & 0.0 & 0.0 & 0.0 & 0.0 & 0.0 \\
  &SensingLinear(Consistency)  &0.596 & 79.4 & 78.2 & 76.4 & 72.2 & 60.2 & 0.0 & 0.0 & 0.0 & 0.0 & 0.0 & 0.0 & 0.0 & 0.0       \\
    &\name  & \textbf{0.709} & \textbf{96.6} & \textbf{94.2} & \textbf{91.4} & \textbf{85.4} & \textbf{75.0} & 0.0 & 0.0 & 0.0 & 0.0 & 0.0 & 0.0 & 0.0 & 0.0    \\
  \hline\hline
\multirow{4}{*}{0.50}  
  & SensingLinear(Gaussian)   & 0.842 & 69.6 & 67.6 & 63.2 & 58.2 & 53.4 & 49.4 & 42.4 & 36.8 & 27.2 & 0.0 & 0.0 & 0.0 & 0.0                  \\
  &SensingLinear(SmoothAdv)   & 0.865 & 70.0 & 67.6 & 64.2 & 59.6 & 55.2 & 50.4 & 45.2 & 40.0 & 31.6 & 0.0 & 0.0 & 0.0 & 0.0 \\
  & SensingLinear(Consistency)  & 0.877 & 69.8 & 68.0 & 65.4 & 60.0 & 56.0 & 51.0 & 45.6 & 40.4 & 31.6 & 0.0 & 0.0 & 0.0 & 0.0         \\
      &\name  &\textbf{1.141} & \textbf{91.2} & \textbf{88.2} & \textbf{84.2} & \textbf{78.8} & \textbf{73.4} & \textbf{68.4} & \textbf{63.2} & \textbf{56.2} & \textbf{44.0} & 0.0 & 0.0 & 0.0 & 0.0      \\
\hline\hline
\multirow{4}{*}{1.00} 
& SensingLinear(Gaussian)    & 1.192 & 51.6 & 49.8 & 48.4 & 46.8 & 46.0 & 45.0 & 42.0 & 40.0 & 38.2 & 36.0 & 34.0 & 31.2 & 29.2      \\ 
& SensingLinear(SmoothAdv)   & 1.265 & 52.8 & 51.6 & 50.2 & 48.4 & 47.4 & 46.4 & 45.0 & 42.2 & 40.0 & 37.8 & 36.0 & 34.2 & 32.0    \\
& SensingLinear(Consistency)  & 1.270 & 52.2 & 51.4 & 50.2 & 48.8 & 47.4 & 46.4 & 44.0 & 43.2 & 40.6 & 38.6 & 36.4 & 34.2 & 32.4        \\
    &\name  & \textbf{2.127} & \textbf{87.0} & \textbf{85.2} & \textbf{84.0} & \textbf{82.0} & \textbf{80.4} & \textbf{78.2} & \textbf{75.6} & \textbf{71.4} & \textbf{68.6} & \textbf{65.8} & \textbf{61.8} & \textbf{59.4} & \textbf{56.0}            \\
\hline
\end{tabular}}
\end{table*}

\renewcommand\arraystretch{1.1}
\begin{table*}[ht]
\centering
    \caption{{\small Certified accuracy for SWEEN with different number of base models under smoothing noise level $\sigma = 0.50$.}}
    \label{tab:ensemble_diff_num}
    \vspace{-2mm}
\resizebox{0.8\linewidth}{!}{
\begin{tabular}{c|cccccccccc}
\hline
 \multirow{2}{*}{\# base models} & \multirow{2}{*}{ACR} &\multicolumn{8}{c}{Certified Accuracy under $\ell_2$ Radius $r$}      & \multicolumn{1}{c}{}     \\  & &  \multicolumn{1}{c}{0.00}   & \multicolumn{1}{c}{0.20} & \multicolumn{1}{c}{0.40} & \multicolumn{1}{c}{0.60} & \multicolumn{1}{c}{0.80} & \multicolumn{1}{c}{1.00} & \multicolumn{1}{c}{1.20} & \multicolumn{1}{c}{1.40} & \multicolumn{1}{c}{1.60}  \\ \hline\hline
$m=3$ & 0.846 & \textbf{76.8} & 72.8 & 66.6 & 60.2 & 53.8 & 46.4 & 39.0 & 34.4 & 22.4     \\
 $m=6$ &0.854 & 76.4 & \textbf{73.8} & 67.8 & 60.4 & 53.6 & 47.4 & 39.6 & 34.6 & 22.4            \\
$m=10$   & 0.856 & 76.4 & 73.4 & \textbf{68.2} & 59.8 & 53.6 & 47.8 & 40.6 & 34.6 & \textbf{23.6} \\
  $m=15$     & 0.859 & 76.4 & 73.0 & \textbf{68.2} & 59.4 & 53.4 & 47.8 & 40.4 & 35.0 & \textbf{23.6}            \\
        $m=20$     & 0.860 & 76.6 & 72.8 & \textbf{68.2} & 59.4 & 53.6 & 48.0 & 40.6 & 35.0 & 23.4       \\
  $m=30$  & \textbf{0.863} & \textbf{76.8} & 73.2 & 67.4 & \textbf{60.8} & \textbf{54.4} & \textbf{48.2} & \textbf{41.2} & \textbf{35.2} & 23.4       \\ \hline
\end{tabular}}
\end{table*}

\renewcommand\arraystretch{1.15}
\begin{table*}[htbp]
    \centering
    \caption{\small The empirical robust accuracy of different methods for AwA2, Word50 and GTSRB under $\ell_{\infty}$ metric.}
    \label{tab:attack_inf_all}
\resizebox{0.8\linewidth}{!}{
\begin{tabular}{c|cccc|cccc|cccc}
\hline
\multirow{3}{*}{Method} & \multicolumn{4}{c|}{AwA2}& \multicolumn{4}{c|}{Word50}& \multicolumn{4}{c}{GTSRB}\\ \cline{2-13} 
& \multicolumn{1}{c|}{\multirow{2}{*}{$\sigma$}} & \multicolumn{3}{c|}{$\epsilon$} & \multicolumn{1}{c|}{\multirow{2}{*}{$\sigma$}} & \multicolumn{3}{c|}{$\epsilon$} & \multicolumn{1}{c|}{\multirow{2}{*}{$\sigma$}} & \multicolumn{3}{c}{$\epsilon$} \\ 
& \multicolumn{1}{c|}{}& 2/255          & 4/255          & 8/255          & \multicolumn{1}{c|}{}                                       & 2/255  & 4/255  & 8/255         & \multicolumn{1}{c|}{}& 2/255  & 4/255  & 8/255        \\ \hline
Gaussian& \multicolumn{1}{c|}{\multirow{6}{*}{0.25}}&  42.4    &   8.6     &   0.0         & \multicolumn{1}{c|}{\multirow{6}{*}{0.12}}  &  11.4 & 1.4 & 0.0          & \multicolumn{1}{c|}{\multirow{6}{*}{0.12}}                  & 89.9 & 73.7 & 47.3\\
SWEEN& \multicolumn{1}{c|}{}&  45.8   &   10.6    &  0.4        & \multicolumn{1}{c|}{}&  30.4 & 13.4 & 2.6       & \multicolumn{1}{c|}{} & 94.2 &\textbf{84.2} &65.6\\
SmoothAdv& \multicolumn{1}{c|}{} &  53.6   &   23.0    &  1.6       & \multicolumn{1}{c|}{}     & 32.4 & 10.4 & 1.2         & \multicolumn{1}{c|}{}&87.2&74.7&56.6\\
Consistency& \multicolumn{1}{c|}{}&  45.6   &   13.6    &  0.4        & \multicolumn{1}{c|}{} & 22.6 & 5.6 & 0.0  & \multicolumn{1}{c|}{}&92.4& 79.0  & 57.2\\
MultiTask& \multicolumn{1}{c|}{}&  46.8   &   18.2    &  0.8        & \multicolumn{1}{c|}{} & 16.6 & 2.4 & 0.2  & \multicolumn{1}{c|}{}& 87.7 & 70.2  & 41.4\\
\name& \multicolumn{1}{c|}{} & \textbf{80.6}    &  \textbf{32.6}   &   \textbf{3.4}       & \multicolumn{1}{c|}{} & \textbf{87.0} & \textbf{83.0} & \textbf{81.2}         & \multicolumn{1}{c|}{}&        \textbf{94.7}       & 84.0 &   \textbf{68.7}          \\ \hline\hline

Gaussian& \multicolumn{1}{c|}{\multirow{6}{*}{0.50}}& 47.0    &   18.2     &   1.6         & \multicolumn{1}{c|}{\multirow{6}{*}{0.25}}& 14.4 & 2.6 & 0.0        & \multicolumn{1}{c|}{\multirow{6}{*}{0.25}}                  &       86.2        &     75.1         &     51.9         \\
SWEEN& \multicolumn{1}{c|}{}&  47.6   &   21.6    &  2.4         & \multicolumn{1}{c|}{} &  30.0 & 18.2 & 2.8        & \multicolumn{1}{c|}{} &         87.4      &   76.7           &    58.4          \\
SmoothAdv& \multicolumn{1}{c|}{}&  48.6   &   25.8    &  3.4      & \multicolumn{1}{c|}{}&  30.8 & 13.6 & 1.0& \multicolumn{1}{c|}{}&      85.6         &       74.5       &      57.4        \\
Consistency& \multicolumn{1}{c|}{}&  48.4   &  25.0    &  3.0       & \multicolumn{1}{c|}{}& 25.6 & 10.8 & 0.4    & \multicolumn{1}{c|}{}&      86.4         &    75.7          &       56.6       \\
MultiTask& \multicolumn{1}{c|}{}&  48.6   &   27.4    &  3.6       & \multicolumn{1}{c|}{} & 19.0 & 2.8 & 0.2  & \multicolumn{1}{c|}{}& 86.4 & 72.8  & 49.4\\
\name& \multicolumn{1}{c|}{} & \textbf{72.6}     &  \textbf{44.8}    &   \textbf{8.6}    & \multicolumn{1}{c|}{} &  \textbf{90.0} & \textbf{85.4} & \textbf{79.4} & \multicolumn{1}{c|}{}&       \textbf{88.7}        &         \textbf{78.2}     &         \textbf{61.3}     \\ \hline\hline

Gaussian& \multicolumn{1}{c|}{\multirow{6}{*}{1.00}}& 40.4    &   22.0     &   4.4    & \multicolumn{1}{c|}{\multirow{6}{*}{0.50}}&  12.8 & 4.2 & 0.2     & \multicolumn{1}{c|}{\multirow{6}{*}{0.50}}                  &    78.0           &        67.9     &       48.1       \\
SWEEN& \multicolumn{1}{c|}{} &  41.6   &   25.0    &  6.8      & \multicolumn{1}{c|}{} &  21.4 & 12.8 & 3.6       & \multicolumn{1}{c|}{} &     {78.4}          &   69.1           &       51.4       \\
SmoothAdv& \multicolumn{1}{c|}{} &  39.8   &   25.6    &  6.6& \multicolumn{1}{c|}{} &  22.6 & 10.8 & 2.2  & \multicolumn{1}{c|}{}&         72.6      &        65.2      &     49.8         \\
Consistency& \multicolumn{1}{c|}{}&  40.0   &   27.2    &  9.2& \multicolumn{1}{c|}{} &  18.0 & 9.4 & 1.4& \multicolumn{1}{c|}{}&         75.5      &      70.8        &      52.1        \\
MultiTask& \multicolumn{1}{c|}{}&  37.0   &   24.8    &  7.4        & \multicolumn{1}{c|}{} & 16.2 & 6.4 & 0.4  & \multicolumn{1}{c|}{}&\textbf{78.6}& 69.5  & 50.6\\
\name& \multicolumn{1}{c|}{}& \textbf{78.8}    & \textbf{66.8}   &   \textbf{37.2}   & \multicolumn{1}{c|}{}&  \textbf{81.0} & \textbf{74.2} & \textbf{63.0}    & \multicolumn{1}{c|}{}&        {78.4}       &   \textbf{72.4}           &       \textbf{57.4}       \\ \hline
\end{tabular}}
\end{table*}

\renewcommand\arraystretch{1.15}
\begin{table*}[htbp]
\centering
\caption{{\small The empirical robust accuracy of different methods for Word50 when attacking on both word classification and letter classification under different metrics including $\ell_2$ and $\ell_{\infty}$.}}
\label{tab:attack_main_attribute}
\resizebox{0.65\linewidth}{!}{
\begin{tabular}{c|c|ccc|ccc}
\hline
\multirow{2}{*}{Method}        & \multirow{2}{*}{$\sigma$} & \multicolumn{3}{c|}{$\epsilon (\ell_2)$} & \multicolumn{3}{c}{$\epsilon (\ell_{\infty})$} \\ \cline{3-8} 
 & & 0.4          & 0.8         & 1.2         & 2/255           & 4/255          & 8/255          \\ \hline
Gaussian & \multirow{6}{*}{0.12} & 22.8 & 16.0 & 12.0 & 18.0 & 7.0 & 3.2 \\
SmoothAdv & & 46.8 & 40.2 & 34.0 & 42.2 & 29.8 & 15.8 \\
Consistency & & 34.8 & 26.6 & 22.6 & 29.2 & 17.8 & 9.2 \\
\name(Gaussian)& & 81.0 & 65.6 & 47.4 & 83.4 & 69.8 & 40.6  \\
\name(SmoothAdv)& &81.8 & 65.2 & 49.6 & 83.0 & 65.8 & 30.6  \\
\name(Consistency) & & \textbf{85.4} & \textbf{69.2} & \textbf{52.6} & \textbf{90.2} & \textbf{78.6} & \textbf{44.2} \\
\hline \hline
Gaussian & \multirow{6}{*}{0.25} & 21.4 & 11.4 & 7.0 & 18.0 & 7.2 & 2.0 \\
SmoothAdv & & 39.0 & 31.2 & 25.0 & 37.6 & 27.2 & 12.8 \\
Consistency & &33.2 & 26.0 & 19.8  & 31.6 & 20.8 & 8.6 \\
\name(Gaussian)& &84.0 & 69.6 & 54.4 & 86.2 & 76.2 & \textbf{55.4} \\
\name(SmoothAdv)& & 77.0 & 62.6 & 48.2  & 79.2 & 66.8 & 34.8 \\
\name(Consistency) & & \textbf{88.6} & \textbf{74.2} & \textbf{58.6} & \textbf{92.0} & \textbf{81.4} & 53.8 \\
\hline \hline
Gaussian & \multirow{6}{*}{0.50} & 17.0 & 8.2 & 4.2 & 16.2 & 7.2 & 1.8 \\
SmoothAdv & & 25.6 & 16.6 & 11.4 & 27.4 & 18.4 & 8.0 \\
Consistency & & 14.0 & 7.8 & 4.8 & 23.4 & 14.0 & 6.8 \\
\name(Gaussian)& & 73.6 & \textbf{61.4} & 45.2 & 76.8 & \textbf{66.6} & 42.4 \\
\name(SmoothAdv)& & 69.6 & 56.8 & 43.0 & 73.6 & 62.4 & 39.8 \\
\name(Consistency) & & \textbf{74.0} & 60.4 & \textbf{46.2} & \textbf{77.6} & 65.8 & \textbf{45.0} \\ \hline
\end{tabular}}
\end{table*}



\subsection{Exploration for the importance of the reasoning module}
\label{adx:n-way}
In this section, we explore the importance of the reasoning module by simply replacing the GCN in~\name with a linear classifier, which learns to discriminate along both the main sensor and knowledge sensors directly without a reasoning component. For simplicity, we denote this method as~\emph{SensingLinear} and similarly, we also train it with different main sensors, including Gaussian, SmoothAdv, and Consistency; the final certified results are shown in~\Cref{tab:n-way}. As we can see, although both main and knowledge sensors are aggregated in this linear classifier, the performance will still drop significantly without the reasoning part, which demonstrates the necessity of the construction of logical reasoning based on the output of both the main sensor and the knowledge sensors.

\subsection{Increasing the number of the base models for SWEEN}
\label{adx:ensemble}
We provide the experiment results for increasing the number of base models for SWEEN on AwA2. The magnitude of the smoothing noise is set to $0.50$ here, and we certify the ensemble method SWEEN with the number of base model $m \in \{ 3,6,10,15,20,30\}$, the corresponding result is shown in~\Cref{tab:ensemble_diff_num}. As we can see, the performance improvement from the increase of the base model is marginal, and the certified accuracy only improves a bit ($1\sim 2\%$) when we increase the base models from $3$ to $30$. On the contrary, with the domain knowledge and logic, such a phenomenon is alleviated as shown in~\Cref{tab:pick} with a $10\sim30\%$ improvement.

\subsection{Additional Experiment Results and Details for Empirical Attack}
\label{sec:emp_attack_detail}
    \subsubsection{Experiment details} We implement the empirical untargeted attack as follows: $(1.)$ Take the mean of the output confidence from the soft base main sensor for the corrupted input image with $100$ Gaussian noise; $(2.)$ Use projected gradient descent (PGD)~\cite{madry2018towards} to minimize the mean confidence for the truth label and get the corresponding adversarial image; $(3.)$ Next, this adversarial image will be sent to all the knowledge sensors to get the new adversarial sensing vector $\bm{z'}$; $(4.)$ Do the same hypothesis-test-based prediction procedure in~\cite{cohen2019certified} with our method~\name to check if the attack is successful with $\bm{z'}$. The test images here are the same as those in the certification part. For $\ell_{\infty}$ attack, the number of update steps is also fixed to $40$, the attack step size is set to $1/255$, and the full results are shown in~\Cref{tab:attack_inf_all}. 
    
    \subsubsection{Attack transferability} Based on these test images, we also explore the attack transferability between $12$ sensors (one main sensor and eleven random picked attribute sensors, all are trained under $\sigma = 0.50$) on AwA2 under $\ell_2$ perturbation size $\epsilon = 3.0$. Besides, the attack step size is set to $0.2$, the number of update steps is set to $100$, and the final results are shown in~\Cref{fig:heatmap}.
    
    \subsubsection{Attacking with the attributes} We also conduct the experiments which construct the adversarial image by attacking the main sensor and all the attribute sensors at the same time on Word50. In other words, the PGD attack is implemented to increase the mean of the loss from both the word and letter classifications. In this case, we also report the empirical robust accuracy of the main sensor (Gaussian/SmoothAdv/Consistency) used in~\name on these new adversarial images, and the results are shown in~\Cref{tab:attack_main_attribute}. As we can see, even though the attack seems stronger here, with the incorporation of the domain knowledge, our method is still much more robust than the baselines.

\vspace{-1mm}
\subsection{Case Study on AwA2}
\label{sec:case_awa}
\begin{figure*}
    \centering
    \includegraphics[width=\textwidth]{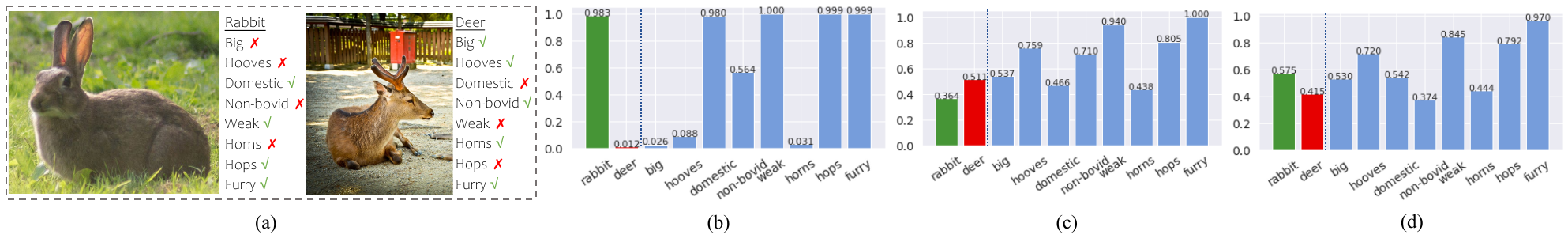}
    \vspace{-4mm}
    \caption{The illustration of the change of confidence. (a) the attributes for the rabbit and mole; (b) the original confidence before the attack; (c) the confidence for the adversarial image, which is obtained by attacking the main sensor; (d) the recovered confidence from our method~\name for the adversarial image. The ground truth is~\emph{``rabbit''}.}
    \label{fig:case1}
\end{figure*}

\begin{figure*}
    \centering
    \includegraphics[width=\textwidth]{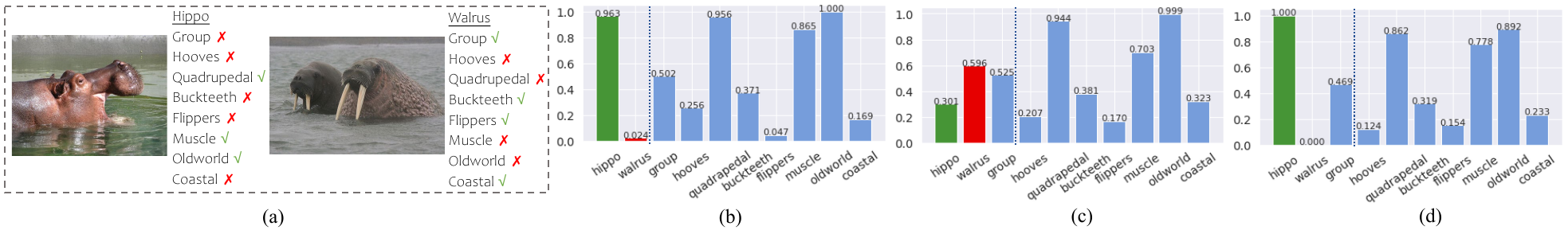}
    \vspace{-4mm}
    \caption{The illustration of the change of confidence. (a) the attributes for the hippopotamus and walrus; (b) the original confidence before the attack; (c) the confidence for the adversarial image, which is obtained by attacking the main sensor; (d) the recovered confidence from our method~\name for the adversarial image. The ground truth is~\emph{``hippopotamus''}.}
    \label{fig:case2}
\end{figure*}

\begin{figure*}
    \centering
    \includegraphics[width=\textwidth]{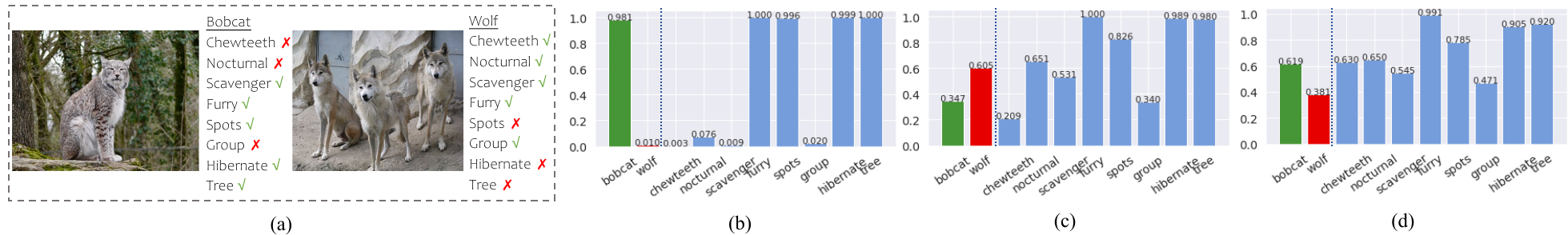}
    \vspace{-4mm}
    \caption{The illustration of the change of confidence. (a) the attributes for the bobcat and wolf; (b) the original confidence before the attack; (c) the confidence for the adversarial image which is obtained by attacking the main sensor; (d) the recovered confidence from our method~\name for the adversarial image. The ground truth is~\emph{``bobcat''}.}
    \label{fig:case3}
\end{figure*}

\begin{figure*}
    \centering
    \includegraphics[width=\textwidth]{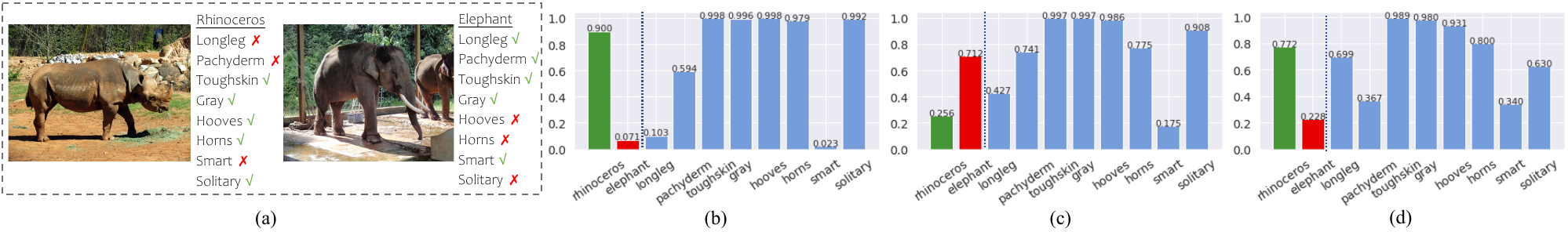}
    \vspace{-4mm}
    \caption{The illustration of the change of confidence. (a) the attributes for the rhinoceros and elephant; (b) the original confidence before the attack; (c) the confidence for the adversarial image, which is obtained by attacking the main sensor; (d) the recovered confidence from our method~\name for the adversarial image. The ground truth is~\emph{``rhinoceros''}.}
    \label{fig:case4}
\end{figure*}

\begin{figure*}
    \centering
    \includegraphics[width=\textwidth]{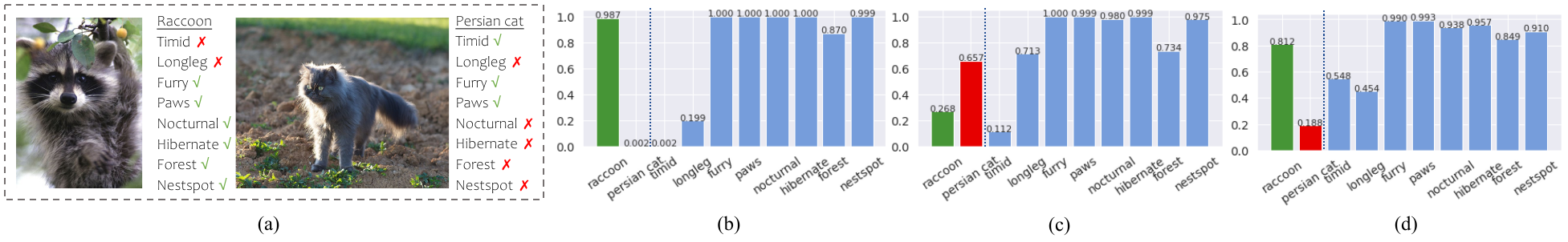}
    \vspace{-4mm}
    \caption{The illustration of the change of confidence. (a) the attributes for the raccoon and Persian cat; (b) the original confidence before the attack; (c) the confidence for the adversarial image, which is obtained by attacking the main sensor; (d) the recovered confidence from our method~\name for the adversarial image. The ground truth is~\emph{``raccoon''}.}
    \label{fig:case5}
\end{figure*}

In this section, we provide more case studies on AwA2, which are shown in~\Cref{fig:case1,fig:case2,fig:case3,fig:case4,fig:case5}. In specific, we adopt untargeted attacks here; the $\ell_2$ perturbation size is set to $3.0$, the attack step size is set to $0.2$, the number of update steps is set to $100$, and all the sensors used here are trained under $\sigma = 0.50$. 

\end{document}